\pdfoutput=1
\documentclass[11pt]{article}

\usepackage[preprint]{acl}

\usepackage{times}
\usepackage{latexsym}
\usepackage[T1]{fontenc}
\usepackage[utf8]{inputenc}
\usepackage{microtype}
\usepackage{inconsolata}

\usepackage{amsmath,amssymb,mathtools,amsthm,bm,nicefrac}
\usepackage{graphicx,booktabs,multirow,xcolor,longtable,makecell}
\usepackage{colortbl}
\usepackage{tikz}
\usetikzlibrary{arrows.meta,positioning,fit,calc}
\usepackage{algorithm}
\usepackage[noend]{algorithmic}
\usepackage{placeins}
\usepackage{xurl}
\usepackage[capitalize,noabbrev]{cleveref}

\theoremstyle{plain}
\newtheorem{theorem}{Theorem}
\newtheorem{proposition}[theorem]{Proposition}
\newtheorem{lemma}[theorem]{Lemma}

\AddToHook{env/proposition/begin}{\crefalias{theorem}{proposition}}
\AddToHook{env/lemma/begin}{\crefalias{theorem}{lemma}}
\AddToHook{env/corollary/begin}{\crefalias{theorem}{corollary}}
\theoremstyle{definition}
\newtheorem{definition}{Definition}

\theoremstyle{remark}

\title{SOLO: Towards Sample-Faithful and Detectable LLM Watermarking under Repeated Sampling via Content-Anchored Single-Owner Contexts}

\author{
  \textbf{Zhongliang Guo}\quad
  \textbf{Yan Lin}\quad
  \textbf{Yifei Qian}\quad
  \textbf{Shuai Zhao}\\
  \textbf{Luwei Xiao}\quad
  \textbf{Daizong Liu}\quad
  \textbf{Chun Pong Lau}
}

\begin{document}
\maketitle

\begin{abstract}
Providers watermark the outputs of large language models, while applications sample many responses per prompt for self-consistency, best-of-n selection, and uncertainty estimation. We show that deployed configurations of distortion-free watermarks, which deduplicate keyed contexts only within a response, couple the responses of a request: answer templates make all responses query the watermark with the same context before the answer, so collisions across responses, usually assumed rare, become the norm. Key pools reduce or avoid the coupling at the cost of multiple testing, and the exact remedy of a context history shared across the request leaves many positions unwatermarked. We propose SOLO, which keeps a request-scoped context history but anchors its keyed contexts on content selected by min-hash, so that siblings rarely compete for a context and random edits stay local. Its responses are jointly distributed as independent unwatermarked responses, each detectable with a single key. On five language models from three families and three benchmarks, SOLO preserves self-consistency and uncertainty estimation and, at a 1\% false positive rate, detects more responses than four-token key pools on most tasks, also after randomly editing 10\% of the tokens, and more than the context history on most clean tasks. Code is available at \href{https://github.com/ZhongliangGuo/solo-watermark}{SOLO}.
\end{abstract}

\section{Introduction}\label{sec:intro}

As large language models (LLMs) write an increasing share of online text, providers watermark their outputs at generation time, so that a text can be traced to its source model without changing what the model would have said~\cite{dathathri2024scalable,aaronson2023watermarking,kuditipudi2024robust}. Distortion-free schemes run in production assistants~\cite{dathathri2024scalable} and in vLLM, which watermarks every request of a watermarked deployment unless it opts out~\cite{vllm2026watermarking}. Meanwhile, applications read several responses per prompt: self-consistency votes over sampled solutions~\cite{wang2023selfconsistency}, test-time scaling selects among candidates~\cite{brown2024monkeys,snell2025scaling}, and uncertainty estimators flag answers on which the responses disagree~\cite{kuhn2023semantic,farquhar2024detecting,manakul2023selfcheckgpt}. These pipelines assume that the $K$ responses to one prompt are independent draws, and a watermark that couples them breaks this premise silently.

\begin{figure}[t]
\centering
\parbox[c][0.75\columnwidth][c]{\columnwidth}{\centering\includegraphics[width=\columnwidth,height=0.75\columnwidth,keepaspectratio]{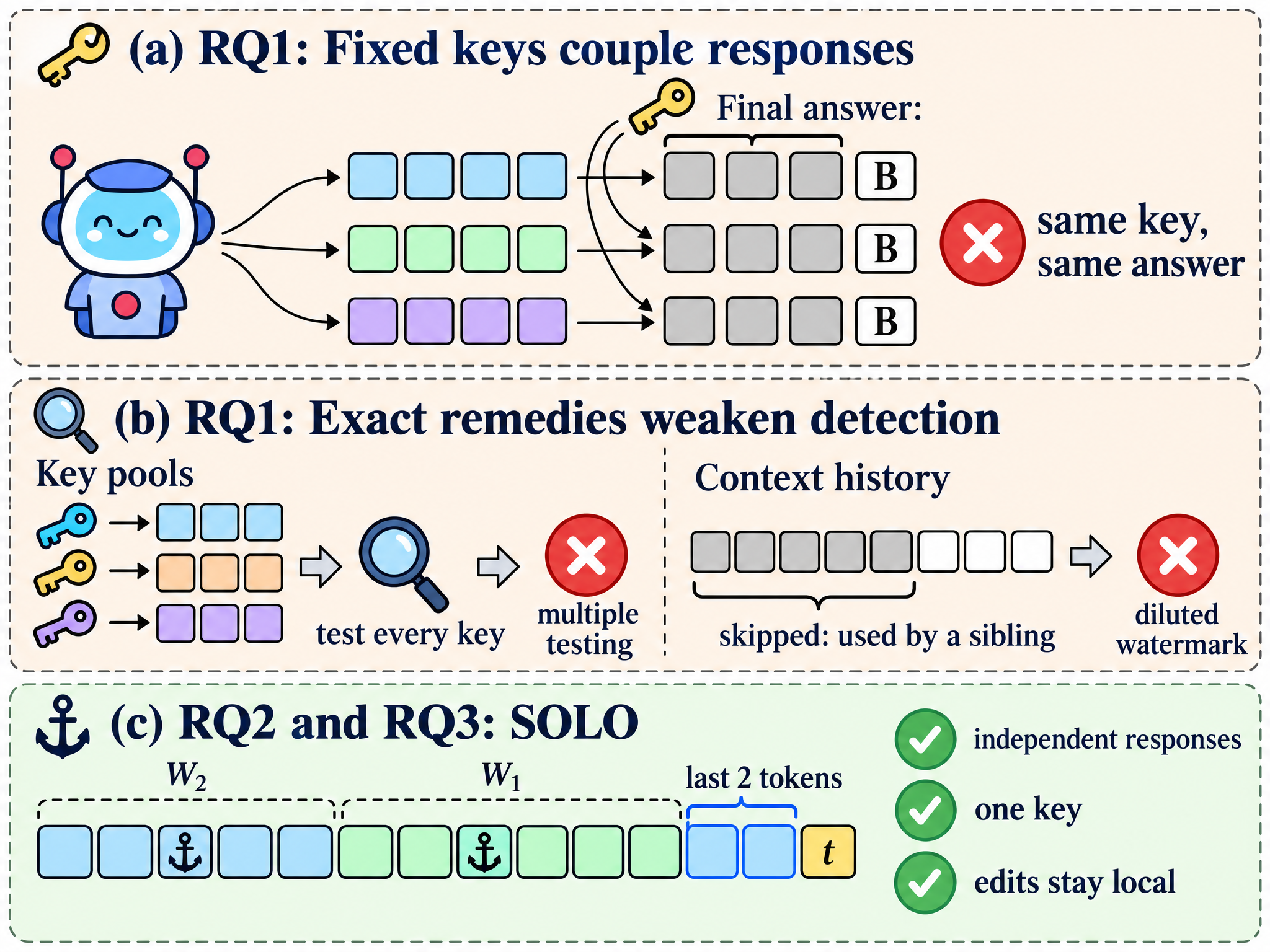}}
\caption{Overview. (a) A fixed key couples the responses at the answer. (b) Exact remedies pay at detection. (c) SOLO keeps them independent with one key.}\label{fig:overview}
\end{figure}

Existing watermarks reuse one key across responses~\cite{aaronson2023watermarking,dathathri2024scalable}, inject true randomness~\cite{fu2024gumbelsoft}, give responses different keys~\cite{liu2025waterbag,aremu2026randomized}, or skip contexts that earlier generations used~\cite{hu2024unbiased,dathathri2024scalable}. We ask three questions (\cref{fig:overview}). \textbf{RQ1:} Do deployed distortion-free watermarks keep the responses of a request independent, and what do remedies cost? \textbf{RQ2:} Can one watermark offer (i) \emph{sample faithfulness}, responses exactly as independent as unwatermarked ones, (ii) \emph{single-key detection}, and (iii) \emph{edit locality}? \textbf{RQ3:} Why does it work, and does its lead survive token edits?

For \textbf{RQ1} (\cref{sec:revisiting}), answer templates make all responses query the watermark with the same context right before the answer, which we call \emph{template funnels}. Under the default watermark of vLLM, 99.9\% of TriviaQA requests to Qwen2.5-7B-Instruct return ten identical short answers, and the vote share predicts correctness no better than chance. In theory, dual-key routing leaves at least half of this coupling, and the existing exact remedies pay at detection: stratified key pools through multiple testing, and a request-scoped context history through dilution. For \textbf{RQ2} (\cref{sec:method,sec:theory,sec:exp-main}), we propose SOLO, which keeps the request-scoped history, so that every keyed context has a single owner (hence the name) and the responses are exactly independent, but anchors its contexts on content selected by a min-hash. SOLO is admissible (\cref{sec:exp-setup}) for repeated sampling on all 26 model, task and format combinations we test, and at a 1\% false positive rate (FPR) it detects significantly more clean responses than stratified keys on 21 of them and than the context history on 20. For \textbf{RQ3} (\cref{sec:exp-why}), the anchors raise the watermarked share from 53\% for four-token contexts to 68\% (Qwen2.5-7B, GSM8K, JSON) while an edit still costs only a few positions, and SOLO keeps a significant lead over stratified keys on most combinations after random edits of 10\% of the tokens.

\section{Related Work}\label{sec:related}

\noindent\textbf{Generation-time watermarks.}
One family biases the next-token distribution toward a keyed subset of the vocabulary~\cite{kirchenbauer2023watermark,zhao2024provable}, which changes the output distribution. Distortion-free schemes instead replace the randomness of sampling with keyed pseudorandomness, through the Gumbel-max decoder~\cite{aaronson2023watermarking,fernandez2023threebricks}, inverse transform sampling~\cite{kuditipudi2024robust}, reweighting~\cite{hu2024unbiased,wu2024dipmark}, tournament sampling~\cite{dathathri2024scalable}, undetectable sampling~\cite{christ2024undetectable}, or pseudorandom codes~\cite{christ2024prc}. They are the ones deployed: SynthID-Text in a production assistant~\cite{dathathri2024scalable} and Gumbel-max sampling in vLLM~\cite{kwon2023vllm,vllm2026watermarking}. Several define guarantees over multiple generations~\cite{hu2024unbiased,dathathri2024scalable,kuditipudi2024robust,christ2024undetectable}, but none analyzes the coupling that templates create between the responses of a request.

\noindent\textbf{Keys, context histories, and diversity.}
Key collisions bias distortion-free watermarks~\cite{wu2024collision}, and GumbelSoft replaces the arg max of the Gumbel watermark with softmax sampling to restore diversity, giving up the per-response guarantee~\cite{fu2024gumbelsoft}, a relaxation also adopted for best-of-$n$ selection~\cite{verma2026watermarking}. Key pools give generations different keys~\cite{liu2025waterbag,aremu2026randomized} and must test them all. A history of used context codes, reset after every batch, makes the batch exactly distortion-free~\cite{hu2024unbiased}, and SynthID-Text defines the same masking but configures it for one response because it reduces detectability~\cite{dathathri2024scalable}. \Cref{app:related} discusses these works in detail.

\noindent\textbf{Repeated sampling and uncertainty.}
Self-consistency~\cite{wang2023selfconsistency}, repeated sampling with selection~\cite{brown2024monkeys,snell2025scaling}, and sampling-based uncertainty estimators~\cite{kuhn2023semantic,farquhar2024detecting,manakul2023selfcheckgpt,vashurin2025cocoa} all assume independent responses.

\section{Problem Formulation}\label{sec:formulation}
\begin{table*}[t]
\centering
\resizebox{\textwidth}{!}{%
\begin{tabular}{lccccccccccccccc}
\toprule
 & \multicolumn{3}{c}{TriviaQA} & \multicolumn{3}{c}{TriviaQA, CoT} & \multicolumn{3}{c}{TriviaQA, JSON} & \multicolumn{3}{c}{GSM8K} & \multicolumn{3}{c}{GSM8K, JSON} \\
\cmidrule(lr){2-4} \cmidrule(lr){5-7} \cmidrule(lr){8-10} \cmidrule(lr){11-13} \cmidrule(lr){14-16}
Watermark & \makecell{Ident.\\(\%)} & \makecell{maj@10\\(\%)} & \makecell{AUROC\\(\%)} & \makecell{Ident.\\(\%)} & \makecell{maj@10\\(\%)} & \makecell{AUROC\\(\%)} & \makecell{Ident.\\(\%)} & \makecell{maj@10\\(\%)} & \makecell{AUROC\\(\%)} & \makecell{Ident.\\(\%)} & \makecell{maj@10\\(\%)} & \makecell{AUROC\\(\%)} & \makecell{Ident.\\(\%)} & \makecell{maj@10\\(\%)} & \makecell{AUROC\\(\%)} \\
\midrule
\rowcolor[HTML]{ECECEC} Unwatermarked & 36 & 57.1 & 83 & 2 & 63.5 & 87 & 0 & 59.3 & 85 & 0 & 92.8 & 89 & 0 & 92.8 & 91 \\
Gumbel (default) & \cellcolor[HTML]{FFF0B8} 100 & \cellcolor[HTML]{FBE0DA} 54.7 & \cellcolor[HTML]{FFF0B8} 50 & \cellcolor[HTML]{FBE0DA} 88 & \cellcolor[HTML]{FBE0DA} 59.4 & \cellcolor[HTML]{FBE0DA} 52 & \cellcolor[HTML]{FBE0DA} 99 & \cellcolor[HTML]{FBE0DA} 54.4 & \cellcolor[HTML]{FBE0DA} 50 & 8 & \cellcolor[HTML]{FBE0DA} 90.7 & \cellcolor[HTML]{FBE0DA} 82 & \cellcolor[HTML]{FBE0DA} 23 & \cellcolor[HTML]{FBE0DA} 88.3 & \cellcolor[HTML]{FBE0DA} 76 \\
Gumbel, dedup \texttt{all} & 36 & 57.2 & 83 & 20 & \cellcolor[HTML]{FBE0DA} 60.2 & \cellcolor[HTML]{FBE0DA} 78 & \cellcolor[HTML]{FBE0DA} 91 & \cellcolor[HTML]{FBE0DA} 54.2 & \cellcolor[HTML]{FBE0DA} 52 & 1 & 92.8 & 87 & 3 & \cellcolor[HTML]{FBE0DA} 88.7 & \cellcolor[HTML]{FBE0DA} 85 \\
Gumbel + prompt nonce & \cellcolor[HTML]{FBE0DA} 81 & \cellcolor[HTML]{FBE0DA} 53.6 & \cellcolor[HTML]{FBE0DA} 62 & \cellcolor[HTML]{FBE0DA} 41 & \cellcolor[HTML]{FBE0DA} 59.3 & \cellcolor[HTML]{FBE0DA} 68 & \cellcolor[HTML]{FBE0DA} 42 & \cellcolor[HTML]{FBE0DA} 54.1 & \cellcolor[HTML]{FBE0DA} 69 & 2 & \cellcolor[HTML]{FBE0DA} 90.8 & 87 & 4 & \cellcolor[HTML]{FBE0DA} 89.2 & \cellcolor[HTML]{FBE0DA} 84 \\
Dual-key, $\lambda = 0.1$ & \cellcolor[HTML]{FBE0DA} 69 & \cellcolor[HTML]{FBE0DA} 55.6 & \cellcolor[HTML]{FBE0DA} 69 & \cellcolor[HTML]{FBE0DA} 24 & \cellcolor[HTML]{FBE0DA} 60.6 & \cellcolor[HTML]{FBE0DA} 76 & 12 & \cellcolor[HTML]{FBE0DA} 53.7 & \cellcolor[HTML]{FBE0DA} 78 & 0 & \cellcolor[HTML]{FBE0DA} 91.2 & 87 & 0 & \cellcolor[HTML]{FBE0DA} 90.9 & 91 \\
Dual-key, $\lambda = 0.5$ & \cellcolor[HTML]{FBE0DA} 59 & \cellcolor[HTML]{FBE0DA} 54.9 & \cellcolor[HTML]{FBE0DA} 76 & 14 & \cellcolor[HTML]{FBE0DA} 61.7 & \cellcolor[HTML]{FBE0DA} 81 & 4 & \cellcolor[HTML]{FBE0DA} 56.8 & 82 & 0 & 92.2 & 89 & 0 & 91.9 & 91 \\
\bottomrule
\end{tabular}
}
\caption{RQ1: vLLM 0.30 watermarks under repeated sampling (Qwen2.5-7B, $K = 10$). Ident.: ten identical responses. Coral, yellow (quoted): at least 20 (Ident.), 1.5 (maj@10), or 5 (AUROC) points worse.}\label{tab:revisit}
\end{table*}
\noindent\textbf{Repeated sampling.}
Let $p(\cdot \mid x, y_{<t})$ be the next-token distribution of a language model over a vocabulary $\mathcal{V}$. A \emph{request} asks for $K$ responses $y^{(1)}, \ldots, y^{(K)}$ of one prompt $x$, and repeated-sampling pipelines read the agreement among their answers $A^{(i)} = g(y^{(i)})$ as the model's confidence.

\noindent\textbf{Keyed samplers.}
At step $t$ of response $i$, a context map $\phi$ produces a keyed context $c^{(i)}_t = \phi(y^{(i)}_{<t})$, typically the last $h$ tokens. A pseudorandom function (PRF) $F$ maps the key $k^{(i)}$ of the response and the context to $\xi = F_{k^{(i)}}(c^{(i)}_t) \in (0,1)^{|\mathcal{V}|}$, and a decoder selects $y^{(i)}_t = \Gamma(\pi, \xi)$ from $\pi = p(\cdot \mid x, y^{(i)}_{<t})$. The Gumbel-max decoder of vLLM and many distortion-free schemes~\cite{aaronson2023watermarking,fernandez2023threebricks} is $\Gamma(\pi, \xi) = \arg\max_{v} \xi_v^{1/\pi_v}$. A context that already occurred in the same response is sampled with true randomness and not scored (within-response deduplication), as in SynthID-Text and vLLM~\cite{dathathri2024scalable,vllm2026watermarking}.

\noindent\textbf{Detection and edits.}
For one key, the detector sums $S = \sum_{t} -\log(1 - \xi_{y_t})$ over the first occurrences of keyed contexts, which follows $\mathrm{Gamma}(n, 1)$ for $n$ scored tokens of a text independent of the key~\cite{fernandez2023threebricks}. A scheme that may have used any of $L$ keys tests all of them, for instance with the minimum $p$-value and a \v{S}id\'ak correction~\cite{sidak1967rectangular}. A response may be edited by token substitutions, insertions, and deletions, and the \emph{exposure} $e_{\phi}$ of a context map is the number of scored positions whose (context, token) pair no longer matches any pair produced at generation time after one edit.

\noindent\textbf{Goal.}
Following \citet{hu2024unbiased} and \citet{dathathri2024scalable}, we require the following.

\begin{definition}[$K$-sample distortion-freeness]\label{def:ksample}
A watermarking scheme is \emph{$K$-sample distortion-free} if, for every prompt $x$, the joint distribution of the $K$ responses of a request, taken over the randomness of the key, equals $p(\cdot \mid x)^{\otimes K}$.
\end{definition}

Deployed configurations guarantee only the case $K = 1$. We seek a scheme that is $K$-sample distortion-free for every $K$, whose responses are each detectable with a single key and a large watermarked share of their tokens, and whose watermark is affected only locally by edits.

\noindent\textbf{Assumptions.}
\textbf{(A1) Ideal PRF:} $F_{k}(c)$ are independent uniform vectors for distinct pairs $(k, c)$, independent of the model and of the true randomness (an idealization that cryptographic PRFs approximate). \textbf{(A2) Unbiased decoder:} $\mathbb{P}_{\xi}[\Gamma(\pi, \xi) = v] = \pi_v$, as for Gumbel-max. \textbf{(A3) Within-response deduplication}, as above. Under them, all schemes we study except GumbelSoft~\cite{fu2024gumbelsoft} are per-response distortion-free.

\section{RQ1: Revisiting Watermarks under Repeated Sampling}\label{sec:revisiting}

\textbf{RQ1} asks whether deployed keyed samplers, which deduplicate contexts only within a response because context collisions are rare~\cite{hu2024unbiased}, keep the responses of a request independent. We show that this fails at \emph{template funnels}, deterministic token runs shared by the responses, and quantify what each remedy recovers and costs.

\noindent\textbf{Settings.}
\Cref{tab:revisit} uses the watermarks shipped in vLLM 0.30 (Gumbel-max over $h = 4$ response tokens, \citealp{vllm2026watermarking}), and \cref{sec:revisiting-keys} our keyed sampler with vLLM's default (\cref{app:impl}), with Qwen2.5-7B-Instruct~\cite{qwen2025qwen25} at temperature 1, $K = 10$, three seeds, and 600 held-out TriviaQA~\cite{joshi2017triviaqa} and GSM8K~\cite{cobbe2021gsm8k} items in the formats of \cref{sec:exp-setup} (TriviaQA also as a short answer). We report the share of requests with ten identical responses, the accuracy of the majority answer (maj@10), and the AUROC of its vote share for its correctness. \Cref{app:revisit} adds Ministral-3-8B.

\subsection{Fixed Keys Couple Responses}\label{sec:revisiting-funnel}
With a fixed key, any two responses that reach the same context receive the same randomness.

\begin{proposition}[Template funnels]\label{prop:funnel}
Under (A1) to (A3), consider a fixed key and contexts of the last $h$ response tokens. (a) Up to the first position whose context repeats within a response, all $K$ responses of a request are identical. (b) If every response generates the same deterministic run of $L_f \ge h$ tokens right before a decision, and the context of that decision has not occurred earlier in either response, then any two responses evaluate the PRF on the same pair at that decision, whatever they generated before the run, and prompt-side randomization and an unwatermarked prefix that ends before the run do not change this. (c) With the Gumbel-max decoder, the two decisions then coincide with probability at least $(1 - \mathrm{TV}(\pi, \pi'))/(1 + \mathrm{TV}(\pi, \pi'))$, where $\mathrm{TV}$ is the total variation distance between their decision distributions $\pi$ and $\pi'$.
\end{proposition}

Part (b) explains why remedies on the prompt or the start of the response fail: templates create deterministic runs right before the answer, which re-synchronize responses that diverged earlier.

\noindent\textbf{Results.}
As shown in \cref{tab:revisit}, the default configuration returns ten identical short answers for 99.9\% of the TriviaQA questions, and the AUROC of the vote share falls to 50\%, the value of a random guess. Deduplicating contexts that overlap the prompt (vLLM's \texttt{all} mode) restores short answers and plain GSM8K but not templated formats, and a random nonce in each prompt recovers only part of the gap. Funnels are pervasive even in the \texttt{all} mode, because the responses state their answer right after reasoning phrases they share, and for $h = 8$ a marker that fills the context roughly doubles the excess agreement, as part (b) predicts (\cref{fig:funnel} in \cref{app:revisit}).

\subsection{Injected Randomness Leaves Coupling}\label{sec:revisiting-remedies}
Dual-key routing~\cite{textseal2026}, which vLLM 0.30 ships, and softmax relaxation~\cite{fu2024gumbelsoft} inject true randomness into keyed decoding. \Cref{thm:inflation} quantifies the coupling they leave.

\begin{theorem}[Agreement inflation]\label{thm:inflation}
Under (A1) to (A3), suppose that the answer of each response is the token chosen at a decision step, that the decision contexts of two responses coincide and are new within each response with probability $\rho_F$, independently of the key assignment and of the routing at that step, that given the joint history of both responses, each decision follows the answer distribution $\pi$ and, unless they use the same pseudorandom value, the two decisions are independent. Then
\begin{equation}\label{eq:inflation}
\mathbb{P}\bigl(A^{(i)} = A^{(j)}\bigr) = \sum_a \pi_a^2 + \rho_F\, \kappa_S \Bigl(1 - \sum_a \pi_a^2\Bigr),
\end{equation}
where $\kappa_S$, the probability that the two responses use the same pseudorandom value at a colliding context, is $1$ for a fixed key, $\lambda^2 + (1-\lambda)^2$ for dual-key routing with probability $\lambda$ of the second key, $1/L$ for keys drawn at random from a pool of $L$, and $0$ for stratified keys (a distinct key per response, $K \le L$) and for a request-scoped context history. If this holds for every pair of $K$ exchangeable responses, the variance of the vote share of any fixed answer is inflated by the factor $1 + (K - 1)\rho_F \kappa_S$.
\end{theorem}

Dual-key routing thus leaves at least half of the fixed-key coupling ($\kappa_S \ge 1/2$), $82\%$ at the default $\lambda = 0.1$, and still couples TriviaQA responses (\cref{tab:revisit}), and a toy model matches Eq.~\eqref{eq:inflation} for every scheme (\cref{app:theorychecks}).

\subsection{Exact Remedies Weaken Detection}\label{sec:revisiting-keys}
\noindent\textbf{Key multiplicity.}
Stratified keys ($\kappa_S = 0$) make the responses independent, but detection must test every key: at a 1\% FPR, the \v{S}id\'ak correction over ten keys raises the critical standardized score from $2.33$ to $3.09$. On GSM8K, stratified keys keep maj@10 and the AUROC but detect only 33.1\% of the JSON responses at 1\% FPR, against 51.6\% for the fixed key, which couples them (\cref{tab:main}, and 8.7\% identical requests in \cref{app:uq}).

\noindent\textbf{Context histories.}\label{sec:revisiting-history}
A request-scoped context history removes the coupling exactly~\cite{hu2024unbiased}: a response skips the watermark at a context a sibling has used, so no keyed value is evaluated twice (\cref{thm:exact}). The price is dilution, because siblings share many local phrases. With contexts of the last four tokens, only $\omega$ = 53\% of the scored positions of a held-out GSM8K response in JSON format carry the watermark, and the history detects 28.6\% of the responses, fewer than stratified keys despite testing one key (\cref{tab:main}).

\par\smallskip\noindent\colorbox[HTML]{E8F1FB}{\parbox{\dimexpr\columnwidth-2\fboxsep\relax}{\small\textbf{Finding 1.} Deployed distortion-free watermarks couple the responses of a request at template funnels, and injected randomness leaves part of this coupling. The existing exact remedies keep the responses independent only at a cost in detection power, so we keep the exact history and redesign its context map.}}\par\smallskip

\section{The Proposed Method}\label{sec:method}

\begin{figure*}[t]
\centering
\parbox[c][0.3333\textwidth][c]{\textwidth}{\centering\includegraphics[width=\textwidth,height=0.3333\textwidth,keepaspectratio]{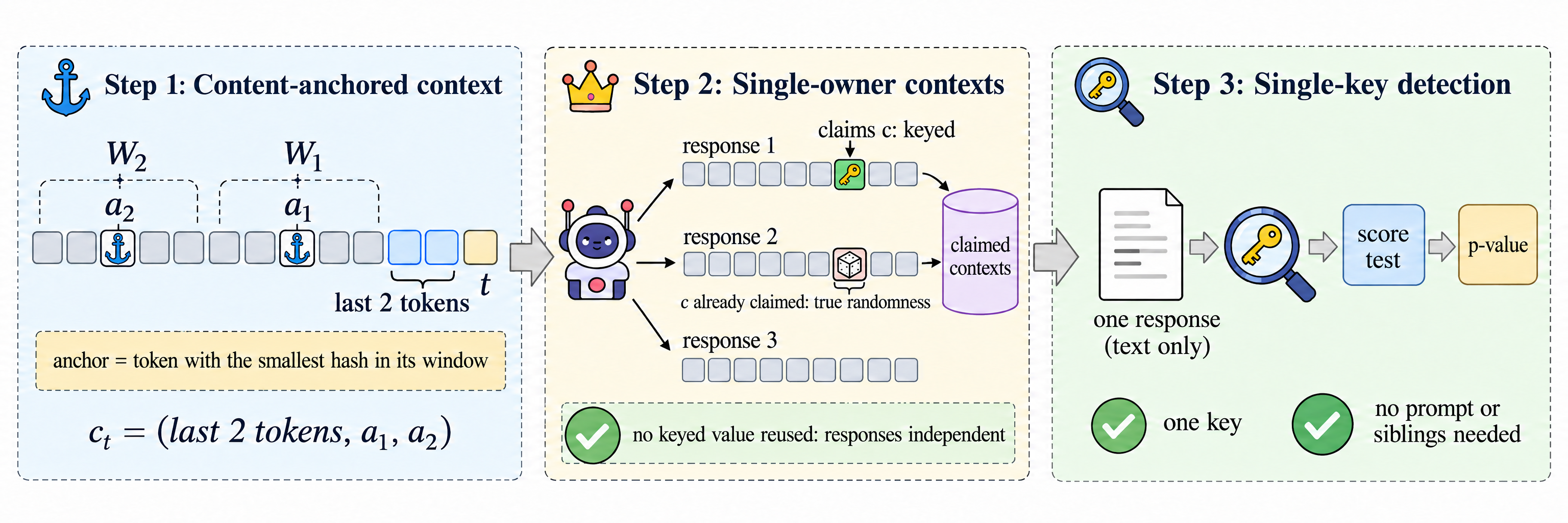}}
\caption{SOLO. Step 1: the keyed context is the last two tokens and two min-hash anchors. Step 2: a context owned by a sibling gets true randomness. Step 3: single-key detection.}\label{fig:pipeline}
\end{figure*}

\noindent\textbf{Design idea.}
To answer \textbf{RQ2}, we keep the exact request-scoped context history and redesign its contexts: since the history is $K$-sample distortion-free for any context map (\cref{thm:exact}), the map can be specific to each response, so that siblings rarely contest a context, and local, so that an edit affects few positions. As shown in \cref{fig:pipeline} and \cref{alg:solo}, SOLO has three components, single-owner contexts, content-anchored contexts, and content-defined anchor selection, which meet requirements (i), (ii), and (iii). The last two are our contribution.

\begin{algorithm}[t]
\caption{SOLO generation for one request}\label{alg:solo}
\small
\begin{algorithmic}[1]
\REQUIRE prompt $x$, $K$ responses, key $k$, local width $h$, windows $W$%
\STATE $\mathcal{U} \leftarrow \emptyset$; $y^{(i)} \leftarrow ()$ for $i = 1, \ldots, K$
\FOR{$t = 1, 2, \ldots$ until all responses end}
  \FOR{$i = 1, \ldots, K$ (unfinished)}
    \STATE $\pi \leftarrow p(\cdot \mid x, y^{(i)}_{<t})$; $c \leftarrow \phi(y^{(i)}_{<t})$ by Eqs.~\eqref{eq:context} and \eqref{eq:minhash}
    \IF{$c \notin \mathcal{U}$ and $c$ is new within $y^{(i)}$}
      \STATE $\mathcal{U} \leftarrow \mathcal{U} \cup \{c\}$; $y^{(i)}_t \leftarrow \arg\max_v F_{k}(c)_v^{1/\pi_v}$
    \ELSE
      \STATE $y^{(i)}_t \sim \pi$ (true randomness)
    \ENDIF
  \ENDFOR
\ENDFOR
\RETURN $y^{(1)}, \ldots, y^{(K)}$
\end{algorithmic}
\end{algorithm}

\subsection{Single-Owner Contexts}\label{sec:method-owner}
For requirement (i), SOLO keeps the set $\mathcal{U}$ of context values claimed in the request. A response that reaches context $c^{(i)}_t \notin \mathcal{U}$ claims it and decodes with $\Gamma(\pi, F_{k}(c^{(i)}_t))$. Otherwise it samples with true randomness. Simultaneous claims are resolved in a fixed response order, and one provider key $k$ serves all responses. This is the context history of \citet{hu2024unbiased} scoped to one request.

\subsection{Content-Anchored Contexts}\label{sec:method-anchor}
With contexts of the last $h$ tokens, siblings share so many phrases that single ownership leaves too few positions watermarked (\cref{sec:revisiting-history}), and longer contexts let one edit destroy more positions (\cref{app:analysis}). For requirement (ii), SOLO keeps the local part short and adds anchors that summarize a longer history:
\begin{equation}\label{eq:context}
\phi(y_{<t}) = \bigl(y_{t-h}, \ldots, y_{t-1},\; a_1(t), \ldots, a_{|W|}(t)\bigr),
\end{equation}
where $a_w(t)$ is one token of the window $[t - u_w, t - \ell_w)$. Siblings share this context only if they agree on the last $h$ tokens and on every anchor, so the anchors raise the fraction $\omega$ of scored positions that carry the watermark. With $h = 2$ and the windows $[t-12, t-2)$ and $[t-32, t-12)$, $\omega$ rises by 7 to 17 points over four-token contexts on all 26 combinations we evaluate.

\subsection{Content-Defined Anchor Selection}\label{sec:method-minhash}
An anchor at a fixed distance $u$ changes for $u$ positions after an insertion or deletion, and a prefix salt~\cite{christ2024undetectable} changes every later context. For requirement (iii), SOLO selects each anchor as the token of the window that minimizes a hash $\eta: \mathcal{V} \to \{0, \ldots, 2^{32} - 1\}$,
\begin{equation}\label{eq:minhash}
a_w(t) = \arg\min_{y_j : \, t - u_w \le j < t - \ell_w} \eta(y_j),
\end{equation}
a min-hash selection~\cite{broder1997resemblance} with an injective $\eta$ (empty windows give a padding symbol). An edit moves an anchor only if a token entering or leaving the window is its minimum, so for a hash independent of the text the expected exposure per edit is bounded independently of the length (\cref{prop:exposure}). Our experiments use a fixed public hash. A deployment should key $\eta$ with the provider key (\cref{app:method}).

\subsection{Detection}\label{sec:method-detect}
SOLO uses the single-key Gumbel score test of \cref{sec:formulation} with the map of Eq.~\eqref{eq:context}, without the prompt or the siblings: an unclaimed position behaves like an unwatermarked token. Generation adds $O(K T)$ hash operations and memory for $K$ responses of length $T$, and detection costs $O(T)$ (\cref{app:method}).

\section{Theoretical Analysis of SOLO}\label{sec:theory}

We analyze the components of SOLO under (A1) to (A3), with proofs in \cref{app:proofs}.

\noindent\textbf{Exactness.}
The next result restates the multi-shot undetectability of a history reset per batch~\citep[Corollary~15]{hu2024unbiased} for one request.

\begin{theorem}[Exact $K$-sample distortion-freeness]\label{thm:exact}
Under (A1) to (A3), for any context map $\phi$ and any $K$, let simultaneous claims be resolved in an order independent of the pseudorandom values of the contested contexts (for instance, a fixed response order). Then the $K$ responses generated by single-owner contexts (\cref{alg:solo}) are independent draws from $p(\cdot \mid x)$, so the scheme is $K$-sample distortion-free (\cref{def:ksample}).
\end{theorem}

\begin{proposition}[Dilution versus multiple testing]\label{prop:dilution}
Consider a response with $n$ scored tokens, of which a fraction $\omega$ carry the watermark of the detector's key, and let $\mu > 0$ be the mean excess score of a watermarked token, so that the standardized score is approximately $z = \omega \mu \sqrt{n}$. A single-key detector at level $\alpha$ detects the response when $z \ge z_{\alpha} = \Phi^{-1}(1 - \alpha)$, and a detector that tests $L$ keys with the \v{S}id\'ak correction needs $z \ge z_{\alpha_L}$ with $\alpha_L = 1 - (1 - \alpha)^{1/L}$. In this normal approximation, single-owner contexts detect more responses than stratified keys ($\omega = 1$) whenever
\begin{equation}\label{eq:dilution}
(1 - \omega)\, \mu \sqrt{n} \;<\; z_{\alpha_L} - z_{\alpha}.
\end{equation}
\end{proposition}

At $\alpha = 1\%$ the right-hand side is $0.76$ for $L = 10$. The condition favors a large watermarked fraction, which the anchors provide (\cref{sec:exp-why}).

\begin{proposition}[Edit exposure]\label{prop:exposure}
Let the hash $\eta$ be random and independent of the text, let the windows $[t - u_w, t - \ell_w)$ satisfy $\ell_w \ge h$, let the tokens within each window be distinct, and let the edit lie at least $\max_w u_w$ tokens after the start of the response. Then the expected exposure of the anchored map in Eqs.~\eqref{eq:context} and \eqref{eq:minhash} is at most $h + 1 + 2|W|$ for a substitution, $h + 1 + \sum_w 2(u_w - h)/(u_w - \ell_w)$ for an insertion, and $h + \sum_w 2(u_w - h)/(u_w - \ell_w)$ for a deletion, that is, $7$, $8$, and $7$ for our windows, independently of the response length.
\end{proposition}

\Cref{sec:experiments} tests these predictions.

\begin{table*}[t]
\centering
{\scriptsize{\setlength{\tabcolsep}{4.6pt}
\begin{tabular}{llcccccccc>{\columncolor[HTML]{DCEBFA}}c}
\toprule
 & & \multicolumn{2}{c}{vLLM} & \multicolumn{2}{c}{GumbelSoft} & \multicolumn{1}{c}{Tournament} & \multicolumn{2}{c}{Key pools} & \multicolumn{2}{c}{Request-scoped history} \\
\cmidrule(lr){3-4} \cmidrule(lr){5-6} \cmidrule(lr){7-7} \cmidrule(lr){8-9} \cmidrule(lr){10-11}
Model & Task & Fixed key & Dedup \texttt{all} & $\tau{=}0.5$ & $\tau{=}0.3$ & SynthID-Text & Random & Stratified & Context history & SOLO \\
\midrule
Qwen2.5-7B & GSM8K & \cellcolor[HTML]{FBE0DA} 30.6$^{*}$ & \cellcolor[HTML]{FBE0DA} 29.3$^{*}$ & 33.3$^{\dagger}$ & 32.1$^{\dagger}$ & \cellcolor[HTML]{FBE0DA} 19.3$^{*}$ & 14.3$^{\dagger}$ & 16.2 & 15.6 & \textbf{21.9} \\
 & GSM8K, JSON & \cellcolor[HTML]{FBE0DA} 51.6$^{*}$ & \cellcolor[HTML]{FBE0DA} 51.4$^{*}$ & 53.5$^{\dagger}$ & 52.8$^{\dagger}$ & \cellcolor[HTML]{FBE0DA} 42.1$^{*}$ & 29.1$^{\dagger}$ & 33.1 & 28.6 & \textbf{37.1} \\
 & MATH-500 & \cellcolor[HTML]{FBE0DA} 34.5$^{*}$ & \cellcolor[HTML]{FBE0DA} 34.3$^{*}$ & 35.7$^{\dagger}$ & 35.2$^{\dagger}$ & \cellcolor[HTML]{FBE0DA} 22.7$^{*}$ & 16.0$^{\dagger}$ & 17.5 & 18.3 & \textbf{20.8} \\
 & MATH-500, JSON & \cellcolor[HTML]{FBE0DA} 83.4$^{*}$ & \cellcolor[HTML]{FBE0DA} 83.6$^{*}$ & \cellcolor[HTML]{FBE0DA} 84.9$^{*}$ & \cellcolor[HTML]{FBE0DA} 84.6$^{*}$ & \cellcolor[HTML]{FBE0DA} 74.4$^{*}$ & \cellcolor[HTML]{FBE0DA} 68.4$^{*}$ & 71.3 & 69.7 & \textbf{74.3} \\
 & TriviaQA, CoT & \cellcolor[HTML]{FBE0DA} 15.6$^{*}$ & \cellcolor[HTML]{FBE0DA} 13.6$^{*}$ & \cellcolor[HTML]{FBE0DA} 15.1$^{*}$ & \cellcolor[HTML]{FBE0DA} 14.8$^{*}$ & \cellcolor[HTML]{FBE0DA} 8.3$^{*}$ & 7.6$^{\dagger}$ & 8.6 & 9.0 & \textbf{10.2} \\
 & TriviaQA, JSON & \cellcolor[HTML]{FBE0DA} 20.8$^{*}$ & \cellcolor[HTML]{FBE0DA} 20.7$^{*}$ & \cellcolor[HTML]{FBE0DA} 22.1$^{*}$ & \cellcolor[HTML]{FBE0DA} 21.6$^{*}$ & \cellcolor[HTML]{FBE0DA} 13.9$^{*}$ & \cellcolor[HTML]{FBE0DA} 9.5$^{*}$ & 9.6 & 10.8 & \textbf{13.1} \\
\midrule
Olmo-3-7B & GSM8K & \cellcolor[HTML]{FBE0DA} 44.0$^{*}$ & 46.0$^{\dagger}$ & 45.2$^{\dagger}$ & 44.2$^{\dagger}$ & \cellcolor[HTML]{FBE0DA} 37.5$^{*}$ & 31.0$^{\dagger}$ & 33.3 & 31.5 & \textbf{36.9} \\
 & GSM8K, JSON & \cellcolor[HTML]{FBE0DA} 26.0$^{*}$ & \cellcolor[HTML]{FBE0DA} 26.6$^{*}$ & 27.6$^{\dagger}$ & \cellcolor[HTML]{FBE0DA} 27.3$^{*}$ & \cellcolor[HTML]{FBE0DA} 19.2$^{*}$ & 13.0$^{\dagger}$ & 14.7 & 13.2 & \textbf{15.2} \\
 & MATH-500, JSON & \cellcolor[HTML]{FBE0DA} 57.4$^{*}$ & \cellcolor[HTML]{FBE0DA} 57.2$^{*}$ & \cellcolor[HTML]{FBE0DA} 59.0$^{*}$ & \cellcolor[HTML]{FBE0DA} 58.2$^{*}$ & \cellcolor[HTML]{FBE0DA} 46.2$^{*}$ & \cellcolor[HTML]{FBE0DA} 42.0$^{*}$ & 44.2 & 44.3 & \textbf{47.3} \\
 & TriviaQA, CoT & \cellcolor[HTML]{FBE0DA} 35.0$^{*}$ & \cellcolor[HTML]{FBE0DA} 31.4$^{*}$ & \cellcolor[HTML]{FBE0DA} 37.4$^{*}$ & \cellcolor[HTML]{FBE0DA} 35.7$^{*}$ & \cellcolor[HTML]{FBE0DA} 24.5$^{*}$ & \cellcolor[HTML]{FBE0DA} 27.1$^{*}$ & \textbf{28.7} & 27.4 & 28.1 \\
 & TriviaQA, JSON & \cellcolor[HTML]{FBE0DA} 43.5$^{*}$ & \cellcolor[HTML]{FBE0DA} 43.5$^{*}$ & \cellcolor[HTML]{FBE0DA} 44.7$^{*}$ & \cellcolor[HTML]{FBE0DA} 43.1$^{*}$ & \cellcolor[HTML]{FBE0DA} 29.8$^{*}$ & 25.0$^{\dagger}$ & 27.3 & 29.2 & \textbf{31.2} \\
\midrule
Ministral-3-8B & GSM8K & \cellcolor[HTML]{FBE0DA} 64.4$^{*}$ & \cellcolor[HTML]{FBE0DA} 68.0$^{*}$ & 68.8$^{\dagger}$ & 65.6$^{\dagger}$ & \cellcolor[HTML]{FBE0DA} 57.5$^{*}$ & 47.3$^{\dagger}$ & 48.2 & 48.4 & \textbf{54.3} \\
 & GSM8K, JSON & \cellcolor[HTML]{FBE0DA} 53.8$^{*}$ & \cellcolor[HTML]{FBE0DA} 54.0$^{*}$ & 59.1$^{\dagger}$ & 56.0$^{\dagger}$ & \cellcolor[HTML]{FBE0DA} 50.5$^{*}$ & 37.4$^{\dagger}$ & 38.9 & 39.6 & \textbf{48.0} \\
 & MATH-500, JSON & \cellcolor[HTML]{FBE0DA} 86.8$^{*}$ & \cellcolor[HTML]{FBE0DA} 86.5$^{*}$ & \cellcolor[HTML]{FBE0DA} 88.5$^{*}$ & 87.2$^{\dagger}$ & \cellcolor[HTML]{FBE0DA} 82.4$^{*}$ & \cellcolor[HTML]{FBE0DA} 78.0$^{*}$ & 79.3 & 78.4 & \textbf{82.0} \\
 & TriviaQA, CoT & \cellcolor[HTML]{FBE0DA} 63.2$^{*}$ & \cellcolor[HTML]{FBE0DA} 55.8$^{*}$ & \cellcolor[HTML]{FBE0DA} 65.2$^{*}$ & \cellcolor[HTML]{FBE0DA} 64.2$^{*}$ & \cellcolor[HTML]{FBE0DA} 45.6$^{*}$ & \cellcolor[HTML]{FBE0DA} 49.1$^{*}$ & \textbf{51.4} & 50.1 & 51.2 \\
 & TriviaQA, JSON & \cellcolor[HTML]{FBE0DA} 72.1$^{*}$ & \cellcolor[HTML]{FBE0DA} 72.3$^{*}$ & \cellcolor[HTML]{FBE0DA} 72.6$^{*}$ & \cellcolor[HTML]{FBE0DA} 71.2$^{*}$ & \cellcolor[HTML]{FBE0DA} 52.4$^{*}$ & \cellcolor[HTML]{FBE0DA} 46.8$^{*}$ & 53.4 & 56.1 & \textbf{57.5} \\
\midrule[\heavyrulewidth]
Qwen2.5-14B & GSM8K & \cellcolor[HTML]{FBE0DA} 66.5$^{*}$ & 61.0$^{\dagger}$ & 67.2$^{\dagger}$ & 67.4$^{\dagger}$ & \cellcolor[HTML]{FBE0DA} 50.8$^{*}$ & 39.7 & 41.2 & 38.7 & \textbf{50.1} \\
 & GSM8K, JSON & \cellcolor[HTML]{FBE0DA} 46.5$^{*}$ & \cellcolor[HTML]{FBE0DA} 46.8$^{*}$ & 49.0$^{\dagger}$ & 47.4$^{\dagger}$ & 40.0$^{\dagger}$ & 27.1 & 29.9 & 23.8 & \textbf{32.8} \\
 & MATH-500, JSON & \cellcolor[HTML]{FBE0DA} 86.5$^{*}$ & \cellcolor[HTML]{FBE0DA} 86.5$^{*}$ & \cellcolor[HTML]{FBE0DA} 87.4$^{*}$ & \cellcolor[HTML]{FBE0DA} 86.6$^{*}$ & \cellcolor[HTML]{FBE0DA} 75.6$^{*}$ & 71.2 & 73.3 & 72.5 & \textbf{74.6} \\
 & TriviaQA, CoT & \cellcolor[HTML]{FBE0DA} 13.3$^{*}$ & \cellcolor[HTML]{FBE0DA} 13.0$^{*}$ & \cellcolor[HTML]{FBE0DA} 13.2$^{*}$ & \cellcolor[HTML]{FBE0DA} 13.3$^{*}$ & \cellcolor[HTML]{FBE0DA} 10.2$^{*}$ & 6.2 & 6.4 & 7.8 & \textbf{9.2} \\
 & TriviaQA, JSON & \cellcolor[HTML]{FBE0DA} 18.5$^{*}$ & \cellcolor[HTML]{FBE0DA} 18.5$^{*}$ & \cellcolor[HTML]{FBE0DA} 18.9$^{*}$ & \cellcolor[HTML]{FBE0DA} 18.8$^{*}$ & \cellcolor[HTML]{FBE0DA} 14.5$^{*}$ & 5.8 & 6.7 & 8.2 & \textbf{10.6} \\
\midrule
Ministral-3-14B & GSM8K & \cellcolor[HTML]{FBE0DA} 56.2$^{*}$ & 59.4$^{\dagger}$ & 62.8$^{\dagger}$ & 58.8$^{\dagger}$ & \cellcolor[HTML]{FBE0DA} 52.3$^{*}$ & 40.5$^{\dagger}$ & 42.7 & 40.2 & \textbf{48.5} \\
 & GSM8K, JSON & \cellcolor[HTML]{FBE0DA} 53.3$^{*}$ & 53.5$^{\dagger}$ & 60.2$^{\dagger}$ & 56.4$^{\dagger}$ & \cellcolor[HTML]{FBE0DA} 49.3$^{*}$ & 37.7$^{\dagger}$ & 41.6 & 40.7 & \textbf{49.2} \\
 & MATH-500, JSON & 85.0$^{\dagger}$ & \cellcolor[HTML]{FBE0DA} 84.8$^{*}$ & \cellcolor[HTML]{FBE0DA} 88.1$^{*}$ & \cellcolor[HTML]{FBE0DA} 86.5$^{*}$ & \cellcolor[HTML]{FBE0DA} 79.3$^{*}$ & \cellcolor[HTML]{FBE0DA} 75.7$^{*}$ & 77.8 & 77.0 & \textbf{81.8} \\
 & TriviaQA, CoT & \cellcolor[HTML]{FBE0DA} 62.3$^{*}$ & \cellcolor[HTML]{FBE0DA} 57.1$^{*}$ & \cellcolor[HTML]{FBE0DA} 64.4$^{*}$ & \cellcolor[HTML]{FBE0DA} 62.7$^{*}$ & \cellcolor[HTML]{FBE0DA} 47.7$^{*}$ & \cellcolor[HTML]{FBE0DA} 44.3$^{*}$ & 46.1 & 46.3 & \textbf{51.7} \\
 & TriviaQA, JSON & \cellcolor[HTML]{FBE0DA} 60.1$^{*}$ & \cellcolor[HTML]{FBE0DA} 61.5$^{*}$ & \cellcolor[HTML]{FBE0DA} 60.7$^{*}$ & \cellcolor[HTML]{FBE0DA} 58.7$^{*}$ & \cellcolor[HTML]{FBE0DA} 44.0$^{*}$ & \cellcolor[HTML]{FBE0DA} 37.3$^{*}$ & 42.0 & 45.4 & \textbf{48.3} \\
\bottomrule
\end{tabular}
}}
\caption{RQ2: clean TPR (\%) at a calibrated 1\% FPR. Coral, $^{*}$: not admissible on the task. $^{\dagger}$: not admissible on another task of the model. Bold: best admissible for the model.}\label{tab:main}
\end{table*}

\section{Experiments}\label{sec:experiments}

\subsection{Experimental Setup}\label{sec:exp-setup}
\noindent\textbf{Models and tasks.}
We use Qwen2.5-7B-Instruct and Qwen2.5-14B-Instruct~\cite{qwen2025qwen25}, Olmo-3-7B-Instruct~\cite{olmo2025olmo3}, and Ministral-3-8B-Instruct and Ministral-3-14B-Instruct~\cite{ministral2026ministral3} on 600 held-out GSM8K problems~\cite{cobbe2021gsm8k}, MATH-500~\cite{hendrycks2021math,lightman2024verify}, and 600 held-out TriviaQA questions~\cite{joshi2017triviaqa}, each in a plain format (``The answer is'', a boxed answer, or the CoT format) and a JSON format with reasoning and answer fields (plain MATH-500 for Qwen2.5-7B only), which gives 26 combinations. Each request draws $K = 10$ responses at temperature 1 without top-$p$ or top-$k$ truncation.

\noindent\textbf{Baselines.}
We compare with watermarks deployable without changing the model and detected from the text with the provider's keys (\cref{tab:props}): the per-response distortion-free samplers of vLLM (fixed key, with and without \texttt{all} deduplication) and SynthID-Text~\cite{dathathri2024scalable}, GumbelSoft~\cite{fu2024gumbelsoft} at $\tau = 0.5$ and $0.3$, random keys (16) and stratified keys (10), and the context history of \citet{hu2024unbiased} reset for every request. All use the four-token contexts of the deployed samplers, within-response deduplication, and one PRF (\cref{app:impl}), so we compare with these configurations, not with key pools over anchored contexts. The choice of two anchor windows was confirmed on the held-out GSM8K items of Qwen2.5-7B, which thus served for model selection.

\begin{table}[t]
\centering
{\scriptsize\setlength{\tabcolsep}{2.4pt}
\begin{tabular}{lcccc}
\toprule
Watermark & \makecell{One\\resp.} & \makecell{$K$\\resp.} & Keys & Dilution \\
\midrule
Fixed key & \cellcolor[HTML]{D9F0F2}\checkmark & \cellcolor[HTML]{FBE0DA}$\times$ & \cellcolor[HTML]{D9F0F2} 1 & \cellcolor[HTML]{D9F0F2} none \\
Dedup \texttt{all} & \cellcolor[HTML]{D9F0F2}\checkmark & \cellcolor[HTML]{FBE0DA}$\times$ & \cellcolor[HTML]{D9F0F2} 1 & \cellcolor[HTML]{D9F0F2} none \\
GumbelSoft & \cellcolor[HTML]{FBE0DA}$\times$ & \cellcolor[HTML]{FBE0DA}$\times$ & \cellcolor[HTML]{D9F0F2} 1 & \cellcolor[HTML]{D9F0F2} none \\
SynthID-Text & \cellcolor[HTML]{D9F0F2}\checkmark & \cellcolor[HTML]{FBE0DA}$\times$ & \cellcolor[HTML]{D9F0F2} 1 & \cellcolor[HTML]{D9F0F2} none \\
Random keys & \cellcolor[HTML]{D9F0F2}\checkmark & \cellcolor[HTML]{FBE0DA}$\times$ & \cellcolor[HTML]{FBE0DA} 16 & \cellcolor[HTML]{D9F0F2} none \\
Stratified keys & \cellcolor[HTML]{D9F0F2}\checkmark & \cellcolor[HTML]{D9F0F2}\checkmark & \cellcolor[HTML]{FBE0DA} 10 & \cellcolor[HTML]{D9F0F2} none \\
Context history & \cellcolor[HTML]{D9F0F2}\checkmark & \cellcolor[HTML]{D9F0F2}\checkmark & \cellcolor[HTML]{D9F0F2} 1 & \cellcolor[HTML]{FBE0DA} yes \\
\midrule
\rowcolor[HTML]{DCEBFA} \textbf{SOLO} & \cellcolor[HTML]{D9F0F2}\checkmark & \cellcolor[HTML]{D9F0F2}\checkmark & \cellcolor[HTML]{D9F0F2}\textbf{1} & \textbf{reduced} \\
\bottomrule
\end{tabular}}
\caption{Watermarks compared: distortion-free per response and per request, keys tested at detection, and dilution by siblings. Teal: desired. Coral: undesired.}\label{tab:props}
\end{table}

\noindent\textbf{Metrics.}
For repeated sampling we report maj@10, the vote-share AUROC, identical responses, coverage (pass@10), and on TriviaQA the AUROC of semantic entropy (SE)~\cite{kuhn2023semantic,farquhar2024detecting}. A watermark is \emph{admissible} on a task if its maj@10 is within 1.5 points of unwatermarked sampling and the 95\% confidence interval (CI) of its AUROC difference includes zero or lies above $-2$ points, and for a model if admissible on all its tasks ($K$-sample distortion-free schemes count as admissible). For detection we report the true positive rate (TPR) at a 1\% FPR, with thresholds calibrated on unwatermarked responses of the same model, task, key, scheme, and edit condition (\cref{app:impl}, unfitted thresholds in \cref{app:stats}), on clean responses and after 10\% random substitutions or insertions and deletions. Numbers average three seeds with fresh keys, and CIs of TPR differences come from a hierarchical bootstrap.

\subsection{RQ2: Faithful and Detectable}\label{sec:exp-main}
\textbf{RQ2} asks whether one watermark keeps repeated sampling intact and still detects each response with a single key.

\noindent\textbf{Repeated sampling.}
As \cref{tab:rs} shows (per task in \cref{app:uq}), the fixed key of vLLM is not admissible for any model, with or without \texttt{all} deduplication, and SynthID-Text also couples TriviaQA responses. GumbelSoft passes on most GSM8K tasks and on plain MATH-500 but misses the rule on TriviaQA with every model at both values of $\tau$, and random keys couple responses that draw the same key (\cref{thm:inflation}). SOLO passes on every task of every model, and stratified keys and the context history pass on 23 and 24 of the 26 combinations, with misses within sampling noise (\cref{app:analysis}). On TriviaQA, maj@10 under the fixed key falls 6.1 points below unwatermarked sampling, whereas SOLO stays within 0.4 points on average (\cref{fig:selfcons} in \cref{app:analysis}). Coverage, which bounds best-of-$n$ selection, falls by up to 21.9 points under the fixed key and changes by at most 1.1 under SOLO.
\begin{table}[t]
\centering
\resizebox{\columnwidth}{!}{\scriptsize{\setlength{\tabcolsep}{2pt}
\begin{tabular}{lrrrrrrr}
\toprule
Watermark & \makecell{Rule\\(of 26)} & \makecell{Adm.\\(of 5)} & \makecell{Ident.\\(\%)} & \makecell{Infl.\\(\%)} & $\Delta$SE & \makecell{$\Delta$maj\\@10} & $\Delta$cov. \\
\midrule
Fixed key & 1 & \cellcolor[HTML]{FBE0DA} 0 & 67.2 & +85.4 & $-$27.8 & $-$6.1 & $-$8.2 \\
Dedup \texttt{all} & 4 & \cellcolor[HTML]{FBE0DA} 0 & 15.5 & +60.8 & $-$17.1 & $-$4.9 & $-$5.7 \\
GumbelSoft ($\tau{=}0.5$) & 11 & \cellcolor[HTML]{FBE0DA} 0 & 3.9 & +14.0 & $-$3.7 & $-$4.2 & $-$1.9 \\
GumbelSoft ($\tau{=}0.3$) & 11 & \cellcolor[HTML]{FBE0DA} 0 & 9.6 & +30.9 & $-$6.7 & $-$4.8 & $-$2.9 \\
SynthID-Text & 1 & \cellcolor[HTML]{FBE0DA} 0 & 27.1 & +43.8 & $-$9.2 & $-$3.7 & $-$4.4 \\
Random keys & 16 & \cellcolor[HTML]{FBE0DA} 1 & 0.9 & +4.8 & $-$0.2 & $-$1.6 & $-$0.6 \\
Stratified keys & 23 & \cellcolor[HTML]{D9F0F2} 5\rlap{\smash{$^{\S}$}} & 0.6 & $-$0.2 & 0.0 & $-$0.7 & 0.0 \\
Context history & 24 & \cellcolor[HTML]{D9F0F2} 5\rlap{\smash{$^{\S}$}} & 0.6 & $-$0.1 & 0.0 & $-$0.3 & +0.1 \\
\midrule
\rowcolor[HTML]{DCEBFA} \bfseries SOLO & \bfseries 26 & \bfseries \cellcolor[HTML]{D9F0F2} 5 & \bfseries 0.5 & \bfseries +0.3 & \bfseries $-$0.2 & \bfseries $-$0.2 & \bfseries +0.2 \\
\bottomrule
\end{tabular}}}
\caption{RQ2: repeated sampling, means over models. Rule: passing combinations. Adm.: admissible models ($^{\S}$: by exactness). Infl.: agreement inflation. SE: AUROC of semantic entropy. $\Delta$: points. Ident.: TriviaQA, CoT. $\Delta$cov.: all tasks. Others: TriviaQA.}\label{tab:rs}
\end{table}

\noindent\textbf{Detection.}
SOLO detects significantly more responses (CI of the difference above zero) than stratified keys, the strongest key pool we evaluate, on 21 of the 26 combinations (\cref{tab:main}, all differences in \cref{app:stats}), than random keys on 24, and than the context history on 20. It significantly beats the best other watermark admissible for the model on 17 of them (13 after a Holm correction), by up to 8.9 points (stratified keys on Qwen2.5-14B, GSM8K). At the nominal Gamma thresholds, fitted to no data, it leads this opponent on 23, 20 and 19 combinations at 1\%, 0.1\% and 0.01\%, though realized FPRs then differ (\cref{app:stats}). The context history itself trails stratified keys in four GSM8K combinations (CIs below zero), which confirms the dilution cost of local contexts.

\par\smallskip\noindent\colorbox[HTML]{E8F1FB}{\parbox{\dimexpr\columnwidth-2\fboxsep\relax}{\small\textbf{Finding 2.} SOLO passes the admissibility rule on all 26 combinations, the other exact schemes on 23 and 24, and at a 1\% FPR it detects significantly more clean responses than stratified keys on 21 and than the context history on 20 of the 26 combinations.}}\par\smallskip

\begin{figure}[t]
\centering
\includegraphics[width=\columnwidth]{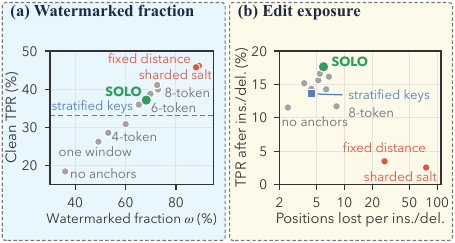}
\caption{RQ3: why SOLO works (Qwen2.5-7B, GSM8K, JSON). Gray, red: other single-owner maps. Red: fragile under insertions and deletions.}\label{fig:why}
\end{figure}

\begin{table}[t]
\centering
{\scriptsize{\setlength{\tabcolsep}{2.6pt}
\begin{tabular}{lrrrrr}
\toprule
 & \multicolumn{5}{c}{GSM8K, JSON} \\
\cmidrule(lr){2-6}
Variant & $\Delta$maj & $\Delta$AUROC & Clean & Subst. & Ins./del. \\
\midrule
\rowcolor[HTML]{DCEBFA} \bfseries SOLO & \bfseries $+0.4$ & \bfseries $-0.2$ & \bfseries 37.1 & \bfseries 14.6 & \bfseries 17.6 \\
w/o anchors & $+0.7$ & $-2.4$ & 18.4 & 10.5 & 11.5 \\
w/o single owner & \cellcolor[HTML]{FBE0DA} $-4.2$ & \cellcolor[HTML]{FBE0DA} $-11.4$ & 55.7 & 24.9 & 28.6 \\
one window $[t-32, t-2)$ & $+0.8$ & $-0.5$ & 26.2 & 12.4 & 15.2 \\
\midrule
$h = 1$ & $+0.7$ & $-1.0$ & 30.8 & 13.5 & 15.6 \\
$h = 3$ & $+0.4$ & $-2.0$ & 41.1 & 13.0 & 16.1 \\
windows $\times 0.6$ & $+0.7$ & $-1.3$ & 38.7 & 15.3 & 17.7 \\
windows $\times 1.5$ & $+0.3$ & $-1.5$ & 35.9 & 14.1 & 16.6 \\
\midrule
local context, $h = 4$ & $+0.6$ & $-1.6$ & 28.6 & 12.9 & 14.3 \\
local context, $h = 6$ & $+0.6$ & $+0.8$ & 36.1 & 12.8 & 14.3 \\
local context, $h = 8$ & $+0.1$ & $+0.5$ & 39.9 & 10.6 & 11.7 \\
\midrule
\rowcolor[HTML]{ECECEC} stratified keys (ref.) & $+0.3$ & $-0.4$ & 33.1 & 11.7 & 13.6 \\
\bottomrule
\end{tabular}}}
\caption{RQ3: ablation (Qwen2.5-7B, GSM8K, JSON). $\Delta$: points against unwatermarked sampling. TPR (\%) without and after 10\% edits. Coral: not admissible.}\label{tab:ablation}
\end{table}

\subsection{RQ3: Why SOLO Works}\label{sec:exp-why}
\textbf{RQ3} asks which component of SOLO drives its detection and whether its lead survives token edits.

\noindent\textbf{Mechanism.}
Replaying the claims of each single-owner context map gives its exact watermarked fraction $\omega$ (\cref{fig:why}, more models in \cref{app:models}). Detection grows with $\omega$ (Spearman 0.99 over 12 maps): two- and four-token contexts watermark 36\% and 53\% of the positions, SOLO 68\%, and single-owner maps overtake stratified keys between $\omega$ = 60\% and 65\%, the trade-off of \cref{prop:dilution}. Anchors at fixed distances and sharded prefix salts reach larger fractions, but one insertion or deletion strips 27.0 and 76.0 positions on average, and their TPR after 10\% insertions and deletions falls to 3.5\% and 2.6\%. Content-defined anchors lose 5.9 positions per insertion or deletion, consistent with \cref{prop:exposure} and close to the 4.4 of four-token contexts, so SOLO raises $\omega$ at the robustness of short contexts.

\noindent\textbf{Ablation.}\label{sec:exp-ablation}
\Cref{tab:ablation} removes one component at a time (two more models in \cref{app:models}). \emph{Without anchors}, siblings compete for short contexts and the TPR drops, as \cref{prop:dilution} predicts. \emph{Without single ownership}, the TPR rises but 15\% of the JSON requests return ten identical responses, so the variant is not admissible. One anchor window also lowers the TPR. The window sizes are not critical: scaling both by $0.6$ or $1.5$ changes the TPR by at most 3 points.

\begin{figure}[t]
\centering
\includegraphics[width=\columnwidth]{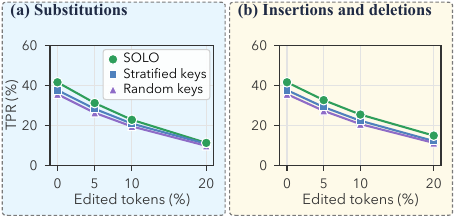}
\caption{RQ3: TPR at a calibrated 1\% FPR after random edits, mean over 26 combinations.}\label{fig:robust}
\end{figure}

\noindent\textbf{Robustness.}\label{sec:exp-robust}
SOLO detects significantly more responses than stratified keys on 16 of the 26 combinations after 10\% random substitutions and on 21 after 10\% random insertions and deletions. The edits replace or insert random token ids, which are valid tokens for the two Qwen models and partly fall outside the vocabularies of Olmo-3 and Ministral-3. On the 11 Qwen combinations alone, SOLO leads on 9 and 10 of them and is behind on none (\cref{app:impl,app:analysis}). After insertions and deletions it is significantly behind neither key pool in any combination at any edit rate, and ahead of stratified keys on 18 to 21 of them (\cref{fig:robust}). At a 20\% edit rate it still detects 14.9\% of the responses on average, against 12.2\% for stratified keys. Its lead over a fixed ten-key pool also grows when a request has fewer responses, holds on average at every response length we test, and adds little generation and detection cost (\cref{app:analysis,app:eff}).

\par\smallskip\noindent\colorbox[HTML]{E8F1FB}{\parbox{\dimexpr\columnwidth-2\fboxsep\relax}{\small\textbf{Finding 3.} Content-defined anchors raise the watermarked fraction at the edit exposure of short local contexts, and each component of SOLO is needed. Its lead over stratified keys mostly survives random edits: it is significantly ahead on 16 of the 26 combinations after substitutions (behind on 1) and on 21 after insertions and deletions.}}\par\smallskip

\section{Conclusion}\label{sec:conclusion}
Deployed distortion-free LLM watermarks couple the responses of a request at template funnels, which undermines self-consistency and uncertainty estimation, and existing exact remedies lose detection power, key pools to multiple testing, local histories to dilution. SOLO keeps a request-scoped context history over content-anchored contexts: its responses are as diverse as unwatermarked ones, each is detected with a single key, and at a 1\% FPR it detects significantly more clean responses than stratified keys over four-token contexts on 21 of 26 model and task combinations, with a significant lead on most of them after random edits. Extending single ownership across requests, keying the anchor hash against adversarial edits, and evaluating execution-based selection in code generation are natural next steps.

\section*{Limitations}\label{sec:limitations}
Our experiments use open instruction-tuned models with up to 14 billion parameters on three reasoning and question-answering benchmarks, with requests of up to ten responses at temperature one, the regime of self-consistency and sampling-based uncertainty estimation. Further applications of repeated sampling, such as code generation with execution-based selection, and larger models remain to be evaluated. 

\section*{Ethical Considerations}\label{sec:ethics}
\noindent\textbf{Data and tools.}
We evaluate on public benchmarks. GSM8K is released under the MIT license. MATH-500 is a subset of MATH selected by \citet{lightman2024verify}, and its source repositories (MATH and PRM800K) are released under the MIT license. The dataset card of TriviaQA states no license. The models (Qwen2.5-7B-Instruct, Qwen2.5-14B-Instruct, Olmo-3-7B-Instruct, Ministral-3-8B-Instruct, and Ministral-3-14B-Instruct) and the software we build on (vLLM, Hugging Face Transformers, and Math-Verify) are released under the Apache 2.0 license. We use all of them for research evaluation. The benchmarks consist of math problems and trivia questions. We generate model responses to them, collect no new human data, and involve no annotators or participants.

\noindent\textbf{Compute.}
All experiments ran on NVIDIA H100 GPUs, and detection runs on CPUs (\cref{app:impl}).

\noindent\textbf{Intended use and risks.}
SOLO is meant for providers who watermark generated text for provenance while applications draw several responses per prompt. A detection is statistical evidence that a text came from a watermarked model at a chosen false positive rate, not proof of authorship, so decisions about people should not rest on it alone. The guarantee of \cref{thm:exact} covers the responses of one request, and responses from separate requests share the provider key, as with every single-key watermark.

\clearpage
\appendix
\makeatletter
\setlength{\@fptop}{0pt}\setlength{\@fpsep}{12pt plus 2pt}
\setlength{\@dblfptop}{0pt}\setlength{\@dblfpsep}{12pt plus 2pt}
\renewcommand{\dbltopfraction}{0.9}
\makeatother
\raggedbottom
\crefalias{section}{appendix}\crefalias{subsection}{subappendix}

\section*{Appendix Overview}\label{app:guide}
The appendices follow the order of the main text. \Cref{app:related} extends the related work (\cref{sec:related}). \Cref{app:revisit} adds the revisiting results with a second model (\cref{sec:revisiting}). \Cref{app:method} gives method details (\cref{sec:method}). \Cref{app:proofs,app:theorychecks} prove and numerically check the results of \cref{sec:revisiting,sec:theory}. \Cref{app:impl,app:analysis,app:uq,app:se,app:models,app:eff} support \cref{sec:experiments}. \Cref{tab:evidence} maps each claim to the comparison that decides it.

\begin{table*}[!htbp]
\centering
{\footnotesize\setlength{\tabcolsep}{4pt}
\begin{tabular}{p{0.27\textwidth}p{0.33\textwidth}p{0.17\textwidth}p{0.14\textwidth}}
\toprule
\textbf{Claim} & \textbf{Decisive comparison} & \textbf{Scope} & \textbf{Evidence} \\
\midrule
\rowcolor[HTML]{ECE8F7} \multicolumn{4}{l}{\emph{RQ1: revisiting watermarks under repeated sampling}} \\
Deployed fixed-key watermarks couple the responses of a request at template funnels & Identical responses, maj@10, and vote-share AUROC against unwatermarked sampling, plus agreement inflation against marker length & vLLM 0.30 with two models on TriviaQA and GSM8K & \cref{sec:revisiting-funnel}, \cref{tab:revisit,fig:funnel}, \cref{app:revisit,app:models} \\
Inexact remedies leave part of the coupling & $\kappa_S$ of \cref{thm:inflation}, a toy simulation, and dual-key routing on real requests & Toy model and vLLM 0.30 & \cref{sec:revisiting-remedies}, \cref{tab:revisit}, \cref{app:revisit,app:proofs,app:theorychecks} \\
Exact remedies lose detection power & TPR of stratified keys against the fixed key, and TPR of the context history against stratified keys & Held-out GSM8K & \cref{sec:revisiting-keys,sec:revisiting-history}, \cref{tab:main} \\
\rowcolor[HTML]{ECE8F7} \multicolumn{4}{l}{\emph{RQ2: faithful and detectable}} \\
SOLO is $K$-sample distortion-free & \Cref{thm:exact} and the admissibility rule on every task & 26 combinations & \cref{sec:theory,sec:exp-main}, \cref{tab:rs}, \cref{app:uq,app:se} \\
SOLO detects more responses than four-token key pools and the context history & Paired TPR differences with hierarchical bootstrap CIs, Holm correction & 26 combinations & \cref{sec:exp-main}, \cref{tab:main} \\
\rowcolor[HTML]{ECE8F7} \multicolumn{4}{l}{\emph{RQ3: why SOLO works}} \\
Anchors raise $\omega$ at the robustness of short contexts & TPR against $\omega$, and exposure per insertion or deletion & GSM8K JSON with three models & \cref{sec:exp-why}, \cref{fig:why}, \cref{app:analysis,app:models} \\
Each component is needed & Removing anchors, single ownership, or one window & GSM8K with three models & \cref{sec:exp-ablation}, \cref{tab:ablation,tab:ablation-full,tab:ablation-b4}, \cref{app:analysis,app:models} \\
The lead holds under edits, for fewer responses, and at every length we test, at low cost & Edit rates of 5\% to 20\%, $K \in \{4, 10\}$, truncated responses, and timing & 26 combinations and GSM8K & \cref{sec:exp-robust}, \cref{fig:robust}, \cref{app:analysis,app:eff} \\
\bottomrule
\end{tabular}}
\caption{Evidence map: each claim of the paper, the comparison that decides it, its scope, and where the evidence is.}\label{tab:evidence}
\end{table*}

\section{Extended Related Work}\label{app:related}
This appendix supports \cref{sec:related} with the full discussion of prior work.

\subsection{Generation-Time Watermarks for Language Models}\label{app:related-wm}
Generation-time watermarks embed a signal while the model samples, so that the provider can later test a text for the signal with a secret key. One family biases the next-token distribution toward a key-dependent subset of the vocabulary~\cite{kirchenbauer2023watermark,zhao2024provable,kirchenbauer2024reliability}. This family is simple and robust but changes the output distribution. A second family keeps the distribution of each response unchanged by replacing the true randomness of sampling with keyed pseudorandomness, through the Gumbel-max decoder~\cite{aaronson2023watermarking,fernandez2023threebricks}, inverse transform sampling over a key sequence~\cite{kuditipudi2024robust}, reweighting~\cite{hu2024unbiased,wu2024dipmark}, tournament sampling~\cite{dathathri2024scalable}, cryptographically undetectable sampling seeded by a high-entropy prefix~\cite{christ2024undetectable}, or pseudorandom codes~\cite{christ2024prc,golowich2024edit}. Statistical analyses of the resulting detectors are given in \citet{li2025statistical}. Distortion-free schemes are the ones deployed at scale: SynthID-Text is used in a production assistant~\cite{dathathri2024scalable}, and the vLLM serving engine~\cite{kwon2023vllm} watermarks every request of a watermarked deployment with Gumbel-max sampling unless the request opts out~\cite{vllm2026watermarking}. Most of these schemes derive the keyed randomness from the last few generated tokens and deduplicate repeated contexts within a response, which is the design we revisit. Robustness evaluations show that such watermarks survive moderate token edits but not aggressive paraphrasing of short texts~\cite{krishna2023paraphrasing,sadasivan2023can,piet2023markmywords}, although the diluted signal remains detectable in long texts~\cite{kirchenbauer2024reliability}. Several of these works define guarantees over multiple generations, exactly through a history of contexts shared across generations~\cite{hu2024unbiased,dathathri2024scalable}, or approximately or computationally through fresh randomness for every query~\cite{kuditipudi2024robust,christ2024undetectable,christ2024prc}, but the deployed configurations deduplicate only within a response, and none of these works analyzes the coupling that templates create between the responses of a request or the detection cost of removing it.

\subsection{Keys, Context Histories, and Output Diversity}
A few works note that keyed sampling interacts with diversity. \citet{wu2024collision} showed that key collisions, which are inevitable with a limited supply of keys, correlate pseudorandom samples and bias distortion-free watermarks, and introduced the $\beta$-watermark family to reduce this bias. GumbelSoft~\cite{fu2024gumbelsoft} observed that the Gumbel-max watermark is deterministic given the prompt and replaced the arg max with sampling from a softmax, trading the per-response guarantee for diversity. \citet{verma2026watermarking} met the same problem when selecting among several Gumbel-watermarked generations with a reward model and adopted a similar relaxation. Key pools assign different keys to different generations, to hide the watermark from users~\cite{liu2025waterbag} or to resist forgery~\cite{aremu2026randomized}. They restore diversity at the cost of testing several keys. \citet{kuditipudi2024robust} shift their key sequence at random for every query, or keep state so that no element of the key sequence is reused, which keeps repeated queries approximately distortion-free at the cost of a detector that searches over the shifts. \citet{hu2024unbiased} keep a history of the context codes used by earlier generations and skip the watermark at a code already used. When reset after every batch of $b$ generations, the history makes the batch exactly distortion-free ($b$-shot undetectable). SynthID-Text defines the same mechanism as repeated context masking over the last $K$ responses, but configures it for a single response because stronger forms of non-distortion reduce detectability~\cite{dathathri2024scalable}. \citet{gloaguen2025blackbox} show that the low diversity of fixed-key watermarks under repeated queries and the caches of context histories let a user detect the presence of a watermark from black-box queries. Our analysis explains why the inexact remedies only partially work under repeated sampling, since template funnels re-synchronize contexts after samples diverge, and why the exact remedy, a request-scoped context history with the usual local contexts, can detect fewer responses than stratified keys on GSM8K. SOLO supplies the context map that makes such a history detection-efficient.

\subsection{Repeated Sampling and Uncertainty Estimation}
Many reliability tools draw several responses for one prompt. Self-consistency votes over sampled reasoning paths~\cite{wang2023selfconsistency}, and repeated sampling with selection scales test-time compute~\cite{brown2024monkeys,snell2025scaling}. Sampling-based uncertainty estimators measure the disagreement among responses, through semantic clustering~\cite{kuhn2023semantic,farquhar2024detecting}, consistency checks for hallucination detection~\cite{manakul2023selfcheckgpt}, prompt perturbations~\cite{gao2024spuq}, or consistency combined with confidence~\cite{vashurin2025cocoa}. All of them assume independent responses. We show when deployed watermarks violate this assumption and how to respect it while detecting, on most tasks, more clean responses than the key pools and context histories that also respect it.

\section{Additional Revisiting Results}\label{app:revisit}
This appendix supports \cref{sec:revisiting}: the coupling of deployed watermarks at template funnels is not specific to one model, and the inexact remedies leave part of it.

\noindent\textbf{Pseudorandom function of vLLM.}
Assumption (A1) is an idealization, which cryptographic PRFs such as HMAC-SHA256 are designed to approximate for anyone without the key, whereas the Philox generator used by vLLM is a statistical, non-cryptographic generator~\cite{vllm2026watermarking}. Under (A1) to (A3), the fixed-key, deduplicated, dual-key, multi-key, and context-history schemes we study, and ours, are per-response distortion-free, and a context history shared by the $K$ responses achieves \cref{def:ksample} for every $K$~\cite{hu2024unbiased}.

\noindent\textbf{Settings.}
Each request asks for $K = 10$ responses through parallel sampling with vLLM's \texttt{n} $= 10$. On TriviaQA we also use Ministral-3-8B-Instruct~\cite{ministral2026ministral3}, with the same settings as Qwen2.5-7B-Instruct.

\noindent\textbf{Fixed-key samplers with Ministral-3-8B-Instruct.}
With Ministral-3-8B-Instruct (same settings, not in \cref{tab:revisit}), 99.6\% of the questions return ten identical answers under the default configuration (5\% without the watermark) and the AUROC falls from 87\% to 50\%. The \texttt{all} mode restores it on the plain format (86\% against 87\%) but not with JSON (59\% of the requests collapse, AUROC 61\% instead of 87\%), and the nonce reaches an AUROC of 59\%. The vLLM documentation remarks that parallel samples can be identical while their contexts coincide~\cite{vllm2026watermarking}. \Cref{prop:funnel}(b) shows that the coupling outlasts the first tokens.

\begin{figure*}[!t]
\begin{minipage}[t]{\columnwidth}\vspace{0pt}
\centering
\includegraphics[width=\columnwidth]{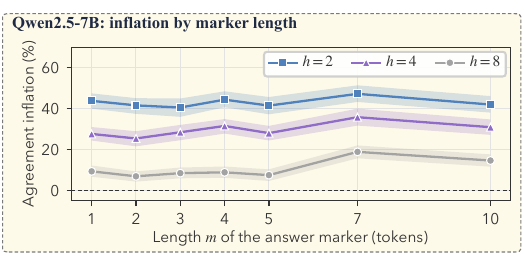}
\caption{Template funnels: agreement inflation among ten TriviaQA responses (CoT) over unwatermarked sampling against the answer-marker length $m$, for Gumbel watermarks of context width $h$ in vLLM's \texttt{all} mode (our sampler, Qwen2.5-7B-Instruct, 300 questions, two seeds, 95\% bootstrap bands).}\label{fig:funnel}
\end{minipage}\hfill
\begin{minipage}[t]{\columnwidth}\vspace{0pt}
\centering
\includegraphics[width=\columnwidth]{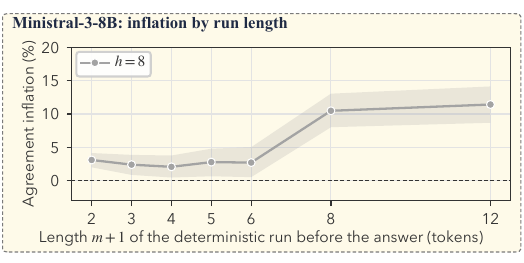}
\caption{Template funnels with Ministral-3-8B-Instruct, as in \cref{fig:funnel}, for $h = 8$ with the first $h$ tokens sampled with true randomness (our sampler, 300 questions, two seeds, 95\% bootstrap band), against the length $L_f = m + 1$ of the deterministic run before the answer.}\label{fig:funnel-ministral}
\end{minipage}
\end{figure*}

\noindent\textbf{Funnels in the \texttt{all} mode.}
In \cref{fig:funnel}, agreement is inflated for every context width even with a single-token answer marker (44\%, 28\%, and 9\% for $h = 2$, $4$, and $8$), because the responses state their answer already in the reasoning sentence, right after phrases that they share, which act as funnels themselves. The marker starts a new line, and the tokenizer merges the end of the reasoning sentence and the line break into one token, so a marker of $m$ tokens gives a deterministic run of $L_f = m + 1$ tokens before the answer. For $h = 8$ it fills the context from $m = 7$ on, and the inflation roughly doubles for these two markers (19\% and 15\% for $m = 7$ and $10$), as \cref{prop:funnel}(b) predicts. With Qwen2.5-7B-Instruct, the step for $h = 4$ is small relative to the coupling already present. Ministral-3-8B-Instruct shows the same rise for $h = 8$ (\cref{app:models}). Longer contexts make shared runs rarer but do not remove them.

\noindent\textbf{Inexact remedies.}
\Cref{thm:inflation} models funnels at which both responses face the same answer distribution, as for short answers. When the reasoning of each response already fixes its answer, \cref{prop:funnel}(c) bounds the coupling instead. On a toy model with a six-token funnel, the measured inflation matches Eq.~\eqref{eq:inflation} for every scheme (82\% and 51\% for $\lambda = 0.1$ and $0.5$, and 25\% and 6\% for pools of 4 and 16 keys, see \cref{app:theorychecks}). With Ministral-3-8B-Instruct (not included in \cref{tab:revisit}), dual-key routing at the default $\lambda$ reduces the AUROC of TriviaQA requests in JSON format from 87\% to 79\%.

\noindent\textbf{Exact context histories.}
In \cref{tab:main}, the context history detects fewer responses than stratified keys on GSM8K with Qwen2.5-14B in both formats, Qwen2.5-7B in JSON format, and Ministral-3-14B in plain format (by 2.5 to 6.0 points, with confidence intervals below zero), where the siblings share long reasoning phrases. The dilution of a request-scoped history over local contexts led SynthID-Text to configure its context masking for a single response~\cite{dathathri2024scalable}. Per-model repeated-sampling metrics of stratified keys and the fixed key are given in \cref{app:uq}.

\section{Method Details}\label{app:method}
\begin{figure*}[!htbp]
\centering
\includegraphics[width=\textwidth]{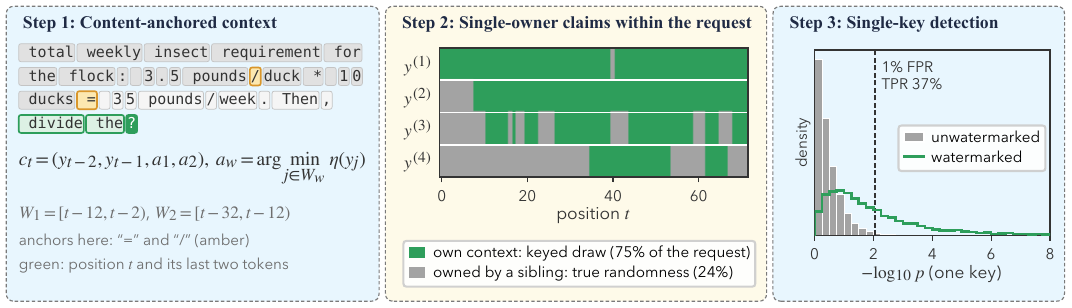}
\caption{SOLO on one real request, as in \cref{fig:pipeline} (Qwen2.5-7B-Instruct, GSM8K, JSON, the median-detection item among items whose ten responses all have at least 60 tokens). Step 1: the keyed context of position $t$ is its last two tokens and the min-hash tokens of $W_1$ and $W_2$. Step 2: the first 72 positions of four of the ten responses. Green: the response owns the context and draws with the key. Gray: a sibling owns it and the position uses true randomness. Step 3: each response is detected with one key ($p$-values of all 6000 watermarked responses of this run and of unwatermarked ones).}\label{fig:pipeline-real}
\end{figure*}
This appendix supports \cref{sec:method}: the relation of single-owner contexts to prior context histories, the keyed anchor hash for deployment, and the cost of SOLO.

\noindent\textbf{Single-owner contexts and prior histories.}
Single-owner contexts are the context history of \citet{hu2024unbiased} scoped to one request: their history of used context codes, reset after every batch, makes the batch exactly distortion-free, and SynthID-Text defines the same masking over the last $K$ responses but deploys it for a single response~\cite{dathathri2024scalable}, as vLLM does~\cite{vllm2026watermarking}. \Cref{thm:exact} restates the guarantee for our setting: the scheme is $K$-sample distortion-free for every context map. A natural alternative is to keep one key per response, either drawn at random from a pool~\cite{liu2025waterbag,aremu2026randomized}, which only lowers the collision probability to $1/L$ (\cref{thm:inflation}), or assigned without repetition inside the request, which removes collisions. In both cases the detector has to test every key of the pool, and the multiple-testing correction lowers its power (\cref{sec:revisiting-keys}). The increase of $\omega$ by the anchors reported in \cref{sec:method-anchor} is measured by replaying the claims on the generated tokens, as in \cref{sec:exp-why}. \Cref{fig:pipeline-real} shows SOLO on a real request. The request is the item with the median detection rate among items whose ten responses all have at least 60 tokens, and its panel shows the first 72 positions of four of the ten responses. Positions that are neither keyed nor owned by a sibling repeat a context of the same response.

\noindent\textbf{Anchor hash.}
$\eta$ is an affine bijection of the token id modulo $2^{32}$ (\cref{app:impl}). In our experiments $\eta$ is a fixed public bijection. An adversary who knows it can insert or substitute a token with a small hash that becomes the minimum of every window that contains it, which exposes up to $h + 1 + \sum_w (u_w - \ell_w) = 33$ positions with one edit. A deployment should therefore key the hash with the provider key, for instance $\eta_k(v) = E_k(v)$ for a pseudorandom permutation $E_k$ of $\{0, \ldots, 2^{32} - 1\}$ keyed with $k$ (a 32-bit block cipher applied to the token id), which keeps $\eta$ injective and leaves \cref{thm:exact,prop:exposure} unchanged for edits chosen without the key. \citet{kirchenbauer2024reliability} seed the pseudorandom function of a single response with the minimum hash over its last few tokens to make it robust. We use the same min-hash selection over windows further back to make contexts specific across siblings without making them fragile.

\noindent\textbf{Complexity.}
Each decoding step adds one context hash and one lookup and insertion in $\mathcal{U}$ per response. The window minima can be maintained in $O(1)$ amortized time with a monotone queue (our reference implementation recomputes them, at most 30 hash evaluations per position), so generation costs $O(K T)$ extra hash operations and $O(K T)$ memory for a request of $K$ responses of length $T$. Detection costs $O(T)$ hash evaluations per response, the same as for the plain Gumbel watermark, and its threshold is calibrated as for every scheme (\cref{sec:exp-setup}). Measured costs are given in \cref{app:eff}.

\noindent\textbf{Relation to prior schemes.}
SOLO combines the exact request-scoped history of \citet{hu2024unbiased} and \citet{dathathri2024scalable} with a context map designed for it. Compared with key pools~\cite{liu2025waterbag,aremu2026randomized}, it tests one key. Compared with a history over local contexts of four tokens, it watermarks a larger fraction $\omega$ of the scored positions on all 26 model and task combinations we evaluate, by 7 to 17 points (\cref{sec:method-anchor}).

\section{Proofs}\label{app:proofs}\FloatBarrier

This appendix supports \cref{sec:revisiting,sec:theory}: it proves \cref{prop:funnel,thm:inflation,thm:exact,prop:dilution,prop:exposure}. We first summarize the proof ideas of the three results of \cref{sec:theory}.

\noindent\textbf{Exactness (\cref{thm:exact}).} Draw the pseudorandom values lazily, the first time a key and context pair is evaluated. Because no context value is evaluated twice within a request, every evaluation returns a fresh uniform vector that is independent of everything generated before. By (A2), each token then follows the model's conditional distribution, and the joint law factorizes. The claim order only decides which response receives the watermark at a shared context, never whether a value is reused. \cref{thm:exact} holds for any context map, including the anchored map of Eq.~\eqref{eq:context}. The guarantee concerns the responses of one request.

\noindent\textbf{Dilution versus multiple testing (\cref{prop:dilution}).} At $\alpha = 1\%$ the right-hand side of Eq.~\eqref{eq:dilution} is $0.76$ for $L = 10$ and $0.90$ for $L = 16$, while at $\alpha = 0.1\%$ it is $0.63$ and $0.75$. The condition favors a large watermarked fraction, which the anchors provide: on held-out GSM8K in JSON format with Qwen2.5-7B-Instruct they raise $\omega$ from 53\% for single-owner contexts of four tokens to 68\% (\cref{sec:exp-why}). Eq.~\eqref{eq:dilution} also shows that, for a fixed $\omega < 1$, the advantage is confined to weakly watermarked responses, those with $\mu \sqrt{n} < (z_{\alpha_L} - z_{\alpha})/(1 - \omega)$ (about $2.4$ for $\omega = 68\%$, $L = 10$, and $\alpha = 1\%$, and $2.0$ at $\alpha = 0.1\%$). Real responses differ in their signal $\mu \sqrt{n}$, and the positions that siblings contest are likely to be mostly shared phrases with little entropy, which would carry little signal anyway (we do not measure this directly). The proposition, with $\omega$ counted over positions, therefore gives the direction of the trade-off rather than its size.

\noindent\textbf{Edit exposure (\cref{prop:exposure}).} For comparison, the exposure of a local $h$-token context is $h + 1$ (substitution, insertion) or $h$ (deletion). That of anchors at fixed distances $D$ is $h + 1 + |D|$ for a substitution, $\max(h, \max D) + 1$ for an insertion, and $\max(h, \max D)$ for a deletion. An edit inside a salted prefix changes every later context. A position loses its pair only if its local part contains the edit or one of its anchors changes. An anchor changes only if one token enters and another leaves its window and one of the two is the window minimum, which has probability at most $2/(u_w - \ell_w)$ for distinct tokens. A substitution changes the windows of $u_w - \ell_w$ positions in this way. An insertion or deletion also shifts by one token the windows of the $\ell_w - h$ positions after the edit whose window lies entirely before it, which gives the bounds. With repeated tokens in a window, the minimum is shared and the probabilities grow with the inverse number of distinct tokens. The bounds assume a hash that the editor does not know. With a public hash, a targeted edit can expose up to $h + 1 + \sum_w (u_w - \ell_w)$ positions (\cref{app:method}).

Throughout, (A1) to (A3) of Section~\ref{sec:formulation} hold, $\pi$ denotes a next-token distribution, and $\Gamma$ is the Gumbel-max decoder $\Gamma(\pi, \xi) = \arg\max_v \xi_v^{1/\pi_v}$, which satisfies (A2): for independent uniforms $\xi_v$, $\xi_v^{1/\pi_v}$ has distribution function $u^{\pi_v}$ on $(0,1)$, so the arg max equals $v$ with probability $\pi_v / \sum_{v'} \pi_{v'} = \pi_v$.

\subsection{A Coupling Lemma}

\begin{lemma}[Collision coupling]\label{lem:coupling}
Consider any scheme that, at every step of every response, either evaluates $F_{k}(c)$ for some key and context and applies $\Gamma$, or samples from $\pi$ with true randomness. Let $C$ be the event that some pair $(k, c)$ is evaluated at two keyed positions of the request (in the same or in different responses). Then the total variation distance between the joint law of the $K$ responses and $p(\cdot \mid x)^{\otimes K}$ is at most $\mathbb{P}(C)$.
\end{lemma}

\begin{proof}
We construct both processes on one probability space. Fix the order in which the positions of the request are generated (step by step, responses in a fixed order within a step, or in any order determined by the history before the step). In the watermarked process, draw $F_{k}(c)$ lazily: the first time a pair is evaluated, draw a fresh uniform vector $U$ and record it. Later evaluations of the same pair read the record. By (A1) this produces the same joint law as a fixed random function. In the reference process, every keyed position instead uses the fresh vector $U$ drawn at that position, and every unkeyed position uses the same true randomness as in the watermarked process. The two processes produce identical outputs until the first time a recorded pair is evaluated again, that is, on the complement of $C$. In the reference process every keyed token is $\Gamma(\pi, U)$ with a fresh $U$, hence distributed as $\pi$ by (A2), conditionally on everything before, and every unkeyed token is distributed as $\pi$ as well. Therefore the reference process generates $K$ independent responses from $p(\cdot \mid x)$, and the coupling inequality gives the bound.
\end{proof}

\subsection{Proof of Proposition~\ref{prop:funnel}}
(a) With response-only contexts and a fixed key, the first context of every response is the all-padding context, so step 1 of every response uses the same vector $\xi$ and the same distribution $\pi = p(\cdot \mid x)$. By the definition of $\Gamma$ all responses select the same token. Suppose that all responses agree up to step $t-1$ and that the context at step $t$ has not occurred earlier in the response. Then the responses share this context (the last $h$ tokens agree and are padded identically) and the distribution $p(\cdot \mid x, y_{<t})$, so they select the same token again. The induction stops only at the first position whose context repeats within the response, where (A3) switches to true randomness, and up to that position the responses are identical. Short answers rarely contain a repeated context, which is why the shipped watermark returns ten identical short answers (\cref{tab:revisit}).

(b) Let the decision of response $i$ be preceded by a run $r = (r_1, \ldots, r_{L_f})$ of $L_f \ge h$ tokens that every response generates deterministically, that is, with $p(r_j \mid x, \cdot) = 1$ at each of these steps. The context of the decision step is then $(r_{L_f-h+1}, \ldots, r_{L_f})$ in every response, regardless of the tokens before the run. A prompt-side randomization changes $x$, which does not enter response-only contexts, and an unwatermarked prefix of $N$ tokens that ends before the run does not change the run. Hence two responses with the same key evaluate $F$ on the same pair at their decision steps whenever the run is shared and the decision context is new within each response, which is the hypothesis of part (b). In the notation of \cref{thm:inflation}, $\rho_F$ is the probability that the decision context is new in both responses, which is one up to the probability that the last $h$ tokens of the run already occurred earlier in a response.

(c) Let the two responses face decision distributions $\pi$ and $\pi'$ with the same $\xi$. Gumbel-max sampling with shared randomness couples the two decisions. Writing $g_v = -\log(-\log \xi_v)$, the decisions are
\[
\arg\max_v \,(\log \pi_v + g_v) \ \text{and}\ \arg\max_v \,(\log \pi'_v + g_v),
\]
and the probability that the two arg maxes coincide is at least $(1 - \mathrm{TV}(\pi, \pi'))/(1 + \mathrm{TV}(\pi, \pi'))$ by the analysis of Gumbel-max sampling as a coupling without communication~\cite{daliri2024coupling}. For $\pi = \pi'$ the two decisions are equal with probability one, whereas independent sampling gives agreement probability $\sum_a \pi_a^2$.

\subsection{Proof of Theorem~\ref{thm:inflation}}
Under the hypotheses of \cref{thm:inflation}, in which the answer is the token chosen at the decision step and the collision event is independent of the key assignment and of the routing, two responses $i \neq j$ reach decision contexts that collide with probability $\rho_F$, and, given the joint history of both responses before their decision steps, each decision follows the answer distribution $\pi$. Condition on the collision, and not yet on the history. Because the collision is independent of the key assignment and of the routing, both responses then evaluate the PRF on the same pair with probability $\kappa_S$, and in that case they return the same answer, because $\Gamma(\pi, \xi)$ is a deterministic function of the distribution and the pseudorandom value (for Gumbel-max this is the case $\pi = \pi'$ of (c) above). Otherwise, by the conditional-independence hypothesis, the two decisions are independent draws from $\pi$ given the joint history, so they agree with probability $\sum_a \pi_a^2$. Without a collision, the same hypothesis gives agreement probability $\sum_a \pi_a^2$. The hypothesis is needed: distinct decision contexts alone do not make the decisions independent, because the decision context of one response may equal a context that the other response evaluated at an earlier step, in which case both read the same pseudorandom value, and the two histories may be correlated through values they shared earlier. Under (A1) the hypothesis holds, for instance, when neither decision pair was evaluated earlier by the other response and the decision distributions do not depend on earlier shared values, as for a short answer that follows a deterministic template run. Hence
\[
\mathbb{P}(A^{(i)} = A^{(j)}) = \sum_a \pi_a^2 + \rho_F \kappa_S \Bigl(1 - \sum_a \pi_a^2\Bigr).
\]
The values of $\kappa_S$ follow from the key assignment. Fixed key: both responses use the same key, $\kappa_S = 1$. Dual-key routing: each response independently uses key A with probability $1 - \lambda$ and key B with probability $\lambda$ at the decision, so they share a key with probability $\lambda^2 + (1 - \lambda)^2$. Random keys from a pool of $L$: they share a key with probability $1/L$. Stratified keys with $K \le L$: distinct keys, $\kappa_S = 0$. Context history shared by the request: only the first claimant evaluates the PRF on the pair, $\kappa_S = 0$. For the vote share $\hat v_a = K^{-1} \sum_i \mathbf{1}[A^{(i)} = a]$ of a fixed answer $a$ among $K$ exchangeable responses, apply the same conditioning to every pair $i \neq j$ (the hypotheses of \cref{thm:inflation} are required for every pair). Then $\mathbb{P}(A^{(i)} = A^{(j)} = a) = \pi_a^2 + \rho_F \kappa_S\, \pi_a (1 - \pi_a)$, so the indicators $\mathbf{1}[A^{(i)} = a]$, each with mean $\pi_a$, have pairwise correlation $\rho_F \kappa_S$, and $\mathrm{Var}(\hat v_a) = K^{-1}\pi_a(1 - \pi_a)\bigl(1 + (K - 1) \rho_F \kappa_S\bigr)$, the variance of independent responses inflated by the design effect $1 + (K - 1) \rho_F \kappa_S$. Summing over $a$ also gives $\mathbb{E} \sum_a \hat v_a^2 = K^{-1} + (1 - K^{-1}) \mathbb{P}(A^{(i)} = A^{(j)})$, so the expected Gini diversity $1 - \mathbb{E} \sum_a \hat v_a^2$ falls by $(1 - K^{-1}) \rho_F \kappa_S (1 - \sum_a \pi_a^2)$.

\subsection{Proof of Theorem~\ref{thm:exact}}
The argument is that of the context code history of \citet{hu2024unbiased} (their Corollary~15, batch variant), restated for one request. In \cref{alg:solo}, a response evaluates $F_{k}(c)$ only if $c \notin \mathcal{U}$ and then inserts $c$ into $\mathcal{U}$ before any later position is generated. Since all responses use the single key $k$, no pair $(k, c)$ is evaluated twice within the request, so the event $C$ of \cref{lem:coupling} is empty for every context map and for every order of resolving simultaneous claims that does not depend on the pseudorandom values of the contested contexts. \cref{lem:coupling} then gives total variation distance zero. The same argument applies to stratified keys with $K \le L$: the responses of a request use distinct keys and each deduplicates its own contexts, so again no pair $(k, c)$ is evaluated twice and $C$ is empty.

\subsection{Proof of Proposition~\ref{prop:dilution}}
Under the null hypothesis, the per-token scores $-\log(1 - \xi_{y_t})$ are independent standard exponential variables, so $S$ has mean $n$ and variance $n$. Under the watermark, a watermarked token has score mean $1 + \mu$ and an unwatermarked token has mean $1$. In the normal approximation the standardized score is $z = (S - n)/\sqrt{n} \approx \omega n \mu / \sqrt{n} = \omega \mu \sqrt{n}$, and the level-$\alpha$ one-key test rejects when $z \ge z_\alpha$. The \v{S}id\'ak-corrected test over $L$ keys rejects when the smallest of the $L$ $p$-values is at most $\alpha_L = 1 - (1 - \alpha)^{1/L}$. The true key dominates the minimum when the response is watermarked, so the test rejects when $z \ge z_{\alpha_L}$ with $\omega = 1$. Comparing $\omega \mu \sqrt{n} - z_{\alpha}$ with $\mu \sqrt{n} - z_{\alpha_L}$ gives Eq.~\eqref{eq:dilution}.

\subsection{Proof of Proposition~\ref{prop:exposure}}
A scored position $t$ of the edited text keeps its pair only if its local tokens, its anchors, and its own token are those of a generation-time position. We compare it with the generation-time position of the same token. \emph{Local part.} A substitution at index $j$ changes the token at $j$ and the local contexts of positions $j+1, \ldots, j+h$, that is, $h+1$ positions. An insertion adds one new position and changes the local contexts of the $h$ following positions, that is, $h+1$ positions. A deletion changes the local contexts of the $h$ following positions. \emph{Anchors.} Fix a window $w$ with offsets $[t - u, t - \ell)$, $\ell \ge h$, and width $m = u - \ell$. Whenever the content of the window of a position differs from that of its generation-time counterpart by one token that leaves and one token that enters, the anchor changes only if the leaving token was the minimum of the original window or the entering token is smaller than that minimum. For a random hash and distinct window tokens, each event has probability at most $1/m$. \emph{Substitution at $j$:} the windows that contain $j$ are those of the $m$ positions $t \in (j + \ell, j + u]$, and each loses the old and gains the new token, so the expected number of changed anchors is at most $2$. \emph{Insertion before original index $j$:} for the $m$ positions whose window contains the inserted token, the window gains it and loses its farthest token. For the $\ell - h$ positions after the edit and outside the local part whose window lies entirely before the edit, the window is shifted by one token relative to the counterpart, gaining its nearest and losing its farthest token. Windows that lie entirely after the edit are unchanged. This gives at most $2(m + \ell - h)/m = 2(u - h)/(u - \ell)$ changed anchors in expectation. \emph{Deletion at $j$:} in the same way, the window of each of the $u - h$ positions that follow the local part and precede index $j + u$ differs from its counterpart by one token that leaves and one that enters, which gives the same bound $2(u - h)/(u - \ell)$. Summing over the $|W|$ windows and adding the local part by a union bound gives the stated bounds. For our windows $(u, \ell) \in \{(12, 2), (32, 12)\}$ and $h = 2$, they equal $3 + 2 + 2 = 7$, $3 + 2 + 3 = 8$, and $2 + 2 + 3 = 7$. If the edit lies within the first $\max_w u_w$ tokens, windows are truncated at the start of the response and each changes with a larger probability. With repeated tokens in a window, the minimum is shared and the probabilities grow with the inverse number of distinct tokens. The bounds assume a hash that the editor does not know (\cref{sec:method-minhash}). For anchors at fixed distances $D$, every position $t$ with $t - d \le j < t$ for some $d \in D$ reads a different token at distance $d$ after an insertion or deletion, which gives $\max(h, \max D)$ positions, plus the inserted position for an insertion. For a salted prefix, every later context contains the salt.

\section{Numerical Checks of the Theory}\label{app:theorychecks}
This appendix supports \cref{sec:revisiting,sec:theory}: simulations and measured exposures agree with \cref{prop:funnel,thm:inflation,thm:exact,prop:exposure}.

\noindent\textbf{Toy model.}
To check \cref{prop:funnel,thm:inflation,thm:exact} without the noise of a real model, we simulate requests of $K = 10$ responses on a toy language model with a vocabulary of $V = 64$ tokens. Each response consists of a free prefix of 12 tokens drawn uniformly at random, a deterministic template run of $L_f$ tokens shared by all responses, and one decision token drawn from $\pi = (0.5, 0.3, 0.2)$. The pseudorandom function is a keyed cryptographic hash, and the requests use disjoint keys. We measure the pairwise agreement of the decision tokens and report the inflation $(\mathbb{P}(A^{(i)} = A^{(j)}) - \sum_a \pi_a^2)/(1 - \sum_a \pi_a^2)$ over 3000 requests, which \cref{thm:inflation} predicts to be $\rho_F \kappa_S$. \cref{tab:theorysim} shows that every measured value lies within three standard errors of the prediction, except a deficit of about 0.6 points for $h = 2$ and $L_f \ge h$. There, a two-token decision context already occurs in the random prefix with probability about $11/V^2$, and within-response deduplication (A3) then samples that decision with true randomness, so that the prediction is about 99.5\% rather than 100\%. Otherwise, a fixed key with response-only contexts couples the whole response. Once the prefix diverges, the coupling appears only when the run is at least as long as the context ($L_f \ge h$). vLLM's \texttt{all} deduplication does not remove a later funnel. Dual-key routing and random keys leave the predicted fractions of the coupling, and stratified keys, the context history, and SOLO remove it.

\begin{table*}[!t]
\begin{minipage}[t]{0.47\textwidth}\vspace{0pt}
\centering
{\scriptsize\setlength{\tabcolsep}{3pt}%
\begin{tabular}{lcccc}
\toprule
Scheme & $L_f$ & $h$ & Measured (\%) & Predicted (\%) \\
\midrule
Fixed key, response only & 0 & 4 & \cellcolor[HTML]{FBE0DA} +100.0 $\pm$ 0.0 & 100.0 \\
Fixed key, diverged prefix & 0 & 2 & -0.2 $\pm$ 0.3 & 0.0 \\
Fixed key, diverged prefix & 1 & 2 & +2.0 $\pm$ 0.3 & 1.6 \\
Fixed key, diverged prefix & 2 & 2 & \cellcolor[HTML]{FBE0DA} +99.4 $\pm$ 0.1 & 100.0 \\
Fixed key, diverged prefix & 4 & 2 & \cellcolor[HTML]{FBE0DA} +99.4 $\pm$ 0.1 & 100.0 \\
Fixed key, diverged prefix & 0 & 4 & +0.7 $\pm$ 0.3 & 0.0 \\
Fixed key, diverged prefix & 1 & 4 & +0.3 $\pm$ 0.3 & 0.0 \\
Fixed key, diverged prefix & 3 & 4 & +1.2 $\pm$ 0.3 & 1.6 \\
Fixed key, diverged prefix & 4 & 4 & \cellcolor[HTML]{FBE0DA} +100.0 $\pm$ 0.0 & 100.0 \\
Fixed key, diverged prefix & 6 & 4 & \cellcolor[HTML]{FBE0DA} +100.0 $\pm$ 0.0 & 100.0 \\
Fixed key, dedup \texttt{all} & 6 & 4 & \cellcolor[HTML]{FBE0DA} +100.0 $\pm$ 0.0 & 100.0 \\
Dual-key, $\lambda = 0.1$ & 6 & 4 & \cellcolor[HTML]{FBE0DA} +82.4 $\pm$ 0.4 & 82.0 \\
Dual-key, $\lambda = 0.5$ & 6 & 4 & \cellcolor[HTML]{FBE0DA} +50.9 $\pm$ 0.7 & 50.0 \\
Random keys, $L = 4$ & 6 & 4 & \cellcolor[HTML]{FBE0DA} +25.0 $\pm$ 0.6 & 25.0 \\
Random keys, $L = 16$ & 6 & 4 & +6.0 $\pm$ 0.4 & 6.2 \\
Stratified keys, $L = 10$ & 6 & 4 & +0.2 $\pm$ 0.3 & 0.0 \\
Context history & 6 & 4 & -0.1 $\pm$ 0.3 & 0.0 \\
\rowcolor[HTML]{DCEBFA} \bfseries SOLO & \bfseries 6 & \bfseries 2 & \bfseries -0.3 $\pm$ 0.3 & \bfseries 0.0 \\
\rowcolor[HTML]{ECECEC} Unwatermarked & 6 & 4 & +0.3 $\pm$ 0.3 & 0.0 \\
\bottomrule
\end{tabular}}
\caption{Agreement inflation on the toy model (mean $\pm$ standard error) and the value predicted by \cref{prop:funnel,thm:inflation,thm:exact}. Coral: measured inflation above 10\%. Gray: unwatermarked.}\label{tab:theorysim}
\end{minipage}\hfill
\begin{minipage}[t]{0.51\textwidth}\vspace{0pt}
\centering
{\scriptsize\setlength{\tabcolsep}{3pt}%
\begin{tabular}{lcccccc}
\toprule
 & \multicolumn{3}{c}{GSM8K} & \multicolumn{3}{c}{GSM8K, JSON} \\
Context map & Subst. & Ins. & Del. & Subst. & Ins. & Del. \\
\midrule
\rowcolor[HTML]{ECE8F7} \multicolumn{7}{l}{\emph{Qwen2.5-7B-Instruct}} \\
4-token context & 5.0 & 5.0 & 3.9 & 4.9 & 5.0 & 3.9 \\
\rowcolor[HTML]{DCEBFA} \bfseries SOLO (min-hash anchors) & \bfseries 5.8 & \bfseries 6.6 & \bfseries 5.4 & \bfseries 6.4 & \bfseries 6.9 & \bfseries 5.0 \\
one min-hash window & 4.2 & 4.3 & 3.0 & 4.5 & 4.4 & 2.9 \\
fixed-distance anchors & 5.7 & \cellcolor[HTML]{FBE0DA} 29.9 & \cellcolor[HTML]{FBE0DA} 28.1 & 5.6 & \cellcolor[HTML]{FBE0DA} 28.1 & \cellcolor[HTML]{FBE0DA} 25.9 \\
sharded prefix salt & 5.9 & \cellcolor[HTML]{FBE0DA} 119.4 & \cellcolor[HTML]{FBE0DA} 113.9 & 5.8 & \cellcolor[HTML]{FBE0DA} 78.3 & \cellcolor[HTML]{FBE0DA} 73.7 \\
prefix salt & \cellcolor[HTML]{FBE0DA} 20.6 & \cellcolor[HTML]{FBE0DA} 16.5 & \cellcolor[HTML]{FBE0DA} 20.2 & \cellcolor[HTML]{FBE0DA} 17.5 & \cellcolor[HTML]{FBE0DA} 18.3 & \cellcolor[HTML]{FBE0DA} 19.6 \\
\midrule
\rowcolor[HTML]{ECE8F7} \multicolumn{7}{l}{\emph{Qwen2.5-14B-Instruct}} \\
4-token context & 5.0 & 5.0 & 3.9 & 4.9 & 5.0 & 3.9 \\
\rowcolor[HTML]{DCEBFA} \bfseries SOLO (min-hash anchors) & \bfseries 6.2 & \bfseries 7.0 & \bfseries 5.4 & \bfseries 6.5 & \bfseries 6.7 & \bfseries 5.4 \\
one min-hash window & 4.5 & 4.2 & 3.0 & 4.5 & 4.3 & 2.9 \\
fixed-distance anchors & 5.8 & \cellcolor[HTML]{FBE0DA} 30.2 & \cellcolor[HTML]{FBE0DA} 28.6 & 5.6 & \cellcolor[HTML]{FBE0DA} 28.0 & \cellcolor[HTML]{FBE0DA} 26.0 \\
sharded prefix salt & 5.9 & \cellcolor[HTML]{FBE0DA} 130.1 & \cellcolor[HTML]{FBE0DA} 126.0 & 5.8 & \cellcolor[HTML]{FBE0DA} 77.7 & \cellcolor[HTML]{FBE0DA} 73.6 \\
prefix salt & \cellcolor[HTML]{FBE0DA} 20.4 & \cellcolor[HTML]{FBE0DA} 17.8 & \cellcolor[HTML]{FBE0DA} 20.9 & \cellcolor[HTML]{FBE0DA} 18.1 & \cellcolor[HTML]{FBE0DA} 18.1 & \cellcolor[HTML]{FBE0DA} 19.6 \\
\bottomrule
\end{tabular}}
\caption{Mean number of positions that lose their watermark after one random edit (600 unwatermarked held-out GSM8K responses of one seed per model and format). Coral: ten or more positions.}\label{tab:exposure}
\end{minipage}
\end{table*}

\noindent\textbf{Edit exposure on real responses.}
To check \cref{prop:exposure}, we apply a single random edit to each of 600 unwatermarked held-out responses of Qwen2.5-7B-Instruct and of Qwen2.5-14B-Instruct and count the scored positions whose (keyed context, token) pair no longer occurs in the original response. \cref{tab:exposure} reports the mean exposure. Local contexts of four tokens lose $h + 1 = 5$ positions per substitution or insertion and $h = 4$ per deletion. The anchored context of SOLO stays below its bounds (7 for a substitution or a deletion and 8 for an insertion) with both models. Fixed-distance anchors lose $\max D + 1 = 33$ positions per insertion and $\max D = 32$ per deletion in the median (fewer on average, since edits near the end of a response affect fewer positions). Prefix salts lose every later position whenever the edit falls into the prefix. The prefix salt keys every position $t \ge 16$ with its last four tokens and a hash of the first 16 response tokens. The sharded prefix salt keys position $t \ge 16$ with the prefix token at index $t \bmod 16$, so that a substitution in the prefix affects only one sixteenth of the later positions, whereas an insertion or deletion anywhere shifts $t \bmod 16$ and changes the salt of every later position. Edits within the first $\max_w u_w = 32$ tokens, where the anchor windows are truncated, expose more positions than the bounds, which assume an edit away from the start. The means include such edits.

\section{Implementation Details}\label{app:impl}
This appendix supports \cref{sec:revisiting,sec:exp-setup}: it describes the shared sampler, the vLLM runs, the baselines, calibration, answer extraction, edits, and compute.

\noindent\textbf{Shared sampler.}
All watermarks except those of the vLLM experiments (described below) run in one keyed sampler implemented as a logits processor for Hugging Face Transformers. At every decoding step it receives the next-token probabilities of all responses of a batch, chooses each token according to the scheme, and forces the model to emit it. All schemes share one pseudorandom function: a context value and a key are hashed into a 32-bit seed with a 32-bit integer mixing function, and the seed and the token index are hashed again into a 64-bit uniform value, so that every scheme draws its randomness in the same way. The 32-bit key and seeds suffice for the statistical evaluation but not for deployment, since the key can be recovered by exhaustive search from one watermarked response. A deployed scheme should derive $\xi$ from a cryptographic PRF with a key of at least 128 bits (e.g., HMAC-SHA256), and the anchor hash of SOLO should be keyed in the same way (\cref{sec:method-minhash}). Responses of one request are decoded in the same batch, which allows request-level bookkeeping. The anchor hash of SOLO is the multiplicative hash $\eta(v) = (2654435761\, v + 2135587861) \bmod 2^{32}$ of the token id, a bijection on the vocabulary that we model as a random hash in \cref{prop:exposure}. An empty window at the start of a response yields a padding anchor. Generation stops at the end of the model's turn or after 160 (TriviaQA), 400 (GSM8K), or 1024 (MATH-500) new tokens, with the same budgets for every model. Truncated responses are scored as they are. The share of responses that reach the budget depends on the model and the task (\cref{tab:trunc}). It is largest for Ministral-3-8B on MATH-500 in JSON format (26.4\% of the unwatermarked responses), and for every watermark it differs from that of unwatermarked sampling with the same model and task by at most 4.4 points, so that it affects all watermarks alike.

\begin{table}[!htbp]
\centering
{\footnotesize\setlength{\tabcolsep}{3pt}
\resizebox{\columnwidth}{!}{\begin{tabular}{lcccccc}
\toprule
Model & \makecell{GSM8K} & \makecell{GSM8K,\\JSON} & \makecell{MATH-500} & \makecell{MATH-500,\\JSON} & \makecell{TriviaQA,\\CoT} & \makecell{TriviaQA,\\JSON} \\
\midrule
Qwen2.5-7B & 4.6 & 0.4 & 8.4 & 0.9 & 0.1 & 0.0 \\
Olmo-3-7B & 17.6 & 1.5 & -- & 4.4 & 3.1 & 1.9 \\
Ministral-3-8B & 12.8 & 5.6 & -- & \cellcolor[HTML]{FFF0B8} 26.4 & 2.7 & 3.2 \\
\midrule[\heavyrulewidth]
Qwen2.5-14B & 6.6 & 0.3 & -- & 0.2 & 0.6 & 0.0 \\
Ministral-3-14B & 12.4 & 5.0 & -- & 24.3 & 2.0 & 2.7 \\
\bottomrule
\end{tabular}}}
\caption{Share (\%) of unwatermarked responses that reach the generation budget (160, 400, and 1024 new tokens for TriviaQA, GSM8K, and MATH-500), mean over three seeds. --: plain-format MATH-500 is evaluated with Qwen2.5-7B-Instruct only. Yellow: largest share.}\label{tab:trunc}
\end{table}

\noindent\textbf{Models and prompts.}
Every prompt is the model's chat template applied to one user turn, with the template's default system prompt where the template has one. For Ministral-3-8B-Instruct and Ministral-3-14B-Instruct~\cite{ministral2026ministral3}, we use the BF16 weights of version 2512. They are vision-language checkpoints, which Transformers loads as \texttt{Mistral3ForConditionalGeneration}. We feed them text only, so that only their language model runs. Their tokenizer is loaded with the Hugging Face \texttt{tokenizers} backend and the corrected pre-tokenizer regular expression (the \texttt{fix\_mistral\_regex} option of Transformers). Every prompt carries exactly one beginning-of-sequence token, and a turn ends with \texttt{</s>}. The default system prompt of their template contains date placeholders such as \texttt{\{today\}}, which the Hugging Face template leaves unfilled, so the system prompt is the same in every run and for every scheme.

\noindent\textbf{vLLM experiments.}
\Cref{tab:revisit} and the Ministral-3-8B-Instruct numbers on vLLM's shipped watermarks in \cref{app:revisit} use vLLM 0.30 and its built-in watermarking without modification, with the key $42 + 1000\,s$ in the run with seed $s$. With Ministral-3-8B-Instruct, vLLM's code for the Pixtral vision encoder of this checkpoint imports two names from Transformers that the installed Transformers 5.18 has renamed or removed. A compatibility shim, loaded only for these runs, makes \texttt{PixtralRotaryEmbedding} an alias of its new name \texttt{PixtralVisionRotaryEmbedding}, which has the same constructor, and restores \texttt{position\_ids\_in\_meshgrid} as implemented in earlier Transformers versions. Both belong to the vision encoder, which text-only requests never run. The prompts are encoded with the same tokenizer as in our sampler and passed to vLLM as token ids, the responses are decoded with the same tokenizer, and image inputs are disabled (\texttt{limit\_mm\_per\_prompt} set to zero images).

\noindent\textbf{Baselines.}
\emph{Fixed key} uses the Gumbel-max decoder with a context of the last four response tokens and within-response deduplication, which follows vLLM's default configuration (positions before the response are padding). \emph{Fixed key with dedup all} additionally samples the first four response tokens with true randomness. vLLM's \texttt{all} mode also skips contexts that occur in the prompt, and \cref{tab:revisit} reports the shipped implementation itself. \emph{GumbelSoft} samples from the softmax of the logits plus the keyed Gumbel noise at temperature $\tau = 0.5$, the most diverse value of the sensitivity study of its paper, and also at $\tau = 0.3$, the value of its main experiments. It is detected with the $z$-test on the Gumbel values of its paper. \emph{SynthID-Text} implements the vectorized non-distortionary tournament of \citet{dathathri2024scalable} with depth 30, a context of four tokens, and the unweighted mean $g$-value score, which its authors report to be weaker than their weighted-mean and Bayesian scores, so its TPRs are lower bounds (SynthID-Text is not admissible on every task of any model, so this does not affect the comparison). The 30 layers use 30 bits of one keyed hash per token. \emph{Random keys} draws one of 16 keys per response, and \emph{stratified keys} gives response $i$ key $i \bmod 10$. Both are detected with the minimum $p$-value over their keys and the \v{S}id\'ak correction. The \emph{context history} of \citet{hu2024unbiased} is implemented in the Gumbel-max form of their paper with one key (the masking of SynthID-Text is evaluated only in this Gumbel-max form): within a request, a response skips the watermark at any four-token context that it or a sibling has already used, and the history is reset for every request. It is detected with the single-key Gamma test over first occurrences, like SOLO. Dual-key routing (Section~\ref{sec:revisiting}) routes each token to the second key with probability $\lambda$ using true randomness and is detected with the fused score of TextSeal~\cite{textseal2026} with a Gamma moment-matching null, without the entropy weighting and localization that require the model at detection time.

\noindent\textbf{Calibration.}
For every model, task, seed, scheme, and edit condition, the detection threshold is the 1\% quantile of the scheme's own $p$-values on the unwatermarked responses of the same model, task, and seed under the same edit, computed over the 6000 (GSM8K, TriviaQA) or 5000 (MATH-500) unwatermarked responses. We calibrate because the nominal Gamma null of the Gumbel score is exact only on average over keys: for one fixed key, unwatermarked responses that pass through the same template evaluate the same keyed contexts there and share a key-dependent shift of their scores. On the held-out GSM8K runs of Qwen2.5-7B-Instruct, the realized false positive rate of the Gamma null at the nominal 1\% level ranges from 0.28\% to 6.68\% across schemes with a Gamma null (including the ablation variants), formats, and keys (0.42\% to 1.10\% for SOLO). Over all tasks and keys of Qwen2.5-7B, Olmo-3-7B and Qwen2.5-14B, the per-key rate reaches 6.83\% for SOLO and 2.40\% for stratified keys. GumbelSoft and SynthID-Text use normal approximations, which are calibrated in the same way. Calibrating per task and edit condition puts all schemes on the same footing. The edit-rate sweep of \cref{fig:robust} and the length analysis of \cref{app:analysis} use the first 300 requests of each run.

\noindent\textbf{Answers and metrics.}
Answers are extracted from the final ``The answer is'' statement or else the last number of the response (GSM8K), the last boxed expression (MATH-500), the ``Final answer:'' line (TriviaQA), or the \texttt{answer} field of the JSON object. GSM8K answers are compared as numbers. MATH-500 answers are grouped into equivalence classes with the Math-Verify library and compared with the reference by the same library. TriviaQA answers are normalized and count as correct if they equal a reference alias or contain an alias of more than three characters. All responses without an extractable answer form one class that is never correct and takes part in the vote like any other class. The vote share is the relative frequency of the majority class among the ten responses.

\noindent\textbf{Edits.}
Substitutions replace each token independently with probability 10\% by a token id drawn uniformly from $[0, 150000)$. Insertions and deletions delete each token with probability 5\% and insert such an id before it with probability 5\%. The edits act on the token-id sequence and are not decoded and re-tokenized. For Qwen2.5-7B-Instruct and Qwen2.5-14B-Instruct, whose special tokens start at id 151643, every drawn id is an ordinary vocabulary token. Olmo-3 (100278 ids) and Ministral-3 (131072 ids) have smaller vocabularies, so about 33\% and 13\% of the drawn ids lie outside them. Such ids act as random tokens for every detector, since all detectors only hash token ids, but they do not correspond to text an editor could produce. Results on the Qwen combinations therefore measure edits with valid tokens (\cref{app:analysis}). The same edit stream is applied to every scheme and to the unwatermarked responses used for calibration.

\noindent\textbf{Compute.}
All experiments ran on NVIDIA H100 GPUs. Generating one configuration of 600 items with ten responses each takes 11 to 15 minutes for the 7B models with our sampler. Detection runs on CPUs.

\section{Additional Experimental Details and Analyses}\label{app:analysis}
This appendix supports \cref{sec:experiments}: it gives the rationale of the setup, the full repeated-sampling discussion, the mechanism and ablation analyses, and the results on the number of responses, edit rates, and response length that \cref{sec:exp-robust} summarizes.

\noindent\textbf{Setup.}
Held-out items share no question with the development items used to design the method: GSM8K problems come from the test split, TriviaQA questions from the validation split, and MATH-500 uses all 500 problems. The choice between one and two anchor windows was confirmed on the held-out GSM8K items of Qwen2.5-7B-Instruct (\cref{sec:exp-ablation}), and the method was frozen before any other model or task was run. We compare with generation-time watermarks that a provider can deploy without changing the model and that are detected from the text with the provider's keys, which is the setting of SOLO. Among them, we evaluate samplers that are distortion-free for each response, the family deployed in production and in vLLM, together with GumbelSoft as the closest relaxation of the Gumbel-max sampler. GumbelSoft runs at $\tau = 0.5$, the most diverse temperature of its sensitivity study, and at $\tau = 0.3$, the value of its main experiments. Stratified keys are the strongest key pool we evaluate, and the context history of \citet{hu2024unbiased} uses the Gumbel-max decoder and is reset for every request, the exact remedy of \cref{sec:revisiting-history}. Stratified keys and the context history are $K$-sample distortion-free. All schemes use contexts of four response tokens unless stated otherwise, within-response deduplication, and the same pseudorandom function. We do not evaluate KGW-type samplers~\cite{kirchenbauer2023watermark}, which bias every response by design, or the random-shift variant of \citet{kuditipudi2024robust}, whose detector searches over the shifts of its key sequence, a form of key multiplicity represented here by random and stratified keys.

\noindent\textbf{Metrics.}
The coverage (pass@10), the fraction of requests with at least one correct response, bounds best-of-$n$ selection. Schemes that are $K$-sample distortion-free (the context history and SOLO by \cref{thm:exact}, stratified keys with $K \le L$ by the same argument, \cref{app:proofs}) count as admissible regardless, because the rule can only reject them by sampling noise. As the rule accepts any difference whose confidence interval includes zero, we also report the pairwise-agreement inflation of \cref{thm:inflation} with its confidence interval. We calibrate because the nominal Gamma null of the Gumbel score is exact only on average over keys: for one fixed key, unwatermarked responses that pass through the same template evaluate the same keyed contexts and share a key-dependent shift of their scores. Calibrating per task and edit condition uses knowledge that a provider may not have, but it puts all schemes on the same footing. Confidence intervals of TPR differences come from a hierarchical bootstrap that resamples the three seeds (keys) and, within each, the items, and re-estimates every calibrated threshold in each replicate, while those of the repeated-sampling metrics (AUROC differences, agreement inflation, semantic entropy) come from a paired bootstrap over items with the keys fixed. The intervals are not adjusted for multiple comparisons, but the count of significant leads over the best other admissible watermark is also reported after a one-sided Holm correction (\cref{app:stats}).

\begin{figure*}[!t]
\begin{minipage}[t]{\columnwidth}\vspace{0pt}
\centering
\includegraphics[width=\columnwidth]{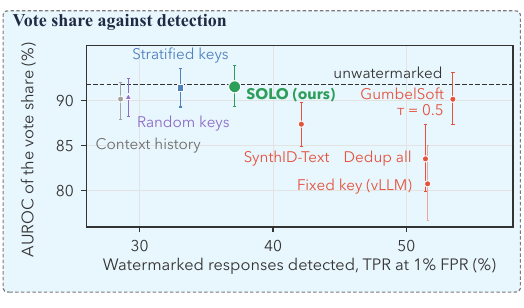}
\caption{Repeated sampling and detection (Qwen2.5-7B-Instruct, $K = 10$, held-out GSM8K, JSON format, three seeds). Vertical: AUROC of the vote share for majority-answer correctness (bars: 95\% CI of the difference to unwatermarked sampling, dashed). Horizontal: TPR at a calibrated 1\% FPR. Red and purple (random keys): not admissible on some task of the model. Green: SOLO. GumbelSoft: $\tau = 0.5$.}\label{fig:teaser}
\end{minipage}\hfill
\begin{minipage}[t]{\columnwidth}\vspace{0pt}
\centering
\includegraphics[width=\columnwidth]{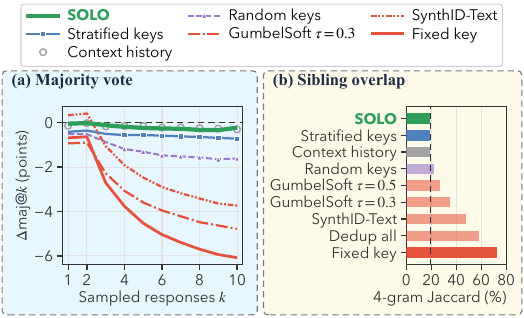}
\caption{Repeated sampling under the watermarks (three seeds, $K = 10$). (a) Difference of maj@$k$ to unwatermarked sampling on TriviaQA (CoT and JSON formats, mean over the five models), for $k$ responses drawn at random from the ten of a request (GumbelSoft: $\tau = 0.3$). (b) Mean pairwise 4-gram Jaccard similarity between the ten responses of a request over the 26 combinations. Black dashed: unwatermarked sampling. Green: SOLO. Blue: stratified keys. Gray: context history. Red and purple (random keys): not admissible for at least one model.}\label{fig:selfcons}
\end{minipage}
\end{figure*}

\noindent\textbf{Repeated sampling against detection.}
\Cref{fig:teaser} places every watermark in the plane of repeated sampling and detection for one setting. Fixed keys detect many responses but lower the AUROC of the vote share, stratified keys and the request-scoped context history keep the AUROC but detect fewer responses, and SOLO keeps the AUROC while detecting more.

\noindent\textbf{Repeated sampling.}
SynthID-Text, whose tournament also reuses keyed values at shared contexts, couples the responses as well on TriviaQA. At $\tau = 0.5$, GumbelSoft passes the admissibility rule on GSM8K with every model and on plain MATH-500, where the reasoning makes most responses diverge before their answers, and at $\tau = 0.3$ it passes on the same tasks except GSM8K in JSON format with Olmo-3-7B, and also on MATH-500 in JSON format with Ministral-3-8B. It still raises the agreement between the answers on some of these tasks, with a confidence interval above zero: at $\tau = 0.5$ only with Olmo-3-7B (GSM8K in both formats) and Ministral-3-14B (GSM8K in JSON format), and at $\tau = 0.3$ with every model (on at least one GSM8K format). On TriviaQA, GumbelSoft misses the rule in both formats at both temperatures with every model and couples the answers there, significantly raising their agreement and lowering the semantic-entropy AUROC on every TriviaQA task except the CoT format of Ministral-3-14B at $\tau = 0.5$. Its lower temperature couples them more with every model. Random keys couple two responses that draw the same key, as \cref{thm:inflation} predicts. SOLO passes the rule on every task of every model. Stratified keys and the context history pass it on 23 and 24 of the 26 combinations. Where they miss it, maj@10 is at most 2.2 points and the AUROC at most 4 points below one unwatermarked run, which is sampling noise. None of the three exact schemes changes the semantic-entropy AUROC significantly.

\noindent\textbf{Self-consistency and diversity.}
\Cref{fig:selfcons} shows what this coupling costs an application that draws more responses. On TriviaQA, unwatermarked sampling gains 5.5 points of majority-vote accuracy from one to ten responses on average over the five models. Under the fixed key, whose single responses are distortion-free, maj@10 falls 6.1 points below unwatermarked sampling, whereas SOLO stays within 0.4 points of unwatermarked sampling on average for every $k$. Its sibling responses also overlap as little as unwatermarked ones: their mean pairwise 4-gram Jaccard similarity is 19\% (unwatermarked sampling: 19\%), against 73\% under the fixed key.

\noindent\textbf{Why SOLO works: watermarked fraction.}
Detection grows with $\omega$ (\cref{fig:why}a): over the 12 single-owner context maps, the Spearman correlation between $\omega$ and the clean TPR is 0.99 with Qwen2.5-7B-Instruct, 0.99 with Qwen2.5-14B-Instruct, and 0.97 with Ministral-3-8B-Instruct. Contexts of the last two and four tokens watermark $\omega$ = 36\% and 53\% of the positions, and SOLO 68\% (JSON clean TPRs in \cref{tab:ablation}). The other two models keep this order (\cref{app:models}). With Qwen2.5-7B-Instruct, stratified keys watermark every scored position but correct for ten keys, and single-owner contexts overtake them between $\omega$ = 60\% and $\omega$ = 65\%. This is the trade-off of \cref{prop:dilution}, although for Qwen2.5-7B-Instruct the crossover comes earlier than its homogeneous normal approximation suggests, consistent with the contested positions carrying less signal than an average position.

\noindent\textbf{Why SOLO works: exposure.}
We measure exposure in \cref{fig:why}b as in \cref{tab:exposure}: one random insertion or deletion on each of 600 unwatermarked responses of one seed per model. Anchors at fixed distances and sharded salts derived from the prefix reach the largest watermarked fractions (89\% and 88\%, and 87\% to 90\% and 87\% to 88\% with the three models), but an insertion or deletion shifts their anchors or changes their salt. One such edit removes the watermark from 27.0 and 76.0 positions on average, and their TPR after 10\% insertions and deletions falls to 3.5\% and 2.6\% (\cref{fig:why}b, and 3.2\% to 5.8\% and 2.0\% to 3.2\% with the three models). Content-defined anchors lose 5.9 positions per insertion or deletion (5.9 to 6.2 with the three models), within the bounds of \cref{prop:exposure} and close to the 4.4 positions of four-token contexts, so SOLO is among the most robust variants after insertions and deletions (17.6\%, on par with the best window setting).

\noindent\textbf{Why SOLO works: longer local contexts.}
A request-scoped history over longer local contexts also raises $\omega$ (65\% for six and 73\% for eight tokens), but every edited token then enters $h$ later contexts. With Qwen2.5-7B-Instruct, after substitutions and after insertions and deletions, SOLO detects more JSON responses than six-token contexts ($+1.8$ and $+3.4$ points) and than eight-token contexts ($+4.0$ and $+5.9$ points, see \cref{tab:ablation-full} for plain GSM8K). Its lead over eight-token contexts after both kinds of edits holds with the other two models as well ($+4.0$ and $+4.9$ points with Qwen2.5-14B-Instruct, $+6.5$ and $+6.7$ points with Ministral-3-8B-Instruct, all with confidence intervals above zero). Longer local contexts thus trade robustness for $\omega$, whereas with Qwen2.5-7B-Instruct in the setting of \cref{fig:why}, content-defined anchors raise $\omega$ at the robustness of short contexts.

\begin{table*}[!htbp]
\centering
{\scriptsize{\setlength{\tabcolsep}{6pt}
\begin{tabular}{lrrrrrrrrrr}
\toprule
 & \multicolumn{5}{c}{GSM8K, JSON} & \multicolumn{5}{c}{GSM8K} \\
\cmidrule(lr){2-6} \cmidrule(lr){7-11}
Variant & $\Delta$maj & $\Delta$AUROC & Clean & Subst. & Ins./del. & $\Delta$maj & $\Delta$AUROC & Clean & Subst. & Ins./del. \\
\midrule
\rowcolor[HTML]{DCEBFA} \bfseries SOLO ($h = 2$, two windows) & \bfseries $+0.4$ & \bfseries $-0.2$ & \bfseries 37.1 & \bfseries 14.6 & \bfseries 17.6 & \bfseries $-0.5$ & \bfseries $+2.0$ & \bfseries 21.9 & \bfseries 8.7 & \bfseries 9.2 \\
w/o anchors & $+0.7$ & $-2.4$ & 18.4 & 10.5 & 11.5 & $-0.2$ & $-0.1$ & 9.1 & 5.4 & 5.6 \\
w/o single-owner contexts & \cellcolor[HTML]{FBE0DA} $-4.2$ & \cellcolor[HTML]{FBE0DA} $-11.4$ & 55.7 & 24.9 & 28.6 & \cellcolor[HTML]{FBE0DA} $-3.5$ & \cellcolor[HTML]{FBE0DA} $-5.4$ & 35.0 & 13.2 & 15.1 \\
one window $[t-32, t-2)$ & $+0.8$ & $-0.5$ & 26.2 & 12.4 & 15.2 & $-0.2$ & $+1.7$ & 14.0 & 6.8 & 7.5 \\
\midrule
$h = 1$ & $+0.7$ & $-1.0$ & 30.8 & 13.5 & 15.6 & $-0.3$ & $+2.1$ & 17.1 & 6.6 & 8.2 \\
$h = 3$ & $+0.4$ & $-2.0$ & 41.1 & 13.0 & 16.1 & $+0.3$ & $-0.1$ & 21.7 & 7.2 & 8.1 \\
windows $\times 0.6$ & $+0.7$ & $-1.3$ & 38.7 & 15.3 & 17.7 & $+0.1$ & $+0.2$ & 21.4 & 7.5 & 9.4 \\
windows $\times 1.5$ & $+0.3$ & $-1.5$ & 35.9 & 14.1 & 16.6 & $-0.1$ & $+1.9$ & 21.0 & 8.8 & 9.0 \\
\midrule
w/o anchors, local context of 4 tokens & $+0.6$ & $-1.6$ & 28.6 & 12.9 & 14.3 & $-0.3$ & $+1.6$ & 15.6 & 7.3 & 7.8 \\
w/o anchors, local context of 6 tokens & $+0.6$ & $+0.8$ & 36.1 & 12.8 & 14.3 & $+0.1$ & $-0.5$ & 19.2 & 6.5 & 6.3 \\
w/o anchors, local context of 8 tokens & $+0.1$ & $+0.5$ & 39.9 & 10.6 & 11.7 & $-0.3$ & $+0.0$ & 21.3 & 5.5 & 6.1 \\
\midrule
\rowcolor[HTML]{ECECEC} stratified keys (reference) & $+0.3$ & $-0.4$ & 33.1 & 11.7 & 13.6 & $-0.3$ & $+2.8$ & 16.2 & 5.2 & 5.6 \\
\bottomrule
\end{tabular}}}
\caption{Ablation and sensitivity in both formats (held-out GSM8K, Qwen2.5-7B-Instruct, three seeds). $\Delta$: difference to unwatermarked sampling in points. Clean, Subst., Ins./del.: TPR (\%) at a calibrated 1\% FPR without edits and after 10\% substitutions or 10\% insertions and deletions. Pale blue: SOLO. Coral: not admissible. Gray: reference.}\label{tab:ablation-full}
\end{table*}

\noindent\textbf{Ablation and sensitivity.}
We remove one component of SOLO at a time (\cref{tab:ablation,tab:ablation-full}) and repeat the ablation in JSON format with Qwen2.5-14B-Instruct and Ministral-3-8B-Instruct, a larger model of the same family and a model of another family (\cref{app:models}). \emph{Without anchors} (single-owner contexts of the last two tokens only), siblings compete for the same short contexts and the TPR drops (likewise with the other two models), as \cref{prop:dilution} predicts. \emph{Without single-owner contexts} (anchored contexts, fixed key), every token carries the watermark and the TPR rises, but the responses couple again: 15\% of the JSON requests return ten identical responses, and maj@10 and the AUROC drop, so this variant is not admissible with any of the three models. A single anchor window also lowers the TPR. Each component is thus needed for one of the three requirements.

The sensitivity rows vary one parameter at a time. A local width of $h = 1$ lowers the TPR with all three models, and $h = 3$ raises the clean TPR on JSON responses but lowers it after substitutions, in line with an exposure that grows with $h$ (\cref{prop:exposure}). Scaling both windows by $0.6$ or $1.5$ changes the TPR by at most 3 points with all three models, so the choice of the windows is not critical. We did not retune them on the held-out data. All single-owner variants are $K$-sample distortion-free by \cref{thm:exact}. Their AUROC differences to unwatermarked sampling range from \ensuremath{-}2.4 to +3.3 points across the variants and formats of \cref{tab:ablation-full} and the JSON format of the other two models. Only 2 of these 40 intervals, both with Qwen2.5-7B-Instruct (without anchors and $h = 3$ in JSON format), exclude zero and reach below $-2$ points, so our admissibility rule would reject these exact variants. Since most JSON differences of Qwen2.5-7B-Instruct are negative (all except the six- and eight-token local contexts) and all share one unwatermarked baseline, we attribute them to the sampling noise of three seeds.

\begin{figure*}[!t]
\begin{minipage}[t]{\columnwidth}\vspace{0pt}
\centering
\includegraphics[width=\columnwidth]{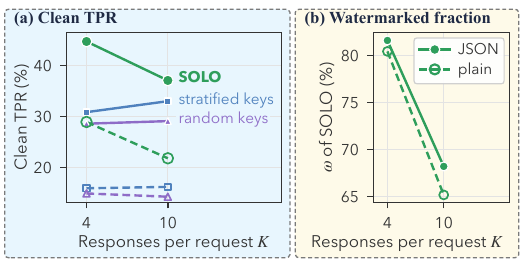}
\caption{Number of responses per request, $K \in \{4, 10\}$ (held-out GSM8K, Qwen2.5-7B-Instruct, three seeds). (a) Clean TPR at a calibrated 1\% FPR of SOLO, stratified keys (ten keys), and random keys (16 keys). (b) Watermarked fraction $\omega$ of SOLO. Solid, filled: JSON format. Dashed, hollow: plain format.}\label{fig:ksweep}
\end{minipage}\hfill
\begin{minipage}[t]{\columnwidth}\vspace{0pt}
\centering
\includegraphics[width=\columnwidth]{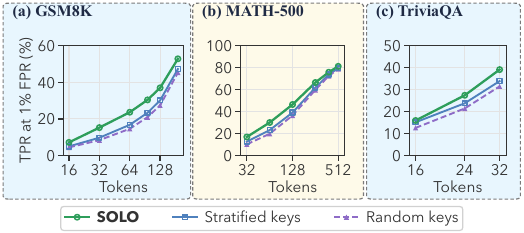}
\caption{TPR at a calibrated 1\% FPR on the first $L$ tokens (horizontal axis) of responses with at least $L$ tokens, for (a) GSM8K, (b) MATH-500, and (c) TriviaQA, mean over models and formats (first 300 requests per run).}\label{fig:length}
\end{minipage}
\end{figure*}

\noindent\textbf{Number of responses per request.}
\Cref{fig:ksweep} varies the number of responses per request. Fewer siblings contest fewer contexts, so the watermarked fraction of SOLO rises from 68\% for $K = 10$ to 82\% for $K = 4$ on JSON responses, and its clean TPR from 37.1\% to 44.8\%, whereas the key pools keep their pools (ten stratified keys, 16 random keys) and are unaffected by $K$. A stratified pool sized to the request, four keys for $K = 4$, would pay a smaller multiple-testing penalty, so these numbers compare SOLO with fixed pools. The lead over the fixed ten-key pool therefore grows from +4.1 to +13.9 points on JSON responses and from +5.6 to +13.0 points on plain responses. Our evaluation covers requests of up to 10 responses at temperature 1, the regime of self-consistency and sampling-based uncertainty estimation that we target.

\noindent\textbf{Robustness to token edits.}
On the 6 Qwen2.5-7B-Instruct and 5 Qwen2.5-14B-Instruct combinations, where every inserted or substituted id is a valid token (\cref{app:impl}), SOLO detects significantly more responses than stratified keys on 4 and 5 of them after 10\% substitutions and on 5 and 5 after 10\% insertions and deletions, and it is significantly behind on none. Against random keys it leads on all of them in both conditions. After 10\% random substitutions, SOLO detects significantly more responses than stratified keys on 16 of the 26 combinations and than random keys on 23, and after 10\% random insertions and deletions, than stratified keys on 21 and than random keys on 26. \cref{fig:robust} varies the share of edited tokens, re-estimating the threshold on unwatermarked responses under the same edits. After insertions and deletions, SOLO is significantly behind neither key pool in any combination at any rate and significantly ahead of stratified keys on 18 to 21 of them. With 20\% insertions and deletions, its mean TPR is 14.9\% against 12.2\%. Its single-key detector loses no power to multiple testing, and its anchors keep most contexts intact under insertions and deletions (\cref{sec:exp-why}), unlike anchors at fixed distances and salts derived from the prefix, whose TPR after insertions and deletions falls to a few percent. We restrict our robustness claims to random token-level edits.

\noindent\textbf{Response length.}
\Cref{fig:length} scores only the first $L$ tokens of the responses with at least $L$ tokens, re-estimating the threshold on unwatermarked responses truncated in the same way. On average, SOLO detects more responses than both key pools at every length of \cref{fig:length} on all three benchmarks (by at least 0.8 points), for instance 23.6\% against 16.7\% for stratified keys after 64 tokens on GSM8K, so its advantage does not depend on long responses.

\section{Detection Statistics}\label{app:stats}
\begin{table*}[!t]
\centering
{\scriptsize{\setlength{\tabcolsep}{4pt}
\begin{tabular*}{\textwidth}{@{\extracolsep{\fill}}llccc@{}}
\toprule
Model & Task & Stratified keys & Random keys & Context history \\
\midrule
Qwen2.5-7B & GSM8K & \cellcolor[HTML]{D9F0F2} +5.6 {\tiny[+3.4, +8.1]} & \cellcolor[HTML]{D9F0F2} +7.5 {\tiny[+5.2, +9.9]} & \cellcolor[HTML]{D9F0F2} +6.2 {\tiny[+4.2, +8.2]} \\
 & GSM8K, JSON & \cellcolor[HTML]{D9F0F2} +4.1 {\tiny[+1.6, +6.8]} & \cellcolor[HTML]{D9F0F2} +8.0 {\tiny[+5.6, +10.3]} & \cellcolor[HTML]{D9F0F2} +8.6 {\tiny[+5.6, +12.0]} \\
 & MATH-500 & \cellcolor[HTML]{D9F0F2} +3.3 {\tiny[+1.8, +4.9]} & \cellcolor[HTML]{D9F0F2} +4.8 {\tiny[+3.2, +6.4]} & \cellcolor[HTML]{D9F0F2} +2.5 {\tiny[+0.7, +4.0]} \\
 & MATH-500, JSON & \cellcolor[HTML]{D9F0F2} +3.0 {\tiny[+0.7, +5.4]} & \cellcolor[HTML]{D9F0F2} +5.9 {\tiny[+4.1, +7.7]} & \cellcolor[HTML]{D9F0F2} +4.6 {\tiny[+2.0, +7.0]} \\
 & TriviaQA, CoT & \cellcolor[HTML]{D9F0F2} +1.6 {\tiny[+0.3, +2.8]} & \cellcolor[HTML]{D9F0F2} +2.6 {\tiny[+1.3, +3.5]} & +1.2 {\tiny[\ensuremath{-}0.1, +2.3]} \\
 & TriviaQA, JSON & \cellcolor[HTML]{D9F0F2} +3.5 {\tiny[+1.5, +5.4]} & \cellcolor[HTML]{D9F0F2} +3.6 {\tiny[+1.9, +5.3]} & \cellcolor[HTML]{D9F0F2} +2.3 {\tiny[+0.6, +4.3]} \\
\midrule
Olmo-3-7B & GSM8K & \cellcolor[HTML]{D9F0F2} +3.7 {\tiny[+2.1, +5.8]} & \cellcolor[HTML]{D9F0F2} +5.9 {\tiny[+4.7, +7.8]} & \cellcolor[HTML]{D9F0F2} +5.4 {\tiny[+3.1, +8.3]} \\
 & GSM8K, JSON & +0.5 {\tiny[\ensuremath{-}0.7, +1.9]} & \cellcolor[HTML]{D9F0F2} +2.2 {\tiny[+0.9, +3.3]} & \cellcolor[HTML]{D9F0F2} +2.0 {\tiny[+0.7, +3.4]} \\
 & MATH-500, JSON & \cellcolor[HTML]{D9F0F2} +3.0 {\tiny[+1.2, +5.2]} & \cellcolor[HTML]{D9F0F2} +5.3 {\tiny[+3.8, +7.0]} & \cellcolor[HTML]{D9F0F2} +2.9 {\tiny[+1.4, +4.7]} \\
 & TriviaQA, CoT & \ensuremath{-}0.6 {\tiny[\ensuremath{-}2.0, +0.6]} & +0.9 {\tiny[\ensuremath{-}0.6, +2.5]} & +0.7 {\tiny[\ensuremath{-}0.5, +1.8]} \\
 & TriviaQA, JSON & \cellcolor[HTML]{D9F0F2} +4.0 {\tiny[+2.1, +6.1]} & \cellcolor[HTML]{D9F0F2} +6.2 {\tiny[+4.6, +7.9]} & +2.0 {\tiny[\ensuremath{-}0.2, +4.2]} \\
\midrule
Ministral-3-8B & GSM8K & \cellcolor[HTML]{D9F0F2} +6.1 {\tiny[+1.6, +10.8]} & \cellcolor[HTML]{D9F0F2} +7.0 {\tiny[+3.6, +10.4]} & \cellcolor[HTML]{D9F0F2} +5.9 {\tiny[+2.8, +9.2]} \\
 & GSM8K, JSON & \cellcolor[HTML]{D9F0F2} +9.1 {\tiny[+5.6, +13.3]} & \cellcolor[HTML]{D9F0F2} +10.6 {\tiny[+6.4, +15.3]} & \cellcolor[HTML]{D9F0F2} +8.4 {\tiny[+4.9, +11.7]} \\
 & MATH-500, JSON & \cellcolor[HTML]{D9F0F2} +2.7 {\tiny[+0.6, +4.5]} & \cellcolor[HTML]{D9F0F2} +4.0 {\tiny[+1.8, +5.7]} & \cellcolor[HTML]{D9F0F2} +3.6 {\tiny[+1.6, +5.8]} \\
 & TriviaQA, CoT & \ensuremath{-}0.2 {\tiny[\ensuremath{-}2.5, +1.9]} & +2.1 {\tiny[\ensuremath{-}0.5, +4.3]} & +1.1 {\tiny[\ensuremath{-}0.8, +2.8]} \\
 & TriviaQA, JSON & \cellcolor[HTML]{D9F0F2} +4.1 {\tiny[+0.4, +7.9]} & \cellcolor[HTML]{D9F0F2} +10.7 {\tiny[+7.9, +12.6]} & +1.5 {\tiny[\ensuremath{-}1.8, +5.6]} \\
\midrule
Qwen2.5-14B & GSM8K & \cellcolor[HTML]{D9F0F2} +8.9 {\tiny[+5.4, +12.8]} & \cellcolor[HTML]{D9F0F2} +10.4 {\tiny[+8.1, +13.6]} & \cellcolor[HTML]{D9F0F2} +11.4 {\tiny[+8.3, +14.7]} \\
 & GSM8K, JSON & +2.9 {\tiny[\ensuremath{-}0.9, +6.3]} & \cellcolor[HTML]{D9F0F2} +5.7 {\tiny[+2.2, +8.7]} & \cellcolor[HTML]{D9F0F2} +9.0 {\tiny[+4.6, +13.6]} \\
 & MATH-500, JSON & +1.3 {\tiny[\ensuremath{-}0.6, +3.0]} & \cellcolor[HTML]{D9F0F2} +3.4 {\tiny[+1.7, +5.0]} & \cellcolor[HTML]{D9F0F2} +2.1 {\tiny[+0.2, +3.7]} \\
 & TriviaQA, CoT & \cellcolor[HTML]{D9F0F2} +2.8 {\tiny[+1.5, +3.9]} & \cellcolor[HTML]{D9F0F2} +3.0 {\tiny[+1.8, +4.2]} & \cellcolor[HTML]{D9F0F2} +1.4 {\tiny[+0.5, +2.5]} \\
 & TriviaQA, JSON & \cellcolor[HTML]{D9F0F2} +4.0 {\tiny[+2.4, +5.5]} & \cellcolor[HTML]{D9F0F2} +4.8 {\tiny[+3.4, +6.4]} & \cellcolor[HTML]{D9F0F2} +2.4 {\tiny[+1.1, +3.6]} \\
\midrule
Ministral-3-14B & GSM8K & \cellcolor[HTML]{D9F0F2} +5.9 {\tiny[+2.4, +9.3]} & \cellcolor[HTML]{D9F0F2} +8.0 {\tiny[+5.7, +10.2]} & \cellcolor[HTML]{D9F0F2} +8.3 {\tiny[+5.4, +10.9]} \\
 & GSM8K, JSON & \cellcolor[HTML]{D9F0F2} +7.7 {\tiny[+5.4, +9.6]} & \cellcolor[HTML]{D9F0F2} +11.6 {\tiny[+9.3, +13.7]} & \cellcolor[HTML]{D9F0F2} +8.6 {\tiny[+5.9, +10.6]} \\
 & MATH-500, JSON & \cellcolor[HTML]{D9F0F2} +4.0 {\tiny[+0.9, +6.6]} & \cellcolor[HTML]{D9F0F2} +6.1 {\tiny[+4.0, +8.3]} & \cellcolor[HTML]{D9F0F2} +4.8 {\tiny[+2.7, +6.8]} \\
 & TriviaQA, CoT & \cellcolor[HTML]{D9F0F2} +5.6 {\tiny[+3.3, +8.4]} & \cellcolor[HTML]{D9F0F2} +7.4 {\tiny[+5.2, +9.9]} & \cellcolor[HTML]{D9F0F2} +5.4 {\tiny[+3.1, +8.4]} \\
 & TriviaQA, JSON & \cellcolor[HTML]{D9F0F2} +6.3 {\tiny[+3.6, +9.0]} & \cellcolor[HTML]{D9F0F2} +11.0 {\tiny[+8.8, +13.1]} & +3.0 {\tiny[\ensuremath{-}2.0, +7.3]} \\
\bottomrule
\end{tabular*}}}
\caption{SOLO minus each opponent, TPR (points) at a calibrated 1\% FPR on clean responses, with the 95\% CI of the hierarchical bootstrap. Teal: CI above zero. Coral: CI below zero. Bounds are rounded, so a colored bound can read 0.0.}\label{tab:pairwise-clean}
\end{table*}

\begin{table*}[!t]
\centering
{\scriptsize{\setlength{\tabcolsep}{4pt}
\begin{tabular*}{\textwidth}{@{\extracolsep{\fill}}llccc@{}}
\toprule
Model & Task & Stratified keys & Random keys & Context history \\
\midrule
Qwen2.5-7B & GSM8K & \cellcolor[HTML]{D9F0F2} +3.4 {\tiny[+2.2, +4.9]} & \cellcolor[HTML]{D9F0F2} +4.5 {\tiny[+3.4, +5.8]} & +1.3 {\tiny[\ensuremath{-}0.3, +3.0]} \\
 & GSM8K, JSON & \cellcolor[HTML]{D9F0F2} +2.8 {\tiny[+0.6, +4.9]} & \cellcolor[HTML]{D9F0F2} +4.1 {\tiny[+2.4, +5.8]} & \cellcolor[HTML]{D9F0F2} +1.7 {\tiny[+0.1, +3.3]} \\
 & MATH-500 & \cellcolor[HTML]{D9F0F2} +2.4 {\tiny[+1.1, +3.8]} & \cellcolor[HTML]{D9F0F2} +3.2 {\tiny[+2.1, +4.5]} & +0.8 {\tiny[\ensuremath{-}0.4, +2.0]} \\
 & MATH-500, JSON & \cellcolor[HTML]{D9F0F2} +3.2 {\tiny[+1.1, +6.0]} & \cellcolor[HTML]{D9F0F2} +5.9 {\tiny[+2.3, +9.7]} & \ensuremath{-}1.2 {\tiny[\ensuremath{-}3.4, +1.3]} \\
 & TriviaQA, CoT & +0.3 {\tiny[\ensuremath{-}1.2, +1.7]} & \cellcolor[HTML]{D9F0F2} +0.8 {\tiny[+0.1, +1.9]} & \ensuremath{-}0.1 {\tiny[\ensuremath{-}1.1, +0.8]} \\
 & TriviaQA, JSON & +0.7 {\tiny[\ensuremath{-}0.5, +1.8]} & \cellcolor[HTML]{D9F0F2} +0.9 {\tiny[+0.0, +1.9]} & \ensuremath{-}0.3 {\tiny[\ensuremath{-}1.4, +0.9]} \\
\midrule
Olmo-3-7B & GSM8K & \cellcolor[HTML]{D9F0F2} +3.6 {\tiny[+2.2, +4.9]} & \cellcolor[HTML]{D9F0F2} +4.9 {\tiny[+3.3, +6.5]} & +1.3 {\tiny[\ensuremath{-}0.8, +3.0]} \\
 & GSM8K, JSON & \cellcolor[HTML]{D9F0F2} +1.2 {\tiny[+0.3, +2.1]} & \cellcolor[HTML]{D9F0F2} +1.9 {\tiny[+0.7, +3.0]} & +0.3 {\tiny[\ensuremath{-}0.9, +1.2]} \\
 & MATH-500, JSON & \cellcolor[HTML]{D9F0F2} +3.0 {\tiny[+1.2, +4.8]} & \cellcolor[HTML]{D9F0F2} +4.5 {\tiny[+3.1, +5.8]} & +0.2 {\tiny[\ensuremath{-}2.4, +2.5]} \\
 & TriviaQA, CoT & \cellcolor[HTML]{FBE0DA} \ensuremath{-}1.7 {\tiny[\ensuremath{-}3.0, \ensuremath{-}0.4]} & \ensuremath{-}0.1 {\tiny[\ensuremath{-}1.1, +0.9]} & \cellcolor[HTML]{FBE0DA} \ensuremath{-}1.7 {\tiny[\ensuremath{-}2.8, \ensuremath{-}0.5]} \\
 & TriviaQA, JSON & \cellcolor[HTML]{D9F0F2} +1.5 {\tiny[+0.1, +2.6]} & \cellcolor[HTML]{D9F0F2} +2.7 {\tiny[+1.6, +3.9]} & \ensuremath{-}1.1 {\tiny[\ensuremath{-}3.0, +0.7]} \\
\midrule
Ministral-3-8B & GSM8K & +1.9 {\tiny[\ensuremath{-}0.8, +4.6]} & \cellcolor[HTML]{D9F0F2} +3.4 {\tiny[+1.7, +5.1]} & \cellcolor[HTML]{FBE0DA} \ensuremath{-}2.9 {\tiny[\ensuremath{-}4.9, \ensuremath{-}1.0]} \\
 & GSM8K, JSON & \cellcolor[HTML]{D9F0F2} +4.8 {\tiny[+2.2, +7.4]} & \cellcolor[HTML]{D9F0F2} +6.3 {\tiny[+3.8, +8.8]} & +2.1 {\tiny[\ensuremath{-}0.3, +4.0]} \\
 & MATH-500, JSON & \cellcolor[HTML]{D9F0F2} +1.8 {\tiny[+0.4, +3.1]} & \cellcolor[HTML]{D9F0F2} +2.6 {\tiny[+1.0, +4.0]} & \ensuremath{-}1.7 {\tiny[\ensuremath{-}3.4, +0.1]} \\
 & TriviaQA, CoT & \ensuremath{-}0.2 {\tiny[\ensuremath{-}2.0, +1.9]} & +1.1 {\tiny[\ensuremath{-}0.8, +3.7]} & \cellcolor[HTML]{FBE0DA} \ensuremath{-}2.1 {\tiny[\ensuremath{-}3.8, \ensuremath{-}0.0]} \\
 & TriviaQA, JSON & +1.2 {\tiny[\ensuremath{-}3.7, +6.5]} & \cellcolor[HTML]{D9F0F2} +4.3 {\tiny[+0.8, +8.2]} & \ensuremath{-}1.9 {\tiny[\ensuremath{-}3.8, +0.4]} \\
\midrule
Qwen2.5-14B & GSM8K & \cellcolor[HTML]{D9F0F2} +4.4 {\tiny[+1.9, +6.5]} & \cellcolor[HTML]{D9F0F2} +6.7 {\tiny[+4.6, +8.3]} & +2.8 {\tiny[\ensuremath{-}0.3, +5.8]} \\
 & GSM8K, JSON & \cellcolor[HTML]{D9F0F2} +2.8 {\tiny[+1.5, +4.2]} & \cellcolor[HTML]{D9F0F2} +3.8 {\tiny[+2.4, +5.1]} & +2.4 {\tiny[\ensuremath{-}0.0, +4.0]} \\
 & MATH-500, JSON & \cellcolor[HTML]{D9F0F2} +2.4 {\tiny[+0.3, +4.4]} & \cellcolor[HTML]{D9F0F2} +4.7 {\tiny[+2.7, +6.6]} & \ensuremath{-}1.4 {\tiny[\ensuremath{-}3.6, +1.3]} \\
 & TriviaQA, CoT & \cellcolor[HTML]{D9F0F2} +0.9 {\tiny[+0.2, +1.6]} & \cellcolor[HTML]{D9F0F2} +1.1 {\tiny[+0.4, +1.7]} & \ensuremath{-}0.4 {\tiny[\ensuremath{-}1.2, +0.4]} \\
 & TriviaQA, JSON & \cellcolor[HTML]{D9F0F2} +1.5 {\tiny[+0.5, +2.1]} & \cellcolor[HTML]{D9F0F2} +1.9 {\tiny[+0.9, +2.8]} & +0.5 {\tiny[\ensuremath{-}0.4, +1.2]} \\
\midrule
Ministral-3-14B & GSM8K & \cellcolor[HTML]{D9F0F2} +2.6 {\tiny[+0.3, +5.2]} & \cellcolor[HTML]{D9F0F2} +4.0 {\tiny[+2.0, +5.6]} & \ensuremath{-}0.5 {\tiny[\ensuremath{-}2.7, +2.0]} \\
 & GSM8K, JSON & \ensuremath{-}0.4 {\tiny[\ensuremath{-}5.1, +3.6]} & +2.4 {\tiny[\ensuremath{-}0.4, +5.4]} & \ensuremath{-}1.4 {\tiny[\ensuremath{-}2.8, +0.5]} \\
 & MATH-500, JSON & +1.8 {\tiny[\ensuremath{-}0.6, +4.2]} & \cellcolor[HTML]{D9F0F2} +3.6 {\tiny[+2.1, +5.1]} & \ensuremath{-}0.5 {\tiny[\ensuremath{-}2.1, +1.3]} \\
 & TriviaQA, CoT & +1.4 {\tiny[\ensuremath{-}0.3, +3.1]} & \cellcolor[HTML]{D9F0F2} +2.6 {\tiny[+1.2, +4.0]} & \ensuremath{-}1.4 {\tiny[\ensuremath{-}4.0, +1.0]} \\
 & TriviaQA, JSON & +2.7 {\tiny[\ensuremath{-}1.1, +5.6]} & \cellcolor[HTML]{D9F0F2} +5.5 {\tiny[+2.5, +7.9]} & \ensuremath{-}1.6 {\tiny[\ensuremath{-}3.8, +0.3]} \\
\bottomrule
\end{tabular*}}}
\caption{As \cref{tab:pairwise-clean}, after 10\% random substitutions (thresholds re-calibrated under the same edits).}\label{tab:pairwise-subst}
\end{table*}

\begin{table*}[!t]
\centering
{\scriptsize{\setlength{\tabcolsep}{4pt}
\begin{tabular*}{\textwidth}{@{\extracolsep{\fill}}llccc@{}}
\toprule
Model & Task & Stratified keys & Random keys & Context history \\
\midrule
Qwen2.5-7B & GSM8K & \cellcolor[HTML]{D9F0F2} +3.6 {\tiny[+2.6, +4.6]} & \cellcolor[HTML]{D9F0F2} +4.2 {\tiny[+3.2, +5.4]} & \cellcolor[HTML]{D9F0F2} +1.4 {\tiny[+0.2, +2.7]} \\
 & GSM8K, JSON & \cellcolor[HTML]{D9F0F2} +4.1 {\tiny[+2.5, +6.1]} & \cellcolor[HTML]{D9F0F2} +6.5 {\tiny[+4.9, +8.6]} & \cellcolor[HTML]{D9F0F2} +3.3 {\tiny[+1.6, +5.0]} \\
 & MATH-500 & \cellcolor[HTML]{D9F0F2} +2.2 {\tiny[+1.0, +3.7]} & \cellcolor[HTML]{D9F0F2} +3.6 {\tiny[+2.4, +4.9]} & +1.2 {\tiny[\ensuremath{-}0.3, +3.0]} \\
 & MATH-500, JSON & \cellcolor[HTML]{D9F0F2} +6.5 {\tiny[+4.4, +8.4]} & \cellcolor[HTML]{D9F0F2} +9.0 {\tiny[+6.7, +11.1]} & \cellcolor[HTML]{D9F0F2} +2.3 {\tiny[+0.1, +4.6]} \\
 & TriviaQA, CoT & +0.7 {\tiny[\ensuremath{-}0.0, +1.3]} & \cellcolor[HTML]{D9F0F2} +1.3 {\tiny[+0.7, +2.1]} & +0.3 {\tiny[\ensuremath{-}0.5, +1.1]} \\
 & TriviaQA, JSON & \cellcolor[HTML]{D9F0F2} +1.5 {\tiny[+0.7, +2.2]} & \cellcolor[HTML]{D9F0F2} +2.1 {\tiny[+1.4, +2.8]} & +0.6 {\tiny[\ensuremath{-}0.4, +1.5]} \\
\midrule
Olmo-3-7B & GSM8K & \cellcolor[HTML]{D9F0F2} +3.6 {\tiny[+2.2, +5.0]} & \cellcolor[HTML]{D9F0F2} +4.8 {\tiny[+3.3, +6.4]} & +0.9 {\tiny[\ensuremath{-}1.3, +3.3]} \\
 & GSM8K, JSON & \cellcolor[HTML]{D9F0F2} +1.9 {\tiny[+1.0, +2.8]} & \cellcolor[HTML]{D9F0F2} +2.3 {\tiny[+1.3, +3.2]} & +0.8 {\tiny[\ensuremath{-}0.4, +2.0]} \\
 & MATH-500, JSON & \cellcolor[HTML]{D9F0F2} +4.0 {\tiny[+2.5, +5.9]} & \cellcolor[HTML]{D9F0F2} +6.7 {\tiny[+5.4, +8.3]} & +1.5 {\tiny[\ensuremath{-}0.6, +3.5]} \\
 & TriviaQA, CoT & \ensuremath{-}0.1 {\tiny[\ensuremath{-}1.1, +1.1]} & \cellcolor[HTML]{D9F0F2} +1.3 {\tiny[+0.2, +2.5]} & +0.0 {\tiny[\ensuremath{-}0.9, +1.1]} \\
 & TriviaQA, JSON & \cellcolor[HTML]{D9F0F2} +2.9 {\tiny[+1.7, +4.2]} & \cellcolor[HTML]{D9F0F2} +4.1 {\tiny[+2.6, +5.5]} & \ensuremath{-}0.1 {\tiny[\ensuremath{-}2.5, +2.1]} \\
\midrule
Ministral-3-8B & GSM8K & \cellcolor[HTML]{D9F0F2} +4.5 {\tiny[+2.7, +6.5]} & \cellcolor[HTML]{D9F0F2} +6.2 {\tiny[+4.0, +8.3]} & +1.1 {\tiny[\ensuremath{-}1.2, +3.0]} \\
 & GSM8K, JSON & \cellcolor[HTML]{D9F0F2} +5.0 {\tiny[+1.9, +8.4]} & \cellcolor[HTML]{D9F0F2} +6.3 {\tiny[+3.3, +10.0]} & +1.9 {\tiny[\ensuremath{-}0.6, +4.4]} \\
 & MATH-500, JSON & \cellcolor[HTML]{D9F0F2} +2.8 {\tiny[+0.3, +5.3]} & \cellcolor[HTML]{D9F0F2} +4.5 {\tiny[+2.5, +6.8]} & +1.2 {\tiny[\ensuremath{-}0.4, +3.2]} \\
 & TriviaQA, CoT & +0.3 {\tiny[\ensuremath{-}1.0, +2.1]} & \cellcolor[HTML]{D9F0F2} +2.0 {\tiny[+0.1, +4.0]} & \ensuremath{-}2.2 {\tiny[\ensuremath{-}4.0, +0.1]} \\
 & TriviaQA, JSON & +2.3 {\tiny[\ensuremath{-}0.9, +4.8]} & \cellcolor[HTML]{D9F0F2} +5.0 {\tiny[+2.9, +6.7]} & \ensuremath{-}2.2 {\tiny[\ensuremath{-}4.5, +0.0]} \\
\midrule
Qwen2.5-14B & GSM8K & \cellcolor[HTML]{D9F0F2} +7.7 {\tiny[+4.1, +11.3]} & \cellcolor[HTML]{D9F0F2} +9.3 {\tiny[+6.1, +12.2]} & \cellcolor[HTML]{D9F0F2} +5.4 {\tiny[+1.5, +9.1]} \\
 & GSM8K, JSON & \cellcolor[HTML]{D9F0F2} +2.7 {\tiny[+1.2, +4.6]} & \cellcolor[HTML]{D9F0F2} +4.4 {\tiny[+3.1, +6.0]} & \cellcolor[HTML]{D9F0F2} +3.0 {\tiny[+1.1, +5.1]} \\
 & MATH-500, JSON & \cellcolor[HTML]{D9F0F2} +4.6 {\tiny[+2.4, +6.6]} & \cellcolor[HTML]{D9F0F2} +7.1 {\tiny[+5.0, +8.8]} & +2.0 {\tiny[\ensuremath{-}0.1, +4.1]} \\
 & TriviaQA, CoT & \cellcolor[HTML]{D9F0F2} +1.4 {\tiny[+0.5, +2.1]} & \cellcolor[HTML]{D9F0F2} +1.6 {\tiny[+0.9, +2.4]} & +0.1 {\tiny[\ensuremath{-}0.7, +0.8]} \\
 & TriviaQA, JSON & \cellcolor[HTML]{D9F0F2} +1.3 {\tiny[+0.6, +2.2]} & \cellcolor[HTML]{D9F0F2} +1.9 {\tiny[+1.1, +2.9]} & +0.1 {\tiny[\ensuremath{-}0.9, +1.3]} \\
\midrule
Ministral-3-14B & GSM8K & \cellcolor[HTML]{D9F0F2} +5.1 {\tiny[+2.9, +7.4]} & \cellcolor[HTML]{D9F0F2} +6.8 {\tiny[+5.5, +8.2]} & \cellcolor[HTML]{D9F0F2} +1.9 {\tiny[+0.0, +4.0]} \\
 & GSM8K, JSON & +3.9 {\tiny[\ensuremath{-}0.3, +7.0]} & \cellcolor[HTML]{D9F0F2} +6.5 {\tiny[+4.4, +8.6]} & \cellcolor[HTML]{D9F0F2} +2.3 {\tiny[+0.7, +3.9]} \\
 & MATH-500, JSON & \cellcolor[HTML]{D9F0F2} +4.4 {\tiny[+0.2, +8.5]} & \cellcolor[HTML]{D9F0F2} +6.4 {\tiny[+2.5, +10.3]} & +1.3 {\tiny[\ensuremath{-}1.8, +4.5]} \\
 & TriviaQA, CoT & \cellcolor[HTML]{D9F0F2} +4.4 {\tiny[+2.7, +5.9]} & \cellcolor[HTML]{D9F0F2} +6.5 {\tiny[+4.8, +7.9]} & +1.1 {\tiny[\ensuremath{-}1.0, +3.2]} \\
 & TriviaQA, JSON & \cellcolor[HTML]{D9F0F2} +2.5 {\tiny[+0.7, +5.0]} & \cellcolor[HTML]{D9F0F2} +5.5 {\tiny[+3.6, +8.3]} & +0.2 {\tiny[\ensuremath{-}1.5, +2.5]} \\
\bottomrule
\end{tabular*}}}
\caption{As \cref{tab:pairwise-clean}, after 10\% random insertions and deletions (thresholds re-calibrated under the same edits).}\label{tab:pairwise-insdel}
\end{table*}

\begin{table*}[!t]
\centering
{\scriptsize\resizebox{\linewidth}{!}{%
\begin{tabular}{ll>{\columncolor[HTML]{DCEBFA}}rr>{\columncolor[HTML]{DCEBFA}}rr>{\columncolor[HTML]{DCEBFA}}rrc}
\toprule
 & & \multicolumn{2}{c}{1\%} & \multicolumn{2}{c}{0.1\%} & \multicolumn{2}{c}{0.01\%} & \\
Model & Task & SOLO & Opp. & SOLO & Opp. & SOLO & Opp. & Opponent \\
\midrule
Qwen2.5-7B & GSM8K & \bfseries 19.7 (0.73) & 15.7 (0.93) & \bfseries 9.0 (0.06) & 7.7 (0.10) & \bfseries 4.8 (0.01) & 4.3 (0.01) & Stratified keys \\
 & GSM8K, JSON & \bfseries 33.8 (0.74) & 33.4 (1.12) & 19.0 (0.09) & \bfseries 20.0 (0.08) & 11.5 (0.02) & \bfseries 12.6 (0.02) & Stratified keys \\
 & MATH-500 & \bfseries 22.8 (1.29) & 17.9 (0.91) & \bfseries 11.7 (0.16) & 8.1 (0.10) & \bfseries 6.3 (0.01) & 3.9 (0.00) & Context history \\
 & MATH-500, JSON & \bfseries 73.7 (0.87) & 72.3 (1.23) & \bfseries 61.1 (0.13) & \bfseries 61.1 (0.14) & 51.0 (0.01) & \bfseries 52.3 (0.02) & Stratified keys \\
 & TriviaQA, CoT & \bfseries 13.0 (2.56) & 12.7 (3.09) & \bfseries 6.6 (0.42) & 6.3 (0.50) & \bfseries 3.7 (0.04) & \bfseries 3.7 (0.06) & Context history \\
 & TriviaQA, JSON & \bfseries 10.7 (1.06) & 8.6 (0.63) & \bfseries 4.8 (0.08) & 3.6 (0.04) & \bfseries 2.5 (0.00) & 1.8 (0.02) & Context history \\
\midrule
Olmo-3-7B & GSM8K & \bfseries 39.1 (1.31) & 32.7 (0.91) & \bfseries 27.3 (0.19) & 24.1 (0.09) & \bfseries 20.7 (0.02) & 19.3 (0.01) & Stratified keys \\
 & GSM8K, JSON & \bfseries 14.4 (0.83) & 13.8 (0.82) & \bfseries 7.8 (0.09) & \bfseries 7.8 (0.07) & 5.4 (0.00) & \bfseries 5.6 (0.00) & Stratified keys \\
 & MATH-500, JSON & \bfseries 46.7 (0.92) & 46.3 (1.37) & \bfseries 36.1 (0.05) & 35.4 (0.15) & \bfseries 29.9 (0.01) & 29.1 (0.01) & Context history \\
 & TriviaQA, CoT & 27.5 (0.86) & \bfseries 28.3 (0.96) & 21.0 (0.07) & \bfseries 22.2 (0.14) & 17.3 (0.01) & \bfseries 18.5 (0.01) & Stratified keys \\
 & TriviaQA, JSON & 30.6 (0.98) & \bfseries 35.0 (2.46) & 20.7 (0.06) & \bfseries 21.7 (0.17) & \bfseries 15.0 (0.01) & 14.8 (0.02) & Context history \\
\midrule
Ministral-3-8B & GSM8K & \bfseries 54.5 (1.01) & 41.9 (0.46) & \bfseries 38.5 (0.11) & 29.2 (0.02) & \bfseries 28.8 (0.02) & 22.1 (0.00) & Context history \\
 & GSM8K, JSON & \bfseries 48.2 (1.66) & 40.3 (1.07) & \bfseries 33.7 (0.19) & 26.6 (0.10) & \bfseries 25.8 (0.01) & 19.4 (0.00) & Context history \\
 & MATH-500, JSON & \bfseries 82.3 (1.42) & 79.6 (1.10) & \bfseries 74.1 (0.19) & 72.4 (0.11) & \bfseries 67.2 (0.01) & 66.5 (0.01) & Stratified keys \\
 & TriviaQA, CoT & \bfseries 52.0 (1.19) & 50.4 (0.83) & \bfseries 40.7 (0.15) & 40.3 (0.06) & 32.8 (0.01) & \bfseries 33.5 (0.01) & Stratified keys \\
 & TriviaQA, JSON & \bfseries 54.8 (2.64) & 50.3 (0.61) & \bfseries 39.2 (0.34) & 34.1 (0.06) & \bfseries 28.9 (0.05) & 24.5 (0.00) & Context history \\
\midrule
Qwen2.5-14B & GSM8K & \bfseries 50.9 (1.08) & 42.0 (1.10) & \bfseries 31.5 (0.11) & 27.0 (0.08) & \bfseries 19.1 (0.00) & 17.4 (0.03) & Stratified keys \\
 & GSM8K, JSON & 28.9 (0.73) & \bfseries 30.0 (1.02) & 15.9 (0.07) & \bfseries 17.5 (0.09) & 9.9 (0.01) & \bfseries 10.8 (0.01) & Stratified keys \\
 & MATH-500, JSON & \bfseries 76.0 (1.28) & 74.2 (1.15) & 63.7 (0.13) & \bfseries 63.8 (0.13) & 54.2 (0.00) & \bfseries 54.9 (0.01) & Stratified keys \\
 & TriviaQA, CoT & \bfseries 12.4 (2.74) & 12.0 (3.11) & \bfseries 5.8 (0.41) & 5.6 (0.55) & \bfseries 3.4 (0.08) & 3.2 (0.11) & Context history \\
 & TriviaQA, JSON & \bfseries 7.9 (0.79) & 7.1 (0.82) & \bfseries 2.8 (0.03) & 2.3 (0.08) & \bfseries 1.2 (0.01) & 0.9 (0.02) & Context history \\
\midrule
Ministral-3-14B & GSM8K & \bfseries 47.2 (0.89) & 39.9 (0.65) & \bfseries 31.4 (0.08) & 28.2 (0.07) & \bfseries 22.5 (0.01) & 21.1 (0.00) & Stratified keys \\
 & GSM8K, JSON & \bfseries 49.1 (1.96) & 41.9 (1.08) & \bfseries 33.0 (0.28) & 30.2 (0.09) & \bfseries 23.5 (0.04) & 22.9 (0.01) & Stratified keys \\
 & MATH-500, JSON & \bfseries 81.9 (1.49) & 78.1 (1.11) & \bfseries 72.7 (0.17) & 69.9 (0.09) & \bfseries 64.9 (0.02) & 63.0 (0.00) & Stratified keys \\
 & TriviaQA, CoT & \bfseries 51.1 (0.97) & 47.2 (1.13) & \bfseries 37.0 (0.12) & 33.8 (0.14) & \bfseries 27.9 (0.00) & 25.4 (0.00) & Context history \\
 & TriviaQA, JSON & \bfseries 47.8 (2.46) & 41.8 (0.96) & \bfseries 32.2 (0.33) & 26.8 (0.10) & \bfseries 22.8 (0.02) & 19.0 (0.00) & Context history \\
\bottomrule
\end{tabular}
}}
\caption{Clean TPR (\%) at the nominal thresholds of the Gamma test, with the realized FPR (\%) on unwatermarked responses in parentheses, for SOLO and the best other watermark admissible for the model (Opp., named in the last column). No threshold is fitted. Bold: the higher TPR at each level.}\label{tab:nominal}
\end{table*}

This appendix supports \cref{sec:exp-main,sec:exp-why}: it gives every pairwise difference behind the counts of significant leads, the tests behind them, and detection at thresholds that are not fitted to any data.

\noindent\textbf{Procedure.}
Each configuration (model, task, scheme) is run with three keys, one per seed, and every TPR is the mean over the three keys. For a difference between SOLO and one opponent, the 95\% CI comes from a hierarchical bootstrap that resamples the three keys and, within each key, the items, and re-estimates every calibrated threshold in each replicate. A lead (deficit) is significant when the whole CI lies above (below) zero. \Cref{tab:pairwise-clean,tab:pairwise-subst,tab:pairwise-insdel} list these differences for the three opponents of \cref{sec:exp-main} and every combination, and the counts in the text are counts of their colored cells. For the comparison with the best other watermark admissible for the model, a one-sided Holm correction at level 2.5\% runs over the 26 combinations of each edit condition, with the bootstrap standard error and a normal approximation. Of the 17 significant clean leads over that opponent, 13 survive the correction. After substitutions, SOLO leads it significantly on 1 of the combinations and trails it on 2, with 0 leads surviving. After insertions and deletions, it leads on 7 and trails on 0, with 2 leads surviving. Because every configuration has three keys, the key-level variability rests on three replicates.

\noindent\textbf{Thresholds without fitting.}
The calibrated thresholds of the main text are fitted per model, task, key, scheme and edit condition on unwatermarked responses, so their FPR on those responses is 1\% by construction, and the TPR is measured on the separate watermarked responses. \Cref{tab:nominal} instead uses the nominal thresholds of the Gamma test at levels 1\%, 0.1\% and 0.01\%, which are fitted to no data, and reports the realized FPR of each threshold on the unwatermarked responses. At these levels, SOLO detects more clean responses than the best other watermark admissible for the model on 23, 20 and 19 of the 26 combinations (point estimates, with the opponent chosen at the calibrated 1\% FPR). In 14 of the 23 leads at 1\%, in most leads at 0.1\% and in about half at 0.01\%, its realized FPR is also higher, so these counts do not compare the schemes at equal FPR. Only the calibrated thresholds of the main text do. Realized FPRs above the nominal level occur on templated tasks, where unwatermarked responses that share a template evaluate the same keyed contexts under a fixed key (\cref{app:impl}).

\section{Results on Further Models}\label{app:models}
This appendix supports \cref{sec:revisiting,sec:exp-why,sec:exp-ablation}: the funnels, the mechanism of SOLO, and the ablation carry over to Qwen2.5-14B-Instruct and Ministral-3-8B-Instruct.
\begin{figure*}[!t]
\centering
\includegraphics[width=\textwidth]{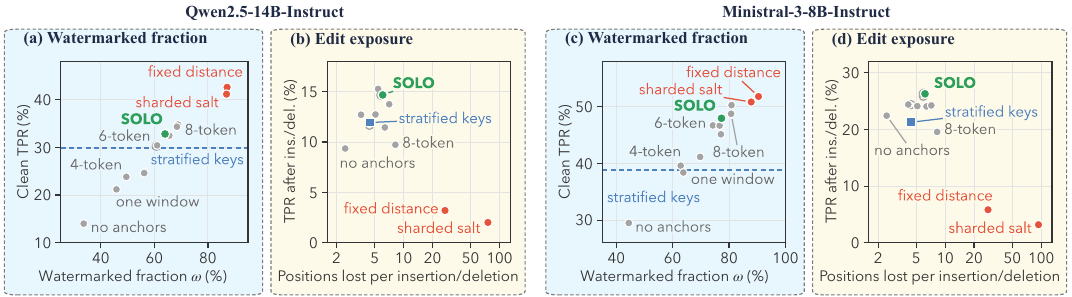}
\caption{Why SOLO works with Qwen2.5-14B-Instruct (a, b) and Ministral-3-8B-Instruct (c, d), as in \cref{fig:why} (held-out GSM8K, JSON format, 600 problems, $K = 10$, three seeds). (a, c) Clean TPR at a calibrated 1\% FPR against the exact watermarked fraction $\omega$ of single-owner context maps. Green: SOLO. Gray points vary $h$ and the windows of SOLO or use $h$-token contexts, and the dashed line marks stratified keys. (b, d) TPR after 10\% insertions and deletions against the number of positions that lose their (context, token) pair per insertion or deletion, measured on 600 unwatermarked responses of one seed. The square marks stratified keys. Anchors at fixed distances and sharded prefix salts (red) reach a high $\omega$ but collapse under insertions and deletions.}\label{fig:why-b4}
\end{figure*}

\noindent\textbf{Ablation.}
\Cref{tab:ablation-b4} repeats the ablation of \cref{tab:ablation} on held-out GSM8K in JSON format, the setting of \cref{fig:why}, with Qwen2.5-14B-Instruct, a larger model of the same family, and Ministral-3-8B-Instruct, a model of another family. As with Qwen2.5-7B-Instruct, removing the anchors, keeping a single anchor window, or shortening the local part to $h = 1$ lowers the clean TPR of SOLO, and removing single-owner contexts couples the responses again: 16\% and 16\% of the requests return ten identical responses, maj@10 drops by 1.8 and 6.9 points, and the AUROC drops by 13.0 and 10.3 points, so that this variant is not admissible. Every single-owner variant is admissible with both models: its AUROC difference to unwatermarked sampling lies between \ensuremath{-}2.1 and +3.3 points with Qwen2.5-14B-Instruct and between \ensuremath{-}1.5 and +1.7 points with Ministral-3-8B-Instruct, and none of these confidence intervals excludes zero.

\begin{table*}[!t]
\centering
\resizebox{\textwidth}{!}{%
\begin{tabular}{lrrrrrrrrrr}
\toprule
 & \multicolumn{5}{c}{Qwen2.5-14B-Instruct (GSM8K, JSON)} & \multicolumn{5}{c}{Ministral-3-8B-Instruct (GSM8K, JSON)} \\
\cmidrule(lr){2-6} \cmidrule(lr){7-11}
Variant & $\Delta$maj & $\Delta$AUROC & Clean & Subst. & Ins./del. & $\Delta$maj & $\Delta$AUROC & Clean & Subst. & Ins./del. \\
\midrule
\rowcolor[HTML]{DCEBFA} \bfseries SOLO ($h = 2$, two windows) & \bfseries $+0.5$ & \bfseries $-0.2$ & \bfseries 32.8 & \bfseries 13.6 & \bfseries 14.7 & \bfseries $+0.0$ & \bfseries $-1.2$ & \bfseries 48.0 & \bfseries 24.8 & \bfseries 26.3 \\
w/o anchors & $-0.1$ & $+3.3$ & 14.0 & 8.9 & 9.4 & $+0.1$ & $-0.6$ & 29.5 & 21.4 & 22.4 \\
w/o single-owner contexts & \cellcolor[HTML]{FBE0DA} $-1.8$ & \cellcolor[HTML]{FBE0DA} $-13.0$ & 51.1 & 22.6 & 25.3 & \cellcolor[HTML]{FBE0DA} $-6.9$ & \cellcolor[HTML]{FBE0DA} $-10.3$ & 68.1 & 40.6 & 43.5 \\
one window $[t-32, t-2)$ & $-0.1$ & $+2.9$ & 21.2 & 10.6 & 12.7 & $+0.4$ & $-0.9$ & 38.4 & 21.7 & 24.4 \\
\midrule
$h = 1$ & $-0.0$ & $+2.8$ & 24.6 & 10.3 & 12.7 & $+0.0$ & $+0.3$ & 41.2 & 21.4 & 24.1 \\
$h = 3$ & $+0.1$ & $+1.0$ & 34.3 & 12.3 & 13.8 & $-0.1$ & $-1.5$ & 50.3 & 21.1 & 24.3 \\
windows $\times 0.6$ & $+0.1$ & $+0.8$ & 32.5 & 12.9 & 14.6 & $-0.1$ & $+0.5$ & 46.7 & 23.0 & 25.9 \\
windows $\times 1.5$ & $+0.4$ & $-2.1$ & 30.4 & 12.1 & 15.3 & $-0.3$ & $-0.7$ & 45.1 & 22.6 & 26.3 \\
\midrule
w/o anchors, local context of 4 tokens & $+0.1$ & $+2.5$ & 23.8 & 11.2 & 11.7 & $+0.1$ & $+1.7$ & 39.6 & 22.7 & 24.4 \\
w/o anchors, local context of 6 tokens & $+0.3$ & $+0.6$ & 30.3 & 11.3 & 11.5 & $-0.6$ & $+0.8$ & 46.7 & 22.1 & 24.1 \\
w/o anchors, local context of 8 tokens & $+0.1$ & $+1.1$ & 34.8 & 9.6 & 9.7 & $+0.0$ & $+0.6$ & 48.8 & 18.2 & 19.6 \\
\midrule
\rowcolor[HTML]{ECECEC} stratified keys (reference) & $+0.3$ & $+0.0$ & 29.9 & 10.8 & 11.9 & $-0.1$ & $+0.9$ & 38.9 & 20.0 & 21.3 \\
\bottomrule
\end{tabular}
}
\caption{Ablation and sensitivity on held-out GSM8K in JSON format with Qwen2.5-14B-Instruct and Ministral-3-8B-Instruct (600 problems, $K = 10$, three seeds), with the rows and JSON columns of \cref{tab:ablation-full} for each model. $\Delta$maj and $\Delta$AUROC are differences to unwatermarked sampling in points. Clean, Subst., and Ins./del.\ are TPRs (\%) at a calibrated 1\% FPR without edits, after 10\% substitutions, and after 10\% insertions and deletions. Pale blue: SOLO. Coral: not admissible. Gray: reference.}\label{tab:ablation-b4}
\end{table*}

\noindent\textbf{Why SOLO works.}
\Cref{fig:why-b4} repeats the analysis of \cref{fig:why}. With both models, detection grows with $\omega$ (Spearman correlation 0.99 and 0.97 over the 12 single-owner context maps), and the anchors of SOLO raise $\omega$ from 50\% and 63\% for local contexts of four tokens to 64\% and 77\%. Anchors at fixed distances and sharded prefix salts reach the largest watermarked fractions, but one insertion or deletion removes the watermark from 27.0 and 28.0 (fixed distances) and 75.6 and 93.9 (sharded salts) positions on average, and their TPR after 10\% insertions and deletions falls to a few percent. SOLO loses 6.1 and 6.2 positions per insertion or deletion (7.1 per insertion and 5.3 per deletion with Ministral-3-8B-Instruct, and in \cref{tab:exposure} for Qwen2.5-14B-Instruct), within the bounds of \cref{prop:exposure}, and is on par with the best window setting after insertions and deletions.

\noindent\textbf{Template funnels.}
\Cref{fig:funnel-ministral} repeats \cref{fig:funnel} with Ministral-3-8B-Instruct. Its tokenizer, like that of Qwen2.5-7B-Instruct, merges the end of the reasoning sentence and the line break before the marker into one token, so that a marker of $m$ tokens again gives a deterministic run of $L_f = m + 1$ tokens. Only the longest marker is split into more tokens than with the tokenizer of Qwen2.5-7B-Instruct ($m = 11$ instead of $10$). For $h = 8$, the inflation is 2\% to 3\% for $L_f \le 6$ and rises to 10\% and 11\% once the run fills the context ($L_f = 8$ and $12$), as \cref{prop:funnel}(b) predicts. Our keyed sampler draws the first $h$ tokens of a response with true randomness, so the inflation measures funnels after the prefix has diverged.

\section{Efficiency}\label{app:eff}
This appendix supports \cref{sec:exp-robust}: SOLO adds little generation time, and its single-key detector costs less than key-pool detectors (\cref{tab:eff}).

\begin{table}[!t]
\centering
{\footnotesize\resizebox{\columnwidth}{!}{{\setlength{\tabcolsep}{8pt}
\begin{tabular}{lrrr}
\toprule
Watermark & \makecell{$\Delta$gen.\\time (\%)} & \makecell{Detection\\(ms)} & \makecell{Detection\\($\times$ fixed key)} \\
\midrule
Fixed key & $-$7.3 & 1.45 & 1.0 \\
Dedup \texttt{all} & $-$5.6 & 1.52 & 1.0 \\
GumbelSoft ($\tau{=}0.5$) & +3.9 & 1.50 & 1.0 \\
GumbelSoft ($\tau{=}0.3$) & +1.2 & 1.50\rlap{\smash{$^{\ddagger}$}} & 1.0 \\
SynthID-Text & +15.5 & 1.59 & 1.1 \\
Random keys & +2.4 & 23.40 & \cellcolor[HTML]{FBE0DA} 16.2 \\
Stratified keys & +4.0 & 14.64 & \cellcolor[HTML]{FBE0DA} 10.1 \\
Context history & +6.2 & 1.45\rlap{\smash{$^{\ddagger}$}} & 1.0 \\
\midrule
\rowcolor[HTML]{DCEBFA} \bfseries SOLO & \bfseries +3.5 & \bfseries 2.17 & \bfseries 1.5 \\
\bottomrule
\end{tabular}}}}
\caption{Cost (Qwen2.5-7B-Instruct): generation time per token vs.\ unwatermarked sampling (mean over tasks and seeds) and detection time (250 tokens, one CPU thread). $^{\ddagger}$Not timed separately: the context history runs the fixed-key detector, and GumbelSoft at $\tau = 0.3$ the detector of $\tau = 0.5$. Coral: detection over ten times the cost of a fixed key.}\label{tab:eff}
\end{table}

\noindent\textbf{Generation.}
The claim set and the anchors add hashing and set operations on the CPU, while decoding is dominated by the forward passes of the model. In our reference implementation on top of Hugging Face Transformers, the generation time per token of SOLO differs from that of unwatermarked sampling by +3.5\% on average over the tasks and seeds of Qwen2.5-7B-Instruct (fixed key \ensuremath{-}7.3\%, stratified keys +4.0\%). Only SynthID-Text, whose tournament draws 30 pseudorandom bits per candidate token, is slower in every task and seed (+15.5\%).

\noindent\textbf{Detection.}
Detection matters more for a provider, since it runs on every text that is checked. On a single CPU thread, the detector of SOLO takes 2.17\,ms for a response of 250 tokens, 1.5 times the cost of the fixed-key detector (1.45\,ms). The overhead comes from recomputing the window minima at every position (\cref{app:method}). Random keys and stratified keys must test every key of the pool and take 23.40\,ms and 14.64\,ms, 16.2 and 10.1 times the cost of the fixed key. The detection cost of key pools thus grows linearly with the pool size, while that of SOLO is independent of $K$.

\begin{table*}[!t]
\centering
{\scriptsize\setlength{\tabcolsep}{6pt}
\begin{tabular}{llrrrrrc}
\toprule
Task & Watermark & $\Delta$maj & $\Delta$AUROC [95\% CI] & Ident. & Infl. (\%) [95\% CI] & $\Delta$cov. & Adm. \\
\midrule
GSM8K & Fixed key & $-2.94$ & $-4.7$ [$-9.0$, $-0.3$] & 4.2 & $+29.2$ [$+23.2$, $+35.0$] & $-3.1$ & \cellcolor[HTML]{FBE0DA} no \\
 & Dedup \texttt{all} & $-1.72$ & $-1.1$ [$-4.3$, $+2.6$] & 1.7 & $+16.6$ [$+11.0$, $+22.0$] & $-1.8$ & \cellcolor[HTML]{FBE0DA} no \\
 & GumbelSoft ($\tau{=}0.5$) & $-0.28$ & $+1.4$ [$-1.0$, $+4.3$] & 0.0 & $-0.9$ [$-6.4$, $+4.2$] & $-0.4$ & \cellcolor[HTML]{D9F0F2} yes \\
 & GumbelSoft ($\tau{=}0.3$) & $-0.61$ & $-0.6$ [$-3.0$, $+1.8$] & 0.0 & $+5.7$ [$+0.2$, $+11.2$] & $-0.6$ & \cellcolor[HTML]{D9F0F2} yes \\
 & SynthID-Text & $-1.89$ & $-2.9$ [$-7.2$, $+1.4$] & 1.1 & $+16.1$ [$+10.2$, $+22.4$] & $-2.0$ & \cellcolor[HTML]{FBE0DA} no \\
 & Random keys & $-0.72$ & $+0.5$ [$-1.9$, $+3.0$] & 0.0 & $+4.3$ [$-0.2$, $+9.0$] & $-0.6$ & \cellcolor[HTML]{D9F0F2} yes \\
 & Stratified keys & $-0.28$ & $+2.8$ [$-0.4$, $+6.3$] & 0.0 & $+2.0$ [$-3.1$, $+6.8$] & $-0.3$ & \cellcolor[HTML]{D9F0F2} yes \\
 & Context history & $-0.28$ & $+1.6$ [$-1.2$, $+4.6$] & 0.0 & $+3.8$ [$-1.1$, $+8.3$] & $-0.1$ & \cellcolor[HTML]{D9F0F2} yes \\
\rowcolor[HTML]{DCEBFA}  & SOLO & $-0.50$ & $+2.0$ [$-1.0$, $+4.9$] & 0.0 & $-1.4$ [$-6.2$, $+3.2$] & $-0.1$ & \cellcolor[HTML]{D9F0F2} yes \\
\midrule
GSM8K, JSON & Fixed key & $-4.00$ & $-10.9$ [$-15.0$, $-6.7$] & 8.7 & $+41.9$ [$+36.4$, $+47.2$] & $-4.6$ & \cellcolor[HTML]{FBE0DA} no \\
 & Dedup \texttt{all} & $-3.94$ & $-8.2$ [$-11.8$, $-4.4$] & 7.9 & $+40.7$ [$+35.2$, $+45.8$] & $-3.9$ & \cellcolor[HTML]{FBE0DA} no \\
 & GumbelSoft ($\tau{=}0.5$) & $+0.06$ & $-1.6$ [$-4.4$, $+1.3$] & 0.0 & $-4.1$ [$-9.5$, $+1.1$] & $+0.3$ & \cellcolor[HTML]{D9F0F2} yes \\
 & GumbelSoft ($\tau{=}0.3$) & $-1.00$ & $-1.6$ [$-4.5$, $+1.2$] & 0.0 & $+1.6$ [$-3.9$, $+6.6$] & $-0.2$ & \cellcolor[HTML]{D9F0F2} yes \\
 & SynthID-Text & $-1.44$ & $-4.4$ [$-6.8$, $-2.0$] & 0.3 & $+8.0$ [$+1.9$, $+14.0$] & $-1.2$ & \cellcolor[HTML]{FBE0DA} no \\
 & Random keys & $+0.28$ & $-1.3$ [$-3.5$, $+0.7$] & 0.0 & $-1.5$ [$-6.6$, $+3.2$] & $+0.1$ & \cellcolor[HTML]{D9F0F2} yes \\
 & Stratified keys & $+0.28$ & $-0.4$ [$-2.5$, $+1.8$] & 0.0 & $-2.8$ [$-7.6$, $+1.6$] & $-0.1$ & \cellcolor[HTML]{D9F0F2} yes \\
 & Context history & $+0.56$ & $-1.6$ [$-3.9$, $+0.2$] & 0.0 & $-0.2$ [$-4.6$, $+4.2$] & $+0.0$ & \cellcolor[HTML]{D9F0F2} yes \\
\rowcolor[HTML]{DCEBFA}  & SOLO & $+0.39$ & $-0.2$ [$-2.4$, $+2.1$] & 0.0 & $-0.4$ [$-5.3$, $+4.2$] & $+0.3$ & \cellcolor[HTML]{D9F0F2} yes \\
\midrule
MATH-500 & Fixed key & $-3.00$ & $-1.9$ [$-3.9$, $+0.2$] & 0.5 & $+19.0$ [$+15.4$, $+22.6$] & $-2.7$ & \cellcolor[HTML]{FBE0DA} no \\
 & Dedup \texttt{all} & $-1.33$ & $-2.4$ [$-4.7$, $-0.4$] & 0.2 & $+13.6$ [$+9.9$, $+17.1$] & $-2.0$ & \cellcolor[HTML]{FBE0DA} no \\
 & GumbelSoft ($\tau{=}0.5$) & $-1.13$ & $-0.9$ [$-2.7$, $+1.0$] & 0.0 & $-0.7$ [$-3.6$, $+2.1$] & $-1.1$ & \cellcolor[HTML]{D9F0F2} yes \\
 & GumbelSoft ($\tau{=}0.3$) & $-1.33$ & $-1.7$ [$-3.7$, $+0.2$] & 0.0 & $+4.8$ [$+1.5$, $+8.1$] & $-0.9$ & \cellcolor[HTML]{D9F0F2} yes \\
 & SynthID-Text & $-2.07$ & $+0.3$ [$-2.0$, $+2.5$] & 0.0 & $+10.8$ [$+7.3$, $+14.5$] & $-1.3$ & \cellcolor[HTML]{FBE0DA} no \\
 & Random keys & $-0.40$ & $-0.6$ [$-2.1$, $+1.1$] & 0.0 & $+2.7$ [$-0.1$, $+5.5$] & $-0.5$ & \cellcolor[HTML]{D9F0F2} yes \\
 & Stratified keys & $-0.47$ & $-0.0$ [$-2.0$, $+2.1$] & 0.0 & $+0.1$ [$-2.9$, $+2.9$] & $+0.1$ & \cellcolor[HTML]{D9F0F2} yes \\
 & Context history & $-0.33$ & $-1.4$ [$-3.3$, $+0.7$] & 0.0 & $+0.3$ [$-2.6$, $+3.1$] & $+0.7$ & \cellcolor[HTML]{D9F0F2} yes \\
\rowcolor[HTML]{DCEBFA}  & SOLO & $-0.13$ & $+0.1$ [$-1.6$, $+2.1$] & 0.0 & $+0.3$ [$-2.6$, $+3.2$] & $+0.3$ & \cellcolor[HTML]{D9F0F2} yes \\
\midrule
MATH-500, JSON & Fixed key & $-6.07$ & $-2.2$ [$-4.5$, $+0.1$] & 2.1 & $+14.5$ [$+11.6$, $+17.4$] & $-5.4$ & \cellcolor[HTML]{FBE0DA} no \\
 & Dedup \texttt{all} & $-6.87$ & $-1.2$ [$-3.5$, $+1.0$] & 2.3 & $+14.2$ [$+11.2$, $+17.2$] & $-5.1$ & \cellcolor[HTML]{FBE0DA} no \\
 & GumbelSoft ($\tau{=}0.5$) & $-1.93$ & $+1.6$ [$-0.1$, $+3.5$] & 0.0 & $-0.6$ [$-2.7$, $+1.6$] & $-0.7$ & \cellcolor[HTML]{FBE0DA} no \\
 & GumbelSoft ($\tau{=}0.3$) & $-2.00$ & $+0.4$ [$-1.5$, $+2.1$] & 0.0 & $+0.7$ [$-1.6$, $+2.9$] & $-0.8$ & \cellcolor[HTML]{FBE0DA} no \\
 & SynthID-Text & $-3.87$ & $+0.3$ [$-1.7$, $+2.3$] & 0.1 & $+5.1$ [$+2.8$, $+7.4$] & $-3.1$ & \cellcolor[HTML]{FBE0DA} no \\
 & Random keys & $-2.13$ & $+2.1$ [$+0.4$, $+3.8$] & 0.0 & $+0.9$ [$-1.3$, $+3.1$] & $-0.1$ & \cellcolor[HTML]{FBE0DA} no \\
 & Stratified keys & $-0.67$ & $+0.9$ [$-0.9$, $+2.6$] & 0.0 & $-1.8$ [$-3.8$, $+0.2$] & $-0.9$ & \cellcolor[HTML]{D9F0F2} yes \\
 & Context history & $-0.53$ & $-0.0$ [$-1.7$, $+1.7$] & 0.0 & $-0.4$ [$-2.6$, $+1.6$] & $+0.2$ & \cellcolor[HTML]{D9F0F2} yes \\
\rowcolor[HTML]{DCEBFA}  & SOLO & $-1.20$ & $+1.4$ [$-0.2$, $+3.1$] & 0.0 & $-2.6$ [$-4.6$, $-0.7$] & $+0.1$ & \cellcolor[HTML]{D9F0F2} yes \\
\midrule
TriviaQA, CoT & Fixed key & $-4.89$ & $-33.3$ [$-35.9$, $-30.5$] & 82.4 & $+94.3$ [$+92.8$, $+95.7$] & $-14.4$ & \cellcolor[HTML]{FBE0DA} no \\
 & Dedup \texttt{all} & $-4.06$ & $-11.3$ [$-14.1$, $-8.3$] & 26.9 & $+49.1$ [$+45.5$, $+52.5$] & $-6.1$ & \cellcolor[HTML]{FBE0DA} no \\
 & GumbelSoft ($\tau{=}0.5$) & $-2.67$ & $-5.0$ [$-7.2$, $-2.8$] & 8.2 & $+19.8$ [$+17.3$, $+22.4$] & $-4.2$ & \cellcolor[HTML]{FBE0DA} no \\
 & GumbelSoft ($\tau{=}0.3$) & $-3.78$ & $-9.5$ [$-11.8$, $-7.1$] & 17.5 & $+39.9$ [$+37.1$, $+42.7$] & $-6.6$ & \cellcolor[HTML]{FBE0DA} no \\
 & SynthID-Text & $-5.17$ & $-12.3$ [$-15.0$, $-9.8$] & 44.7 & $+54.1$ [$+50.9$, $+57.4$] & $-10.8$ & \cellcolor[HTML]{FBE0DA} no \\
 & Random keys & $-1.11$ & $-0.2$ [$-1.6$, $+1.2$] & 1.4 & $+5.2$ [$+3.4$, $+6.9$] & $-1.3$ & \cellcolor[HTML]{D9F0F2} yes \\
 & Stratified keys & $+0.67$ & $-0.9$ [$-2.2$, $+0.5$] & 1.1 & $-1.2$ [$-2.9$, $+0.3$] & $-0.1$ & \cellcolor[HTML]{D9F0F2} yes \\
 & Context history & $+0.28$ & $-0.3$ [$-1.6$, $+0.8$] & 1.2 & $-0.3$ [$-1.5$, $+0.8$] & $+0.6$ & \cellcolor[HTML]{D9F0F2} yes \\
\rowcolor[HTML]{DCEBFA}  & SOLO & $+0.00$ & $-0.0$ [$-1.2$, $+1.1$] & 1.1 & $-0.5$ [$-1.7$, $+0.6$] & $+0.2$ & \cellcolor[HTML]{D9F0F2} yes \\
\midrule
TriviaQA, JSON & Fixed key & $-6.61$ & $-31.6$ [$-34.2$, $-28.9$] & 91.2 & $+95.4$ [$+94.4$, $+96.5$] & $-17.6$ & \cellcolor[HTML]{FBE0DA} no \\
 & Dedup \texttt{all} & $-6.72$ & $-31.0$ [$-33.7$, $-28.1$] & 90.9 & $+95.4$ [$+94.3$, $+96.4$] & $-17.5$ & \cellcolor[HTML]{FBE0DA} no \\
 & GumbelSoft ($\tau{=}0.5$) & $-4.89$ & $-3.1$ [$-5.2$, $-0.9$] & 2.7 & $+22.8$ [$+20.4$, $+25.3$] & $-6.2$ & \cellcolor[HTML]{FBE0DA} no \\
 & GumbelSoft ($\tau{=}0.3$) & $-5.78$ & $-6.7$ [$-8.9$, $-4.3$] & 8.4 & $+41.9$ [$+39.5$, $+44.6$] & $-8.6$ & \cellcolor[HTML]{FBE0DA} no \\
 & SynthID-Text & $-5.33$ & $-11.0$ [$-13.4$, $-8.4$] & 32.9 & $+54.4$ [$+51.5$, $+57.3$] & $-11.8$ & \cellcolor[HTML]{FBE0DA} no \\
 & Random keys & $-2.00$ & $-0.6$ [$-2.4$, $+1.0$] & 0.2 & $+6.9$ [$+5.2$, $+8.8$] & $-1.9$ & \cellcolor[HTML]{FBE0DA} no \\
 & Stratified keys & $-1.44$ & $-0.0$ [$-1.6$, $+1.5$] & 0.2 & $+2.1$ [$+0.4$, $+3.7$] & $-0.3$ & \cellcolor[HTML]{D9F0F2} yes \\
 & Context history & $-0.28$ & $+0.9$ [$-0.3$, $+2.2$] & 0.3 & $+0.8$ [$-0.5$, $+2.1$] & $-0.6$ & \cellcolor[HTML]{D9F0F2} yes \\
\rowcolor[HTML]{DCEBFA}  & SOLO & $-0.28$ & $+0.4$ [$-1.0$, $+1.7$] & 0.3 & $+1.6$ [$+0.2$, $+2.9$] & $+0.4$ & \cellcolor[HTML]{D9F0F2} yes \\
\bottomrule
\end{tabular}}
\caption{Repeated-sampling metrics of every watermark relative to unwatermarked sampling, Qwen2.5-7B (three seeds, $K = 10$, 600 items per GSM8K and TriviaQA task and 500 per MATH-500 task, paired bootstrap over items). $\Delta$maj: maj@$K$ difference in points. $\Delta$AUROC: vote-share AUROC difference in points with 95\% CI. Ident.: requests with $K$ identical responses (\%). Infl.: pairwise-agreement inflation $(A - A_0)/(1 - A_0)$ with 95\% CI, the quantity of \cref{thm:inflation}. $\Delta$cov.: difference in coverage, the fraction of requests with at least one correct response (pass@$K$, points). Adm.: admissible for repeated sampling on the task by the rule of \cref{sec:exp-setup} (yes$^{\S}$: misses the rule but is $K$-sample distortion-free by \cref{thm:exact} and counted as admissible).}\label{tab:uqsupp}
\end{table*}

\begin{table*}[!t]
\centering
{\scriptsize\setlength{\tabcolsep}{6pt}
\begin{tabular}{llrrrrrc}
\toprule
Task & Watermark & $\Delta$maj & $\Delta$AUROC [95\% CI] & Ident. & Infl. (\%) [95\% CI] & $\Delta$cov. & Adm. \\
\midrule
GSM8K & Fixed key & $-1.83$ & $-4.1$ [$-6.6$, $-1.8$] & 11.4 & $+28.6$ [$+24.5$, $+32.7$] & $-2.6$ & \cellcolor[HTML]{FBE0DA} no \\
 & Dedup \texttt{all} & $-1.06$ & $-0.1$ [$-1.8$, $+1.4$] & 0.6 & $+7.8$ [$+4.3$, $+11.3$] & $-0.8$ & \cellcolor[HTML]{D9F0F2} yes \\
 & GumbelSoft ($\tau{=}0.5$) & $-0.11$ & $-1.6$ [$-3.7$, $+0.3$] & 0.0 & $+7.7$ [$+4.5$, $+10.8$] & $+1.2$ & \cellcolor[HTML]{D9F0F2} yes \\
 & GumbelSoft ($\tau{=}0.3$) & $-0.22$ & $-1.2$ [$-3.1$, $+0.6$] & 0.0 & $+10.9$ [$+7.8$, $+14.1$] & $+0.6$ & \cellcolor[HTML]{D9F0F2} yes \\
 & SynthID-Text & $-2.78$ & $+0.4$ [$-1.6$, $+2.4$] & 2.8 & $+5.3$ [$+1.7$, $+9.0$] & $-1.3$ & \cellcolor[HTML]{FBE0DA} no \\
 & Random keys & $-0.33$ & $+0.1$ [$-1.4$, $+1.4$] & 0.0 & $+0.3$ [$-2.9$, $+3.4$] & $+0.4$ & \cellcolor[HTML]{D9F0F2} yes \\
 & Stratified keys & $+0.06$ & $-0.5$ [$-2.1$, $+0.9$] & 0.0 & $+1.3$ [$-1.5$, $+4.1$] & $+0.6$ & \cellcolor[HTML]{D9F0F2} yes \\
 & Context history & $-0.17$ & $-0.1$ [$-1.4$, $+1.2$] & 0.0 & $+0.7$ [$-2.4$, $+3.7$] & $+0.8$ & \cellcolor[HTML]{D9F0F2} yes \\
\rowcolor[HTML]{DCEBFA}  & SOLO & $+0.17$ & $-1.0$ [$-3.0$, $+0.6$] & 0.0 & $+1.9$ [$-1.3$, $+4.9$] & $+0.9$ & \cellcolor[HTML]{D9F0F2} yes \\
\midrule
GSM8K, JSON & Fixed key & $-2.17$ & $-17.2$ [$-22.3$, $-12.1$] & 35.6 & $+63.5$ [$+57.5$, $+69.1$] & $-4.7$ & \cellcolor[HTML]{FBE0DA} no \\
 & Dedup \texttt{all} & $-2.39$ & $-17.4$ [$-22.4$, $-12.3$] & 33.5 & $+63.5$ [$+57.7$, $+69.1$] & $-4.8$ & \cellcolor[HTML]{FBE0DA} no \\
 & GumbelSoft ($\tau{=}0.5$) & $-1.00$ & $-2.6$ [$-6.1$, $+0.5$] & 0.1 & $+11.6$ [$+5.8$, $+17.1$] & $-0.9$ & \cellcolor[HTML]{D9F0F2} yes \\
 & GumbelSoft ($\tau{=}0.3$) & $-1.33$ & $-3.6$ [$-7.1$, $-0.1$] & 0.3 & $+18.2$ [$+12.6$, $+24.3$] & $-1.2$ & \cellcolor[HTML]{FBE0DA} no \\
 & SynthID-Text & $-1.56$ & $-7.4$ [$-12.0$, $-3.3$] & 7.6 & $+40.3$ [$+34.1$, $+46.4$] & $-2.6$ & \cellcolor[HTML]{FBE0DA} no \\
 & Random keys & $+0.11$ & $-2.5$ [$-5.4$, $+0.4$] & 0.0 & $+3.8$ [$-1.6$, $+9.3$] & $-0.3$ & \cellcolor[HTML]{D9F0F2} yes \\
 & Stratified keys & $+0.39$ & $-2.1$ [$-4.5$, $+0.5$] & 0.0 & $+2.6$ [$-2.3$, $+7.1$] & $-0.1$ & \cellcolor[HTML]{D9F0F2} yes \\
 & Context history & $+0.00$ & $-2.3$ [$-5.2$, $+0.4$] & 0.0 & $+2.2$ [$-2.5$, $+6.9$] & $-0.3$ & \cellcolor[HTML]{D9F0F2} yes \\
\rowcolor[HTML]{DCEBFA}  & SOLO & $+0.44$ & $-2.2$ [$-5.6$, $+0.8$] & 0.0 & $+4.8$ [$-0.1$, $+9.5$] & $+0.0$ & \cellcolor[HTML]{D9F0F2} yes \\
\midrule
MATH-500, JSON & Fixed key & $-6.27$ & $-3.9$ [$-6.3$, $-1.5$] & 8.7 & $+32.4$ [$+28.8$, $+36.3$] & $-6.9$ & \cellcolor[HTML]{FBE0DA} no \\
 & Dedup \texttt{all} & $-6.27$ & $-4.0$ [$-6.4$, $-1.5$] & 8.3 & $+33.1$ [$+29.5$, $+37.1$] & $-7.1$ & \cellcolor[HTML]{FBE0DA} no \\
 & GumbelSoft ($\tau{=}0.5$) & $-2.40$ & $+0.6$ [$-0.8$, $+2.2$] & 0.0 & $+3.0$ [$+0.2$, $+5.7$] & $-0.7$ & \cellcolor[HTML]{FBE0DA} no \\
 & GumbelSoft ($\tau{=}0.3$) & $-3.53$ & $-0.7$ [$-2.6$, $+1.1$] & 0.0 & $+8.1$ [$+5.1$, $+11.1$] & $-1.7$ & \cellcolor[HTML]{FBE0DA} no \\
 & SynthID-Text & $-4.60$ & $-0.5$ [$-2.7$, $+1.6$] & 1.6 & $+14.7$ [$+11.4$, $+18.2$] & $-3.1$ & \cellcolor[HTML]{FBE0DA} no \\
 & Random keys & $-1.53$ & $+1.5$ [$+0.1$, $+2.9$] & 0.0 & $+2.0$ [$-0.6$, $+4.4$] & $-0.7$ & \cellcolor[HTML]{FBE0DA} no \\
 & Stratified keys & $-1.47$ & $+1.6$ [$+0.2$, $+3.3$] & 0.0 & $-0.1$ [$-2.8$, $+2.3$] & $-0.9$ & \cellcolor[HTML]{D9F0F2} yes \\
 & Context history & $-2.20$ & $+1.6$ [$+0.3$, $+3.1$] & 0.0 & $-1.5$ [$-3.9$, $+0.7$] & $+0.0$ & \cellcolor[HTML]{D9F0F2} yes$^{\S}$ \\
\rowcolor[HTML]{DCEBFA}  & SOLO & $-1.40$ & $+1.8$ [$+0.3$, $+3.6$] & 0.0 & $+0.5$ [$-1.9$, $+2.9$] & $-0.3$ & \cellcolor[HTML]{D9F0F2} yes \\
\midrule
TriviaQA, CoT & Fixed key & $-7.00$ & $-28.1$ [$-30.9$, $-25.2$] & 82.1 & $+89.5$ [$+87.8$, $+91.3$] & $-21.6$ & \cellcolor[HTML]{FBE0DA} no \\
 & Dedup \texttt{all} & $-3.28$ & $-7.4$ [$-10.1$, $-4.8$] & 11.9 & $+29.4$ [$+26.7$, $+32.4$] & $-6.4$ & \cellcolor[HTML]{FBE0DA} no \\
 & GumbelSoft ($\tau{=}0.5$) & $-4.83$ & $-4.4$ [$-6.8$, $-2.1$] & 3.2 & $+17.4$ [$+15.3$, $+19.6$] & $-6.3$ & \cellcolor[HTML]{FBE0DA} no \\
 & GumbelSoft ($\tau{=}0.3$) & $-5.44$ & $-7.1$ [$-9.8$, $-4.6$] & 10.2 & $+33.7$ [$+31.4$, $+36.1$] & $-9.1$ & \cellcolor[HTML]{FBE0DA} no \\
 & SynthID-Text & $-4.83$ & $-6.4$ [$-8.7$, $-4.2$] & 27.3 & $+37.2$ [$+34.5$, $+39.8$] & $-11.9$ & \cellcolor[HTML]{FBE0DA} no \\
 & Random keys & $-2.11$ & $+0.0$ [$-1.6$, $+1.7$] & 0.3 & $+6.3$ [$+5.1$, $+7.6$] & $-2.1$ & \cellcolor[HTML]{FBE0DA} no \\
 & Stratified keys & $-1.39$ & $+0.7$ [$-0.8$, $+2.4$] & 0.1 & $+0.2$ [$-1.0$, $+1.3$] & $+0.2$ & \cellcolor[HTML]{D9F0F2} yes \\
 & Context history & $-0.39$ & $-0.3$ [$-1.8$, $+1.3$] & 0.1 & $-0.2$ [$-1.1$, $+0.6$] & $+0.2$ & \cellcolor[HTML]{D9F0F2} yes \\
\rowcolor[HTML]{DCEBFA}  & SOLO & $-1.39$ & $+1.0$ [$-0.3$, $+2.4$] & 0.1 & $-0.1$ [$-0.9$, $+0.8$] & $+0.0$ & \cellcolor[HTML]{D9F0F2} yes \\
\midrule
TriviaQA, JSON & Fixed key & $-7.28$ & $-33.3$ [$-35.6$, $-30.9$] & 87.6 & $+94.2$ [$+93.0$, $+95.3$] & $-21.9$ & \cellcolor[HTML]{FBE0DA} no \\
 & Dedup \texttt{all} & $-7.28$ & $-32.6$ [$-35.1$, $-30.1$] & 84.7 & $+93.7$ [$+92.5$, $+94.7$] & $-21.7$ & \cellcolor[HTML]{FBE0DA} no \\
 & GumbelSoft ($\tau{=}0.5$) & $-4.94$ & $-4.1$ [$-6.0$, $-2.2$] & 1.8 & $+14.5$ [$+12.7$, $+16.3$] & $-5.6$ & \cellcolor[HTML]{FBE0DA} no \\
 & GumbelSoft ($\tau{=}0.3$) & $-6.78$ & $-7.0$ [$-9.2$, $-5.0$] & 6.2 & $+31.0$ [$+29.0$, $+33.0$] & $-10.0$ & \cellcolor[HTML]{FBE0DA} no \\
 & SynthID-Text & $-2.78$ & $-8.1$ [$-10.3$, $-5.7$] & 17.1 & $+35.0$ [$+32.5$, $+37.6$] & $-9.5$ & \cellcolor[HTML]{FBE0DA} no \\
 & Random keys & $-1.44$ & $+0.4$ [$-1.1$, $+1.8$] & 0.1 & $+6.1$ [$+4.9$, $+7.3$] & $-1.3$ & \cellcolor[HTML]{D9F0F2} yes \\
 & Stratified keys & $+0.22$ & $+0.3$ [$-1.0$, $+1.7$] & 0.0 & $+0.2$ [$-0.9$, $+1.4$] & $+0.9$ & \cellcolor[HTML]{D9F0F2} yes \\
 & Context history & $-0.72$ & $+0.1$ [$-1.3$, $+1.4$] & 0.0 & $-0.5$ [$-1.6$, $+0.6$] & $+1.0$ & \cellcolor[HTML]{D9F0F2} yes \\
\rowcolor[HTML]{DCEBFA}  & SOLO & $+0.28$ & $-1.0$ [$-2.2$, $+0.4$] & 0.0 & $-0.3$ [$-1.4$, $+0.7$] & $+0.8$ & \cellcolor[HTML]{D9F0F2} yes \\
\bottomrule
\end{tabular}}
\caption{Repeated-sampling metrics of every watermark relative to unwatermarked sampling, Olmo-3-7B (continued, see \cref{tab:uqsupp} for the columns).}\label{tab:uqsupp-olmo7b}
\end{table*}

\begin{table*}[!t]
\centering
{\scriptsize\setlength{\tabcolsep}{6pt}
\begin{tabular}{llrrrrrc}
\toprule
Task & Watermark & $\Delta$maj & $\Delta$AUROC [95\% CI] & Ident. & Infl. (\%) [95\% CI] & $\Delta$cov. & Adm. \\
\midrule
GSM8K & Fixed key & $-3.00$ & $-2.0$ [$-5.3$, $+1.0$] & 3.8 & $+17.3$ [$+12.0$, $+22.5$] & $-3.3$ & \cellcolor[HTML]{FBE0DA} no \\
 & Dedup \texttt{all} & $-1.00$ & $-2.2$ [$-4.4$, $-0.1$] & 0.1 & $+5.4$ [$+0.7$, $+10.1$] & $-0.8$ & \cellcolor[HTML]{FBE0DA} no \\
 & GumbelSoft ($\tau{=}0.5$) & $-0.78$ & $+0.4$ [$-1.1$, $+2.0$] & 0.0 & $-10.3$ [$-14.8$, $-6.2$] & $-0.7$ & \cellcolor[HTML]{D9F0F2} yes \\
 & GumbelSoft ($\tau{=}0.3$) & $-0.61$ & $-0.4$ [$-2.7$, $+1.7$] & 0.0 & $-1.1$ [$-5.3$, $+2.8$] & $-0.7$ & \cellcolor[HTML]{D9F0F2} yes \\
 & SynthID-Text & $-3.67$ & $+0.2$ [$-2.1$, $+2.5$] & 0.1 & $-13.7$ [$-19.5$, $-8.3$] & $-2.9$ & \cellcolor[HTML]{FBE0DA} no \\
 & Random keys & $+0.83$ & $-0.8$ [$-2.4$, $+0.5$] & 0.0 & $+0.2$ [$-3.7$, $+4.0$] & $+0.1$ & \cellcolor[HTML]{D9F0F2} yes \\
 & Stratified keys & $+0.44$ & $-1.8$ [$-4.1$, $+0.1$] & 0.0 & $-0.2$ [$-4.1$, $+3.6$] & $+0.4$ & \cellcolor[HTML]{D9F0F2} yes \\
 & Context history & $+0.44$ & $-0.4$ [$-1.8$, $+1.0$] & 0.0 & $-1.5$ [$-5.1$, $+2.1$] & $+0.4$ & \cellcolor[HTML]{D9F0F2} yes \\
\rowcolor[HTML]{DCEBFA}  & SOLO & $+0.17$ & $-1.0$ [$-2.8$, $+0.5$] & 0.0 & $-1.6$ [$-5.2$, $+1.4$] & $+0.0$ & \cellcolor[HTML]{D9F0F2} yes \\
\midrule
GSM8K, JSON & Fixed key & $-2.06$ & $-4.9$ [$-9.4$, $-0.6$] & 8.2 & $+38.4$ [$+32.7$, $+43.9$] & $-2.4$ & \cellcolor[HTML]{FBE0DA} no \\
 & Dedup \texttt{all} & $-1.83$ & $-6.0$ [$-10.4$, $-1.3$] & 7.8 & $+39.5$ [$+33.5$, $+45.0$] & $-2.5$ & \cellcolor[HTML]{FBE0DA} no \\
 & GumbelSoft ($\tau{=}0.5$) & $-1.11$ & $+0.7$ [$-2.3$, $+4.3$] & 0.0 & $-6.2$ [$-12.9$, $+0.1$] & $-0.7$ & \cellcolor[HTML]{D9F0F2} yes \\
 & GumbelSoft ($\tau{=}0.3$) & $-0.72$ & $-0.7$ [$-4.4$, $+3.0$] & 0.0 & $+6.8$ [$+1.2$, $+12.3$] & $-0.3$ & \cellcolor[HTML]{D9F0F2} yes \\
 & SynthID-Text & $-2.67$ & $-2.4$ [$-6.1$, $+1.5$] & 0.3 & $-8.2$ [$-16.1$, $-1.3$] & $-1.2$ & \cellcolor[HTML]{FBE0DA} no \\
 & Random keys & $-0.78$ & $+0.6$ [$-2.4$, $+3.5$] & 0.0 & $+4.0$ [$-1.0$, $+8.9$] & $-0.2$ & \cellcolor[HTML]{D9F0F2} yes \\
 & Stratified keys & $-0.06$ & $+0.9$ [$-2.1$, $+3.8$] & 0.0 & $+0.9$ [$-4.3$, $+6.1$] & $-0.1$ & \cellcolor[HTML]{D9F0F2} yes \\
 & Context history & $+0.06$ & $+1.7$ [$-1.1$, $+5.0$] & 0.0 & $+0.3$ [$-4.7$, $+5.4$] & $-0.1$ & \cellcolor[HTML]{D9F0F2} yes \\
\rowcolor[HTML]{DCEBFA}  & SOLO & $+0.00$ & $-1.2$ [$-4.6$, $+2.0$] & 0.0 & $-8.5$ [$-14.8$, $-3.0$] & $-0.3$ & \cellcolor[HTML]{D9F0F2} yes \\
\midrule
MATH-500, JSON & Fixed key & $-2.33$ & $-0.2$ [$-3.4$, $+2.8$] & 1.1 & $+19.6$ [$+15.6$, $+23.6$] & $-3.1$ & \cellcolor[HTML]{FBE0DA} no \\
 & Dedup \texttt{all} & $-1.80$ & $-1.8$ [$-5.0$, $+1.3$] & 1.3 & $+19.7$ [$+15.5$, $+23.6$] & $-3.7$ & \cellcolor[HTML]{FBE0DA} no \\
 & GumbelSoft ($\tau{=}0.5$) & $-2.73$ & $-2.5$ [$-5.5$, $+0.4$] & 0.0 & $-7.0$ [$-11.6$, $-3.1$] & $-1.7$ & \cellcolor[HTML]{FBE0DA} no \\
 & GumbelSoft ($\tau{=}0.3$) & $-1.13$ & $+0.0$ [$-2.6$, $+3.0$] & 0.0 & $+5.6$ [$+1.6$, $+9.3$] & $-1.3$ & \cellcolor[HTML]{D9F0F2} yes \\
 & SynthID-Text & $-5.93$ & $-5.4$ [$-8.7$, $-2.0$] & 0.0 & $-1.1$ [$-6.0$, $+3.4$] & $-3.4$ & \cellcolor[HTML]{FBE0DA} no \\
 & Random keys & $+0.67$ & $-2.8$ [$-5.5$, $-0.0$] & 0.0 & $-2.2$ [$-5.9$, $+1.4$] & $-0.5$ & \cellcolor[HTML]{FBE0DA} no \\
 & Stratified keys & $+0.40$ & $-1.7$ [$-4.9$, $+1.2$] & 0.0 & $-3.8$ [$-7.5$, $-0.3$] & $-0.1$ & \cellcolor[HTML]{D9F0F2} yes \\
 & Context history & $+0.27$ & $-0.4$ [$-2.9$, $+2.5$] & 0.0 & $-0.0$ [$-3.4$, $+3.4$] & $-0.3$ & \cellcolor[HTML]{D9F0F2} yes \\
\rowcolor[HTML]{DCEBFA}  & SOLO & $-0.67$ & $-0.8$ [$-3.7$, $+2.1$] & 0.0 & $-4.2$ [$-8.0$, $-0.8$] & $-1.0$ & \cellcolor[HTML]{D9F0F2} yes \\
\midrule
TriviaQA, CoT & Fixed key & $-6.50$ & $-22.9$ [$-25.8$, $-20.1$] & 49.7 & $+72.8$ [$+70.6$, $+75.1$] & $-15.6$ & \cellcolor[HTML]{FBE0DA} no \\
 & Dedup \texttt{all} & $-2.61$ & $-4.9$ [$-7.0$, $-2.9$] & 1.4 & $+15.1$ [$+13.2$, $+17.1$] & $-1.9$ & \cellcolor[HTML]{FBE0DA} no \\
 & GumbelSoft ($\tau{=}0.5$) & $-3.06$ & $-4.3$ [$-6.3$, $-2.4$] & 0.1 & $+6.1$ [$+4.3$, $+7.8$] & $-1.9$ & \cellcolor[HTML]{FBE0DA} no \\
 & GumbelSoft ($\tau{=}0.3$) & $-3.67$ & $-6.1$ [$-8.0$, $-4.1$] & 1.1 & $+18.8$ [$+16.7$, $+20.8$] & $-3.4$ & \cellcolor[HTML]{FBE0DA} no \\
 & SynthID-Text & $-3.94$ & $-7.7$ [$-10.1$, $-5.2$] & 12.7 & $+30.7$ [$+28.1$, $+33.5$] & $-7.1$ & \cellcolor[HTML]{FBE0DA} no \\
 & Random keys & $-1.56$ & $-0.6$ [$-2.1$, $+1.0$] & 0.0 & $+6.8$ [$+5.4$, $+8.2$] & $-0.9$ & \cellcolor[HTML]{FBE0DA} no \\
 & Stratified keys & $-0.33$ & $-0.1$ [$-1.4$, $+1.3$] & 0.0 & $+3.3$ [$+1.9$, $+4.5$] & $-0.4$ & \cellcolor[HTML]{D9F0F2} yes \\
 & Context history & $-0.67$ & $-0.6$ [$-1.8$, $+0.8$] & 0.0 & $-0.8$ [$-2.1$, $+0.3$] & $-0.3$ & \cellcolor[HTML]{D9F0F2} yes \\
\rowcolor[HTML]{DCEBFA}  & SOLO & $-0.67$ & $-0.1$ [$-1.5$, $+1.1$] & 0.0 & $+1.4$ [$+0.3$, $+2.6$] & $+0.2$ & \cellcolor[HTML]{D9F0F2} yes \\
\midrule
TriviaQA, JSON & Fixed key & $-7.50$ & $-27.7$ [$-30.0$, $-25.4$] & 57.4 & $+83.0$ [$+81.2$, $+84.8$] & $-18.1$ & \cellcolor[HTML]{FBE0DA} no \\
 & Dedup \texttt{all} & $-7.39$ & $-24.4$ [$-26.9$, $-22.1$] & 43.7 & $+80.1$ [$+78.4$, $+81.8$] & $-16.5$ & \cellcolor[HTML]{FBE0DA} no \\
 & GumbelSoft ($\tau{=}0.5$) & $-7.22$ & $-6.6$ [$-8.8$, $-4.5$] & 0.0 & $+10.8$ [$+8.7$, $+13.0$] & $-4.6$ & \cellcolor[HTML]{FBE0DA} no \\
 & GumbelSoft ($\tau{=}0.3$) & $-7.39$ & $-8.7$ [$-11.0$, $-6.4$] & 0.0 & $+28.2$ [$+25.8$, $+30.4$] & $-7.6$ & \cellcolor[HTML]{FBE0DA} no \\
 & SynthID-Text & $-4.44$ & $-7.7$ [$-10.1$, $-5.4$] & 3.4 & $+40.6$ [$+37.7$, $+43.3$] & $-8.8$ & \cellcolor[HTML]{FBE0DA} no \\
 & Random keys & $-2.50$ & $-1.0$ [$-2.6$, $+0.6$] & 0.0 & $+3.5$ [$+1.7$, $+5.2$] & $-1.4$ & \cellcolor[HTML]{FBE0DA} no \\
 & Stratified keys & $-1.94$ & $+0.5$ [$-0.9$, $+2.1$] & 0.0 & $-1.4$ [$-3.1$, $+0.3$] & $+0.3$ & \cellcolor[HTML]{D9F0F2} yes$^{\S}$ \\
 & Context history & $-0.22$ & $+0.0$ [$-1.3$, $+1.3$] & 0.0 & $-0.7$ [$-2.2$, $+0.9$] & $-0.1$ & \cellcolor[HTML]{D9F0F2} yes \\
\rowcolor[HTML]{DCEBFA}  & SOLO & $-0.50$ & $-0.3$ [$-1.6$, $+1.0$] & 0.0 & $+0.1$ [$-1.5$, $+1.6$] & $+0.6$ & \cellcolor[HTML]{D9F0F2} yes \\
\bottomrule
\end{tabular}}
\caption{Repeated-sampling metrics of every watermark relative to unwatermarked sampling, Ministral-3-8B (continued, see \cref{tab:uqsupp} for the columns).}\label{tab:uqsupp-min8b}
\end{table*}

\begin{table*}[!t]
\centering
{\scriptsize\setlength{\tabcolsep}{6pt}
\begin{tabular}{llrrrrrc}
\toprule
Task & Watermark & $\Delta$maj & $\Delta$AUROC [95\% CI] & Ident. & Infl. (\%) [95\% CI] & $\Delta$cov. & Adm. \\
\midrule
GSM8K & Fixed key & $-1.83$ & $-3.5$ [$-7.8$, $+1.5$] & 0.7 & $+22.7$ [$+15.6$, $+29.9$] & $-1.6$ & \cellcolor[HTML]{FBE0DA} no \\
 & Dedup \texttt{all} & $-0.50$ & $-0.5$ [$-3.5$, $+2.8$] & 0.0 & $+4.8$ [$-1.6$, $+11.2$] & $-0.3$ & \cellcolor[HTML]{D9F0F2} yes \\
 & GumbelSoft ($\tau{=}0.5$) & $-0.22$ & $-1.4$ [$-5.2$, $+2.0$] & 0.0 & $-8.1$ [$-14.7$, $-2.0$] & $-0.2$ & \cellcolor[HTML]{D9F0F2} yes \\
 & GumbelSoft ($\tau{=}0.3$) & $-0.72$ & $-0.9$ [$-3.7$, $+2.1$] & 0.1 & $-1.9$ [$-8.6$, $+4.2$] & $+0.2$ & \cellcolor[HTML]{D9F0F2} yes \\
 & SynthID-Text & $-1.67$ & $+0.2$ [$-4.2$, $+4.9$] & 0.1 & $-4.9$ [$-12.3$, $+2.2$] & $-0.8$ & \cellcolor[HTML]{FBE0DA} no \\
 & Random keys & $-0.72$ & $-1.2$ [$-4.0$, $+2.0$] & 0.0 & $+2.8$ [$-3.0$, $+8.8$] & $+0.1$ & \cellcolor[HTML]{D9F0F2} yes \\
 & Stratified keys & $-0.22$ & $-1.5$ [$-5.3$, $+1.9$] & 0.0 & $+0.4$ [$-5.0$, $+5.6$] & $+0.3$ & \cellcolor[HTML]{D9F0F2} yes \\
 & Context history & $-0.17$ & $+1.1$ [$-1.7$, $+4.5$] & 0.0 & $-3.2$ [$-8.7$, $+2.1$] & $+0.2$ & \cellcolor[HTML]{D9F0F2} yes \\
\rowcolor[HTML]{DCEBFA}  & SOLO & $+0.28$ & $-2.0$ [$-5.6$, $+1.1$] & 0.0 & $-5.5$ [$-11.1$, $-0.0$] & $+0.2$ & \cellcolor[HTML]{D9F0F2} yes \\
\midrule
GSM8K, JSON & Fixed key & $-1.17$ & $-7.3$ [$-13.6$, $-1.1$] & 8.6 & $+43.8$ [$+34.6$, $+52.5$] & $-1.8$ & \cellcolor[HTML]{FBE0DA} no \\
 & Dedup \texttt{all} & $-0.89$ & $-6.7$ [$-13.3$, $-0.3$] & 5.2 & $+43.8$ [$+34.9$, $+52.7$] & $-1.9$ & \cellcolor[HTML]{FBE0DA} no \\
 & GumbelSoft ($\tau{=}0.5$) & $+0.06$ & $-0.6$ [$-6.7$, $+4.8$] & 0.0 & $-1.5$ [$-10.0$, $+6.4$] & $-0.6$ & \cellcolor[HTML]{D9F0F2} yes \\
 & GumbelSoft ($\tau{=}0.3$) & $+0.17$ & $-2.3$ [$-7.7$, $+2.6$] & 0.0 & $+11.9$ [$+3.5$, $+19.9$] & $-0.6$ & \cellcolor[HTML]{D9F0F2} yes \\
 & SynthID-Text & $-0.78$ & $-1.9$ [$-7.7$, $+3.6$] & 1.1 & $+20.7$ [$+13.1$, $+28.0$] & $-0.8$ & \cellcolor[HTML]{D9F0F2} yes \\
 & Random keys & $-0.06$ & $-0.3$ [$-5.8$, $+4.8$] & 0.0 & $+2.1$ [$-5.3$, $+9.0$] & $-0.3$ & \cellcolor[HTML]{D9F0F2} yes \\
 & Stratified keys & $+0.33$ & $+0.0$ [$-5.6$, $+5.4$] & 0.0 & $+4.9$ [$-1.5$, $+11.6$] & $-0.2$ & \cellcolor[HTML]{D9F0F2} yes \\
 & Context history & $+0.11$ & $+2.5$ [$-2.3$, $+7.1$] & 0.0 & $+0.5$ [$-6.7$, $+7.4$] & $-0.1$ & \cellcolor[HTML]{D9F0F2} yes \\
\rowcolor[HTML]{DCEBFA}  & SOLO & $+0.50$ & $-0.2$ [$-5.3$, $+4.7$] & 0.0 & $+3.0$ [$-3.5$, $+9.4$] & $-0.3$ & \cellcolor[HTML]{D9F0F2} yes \\
\midrule
MATH-500, JSON & Fixed key & $-2.80$ & $-1.4$ [$-3.2$, $+0.4$] & 0.9 & $+9.7$ [$+6.6$, $+12.9$] & $-2.7$ & \cellcolor[HTML]{FBE0DA} no \\
 & Dedup \texttt{all} & $-2.80$ & $-1.2$ [$-3.0$, $+0.5$] & 0.2 & $+6.9$ [$+4.0$, $+9.8$] & $-1.9$ & \cellcolor[HTML]{FBE0DA} no \\
 & GumbelSoft ($\tau{=}0.5$) & $-1.53$ & $-0.7$ [$-2.3$, $+0.8$] & 0.0 & $-2.5$ [$-5.1$, $-0.2$] & $-0.9$ & \cellcolor[HTML]{FBE0DA} no \\
 & GumbelSoft ($\tau{=}0.3$) & $-1.87$ & $+0.3$ [$-1.2$, $+1.9$] & 0.0 & $-0.4$ [$-2.9$, $+2.1$] & $-0.7$ & \cellcolor[HTML]{FBE0DA} no \\
 & SynthID-Text & $-3.60$ & $-0.1$ [$-1.8$, $+1.7$] & 0.1 & $+5.2$ [$+2.5$, $+7.9$] & $-1.3$ & \cellcolor[HTML]{FBE0DA} no \\
 & Random keys & $-0.87$ & $-0.5$ [$-2.2$, $+1.0$] & 0.0 & $-2.2$ [$-4.7$, $+0.3$] & $-0.5$ & \cellcolor[HTML]{D9F0F2} yes \\
 & Stratified keys & $-0.87$ & $+0.3$ [$-1.4$, $+1.9$] & 0.0 & $-2.6$ [$-5.0$, $-0.4$] & $+0.2$ & \cellcolor[HTML]{D9F0F2} yes \\
 & Context history & $+0.60$ & $-0.6$ [$-2.2$, $+0.9$] & 0.0 & $-2.4$ [$-4.7$, $-0.1$] & $+0.6$ & \cellcolor[HTML]{D9F0F2} yes \\
\rowcolor[HTML]{DCEBFA}  & SOLO & $-1.13$ & $+0.3$ [$-1.1$, $+1.8$] & 0.0 & $-1.7$ [$-4.2$, $+0.8$] & $+1.1$ & \cellcolor[HTML]{D9F0F2} yes \\
\midrule
TriviaQA, CoT & Fixed key & $-2.44$ & $-27.1$ [$-30.4$, $-23.7$] & 82.7 & $+86.8$ [$+84.2$, $+89.2$] & $-8.6$ & \cellcolor[HTML]{FBE0DA} no \\
 & Dedup \texttt{all} & $-2.22$ & $-10.6$ [$-13.8$, $-7.5$] & 34.2 & $+57.0$ [$+52.9$, $+61.1$] & $-4.8$ & \cellcolor[HTML]{FBE0DA} no \\
 & GumbelSoft ($\tau{=}0.5$) & $-1.50$ & $-3.6$ [$-5.8$, $-1.3$] & 8.2 & $+18.4$ [$+15.0$, $+21.7$] & $-2.1$ & \cellcolor[HTML]{FBE0DA} no \\
 & GumbelSoft ($\tau{=}0.3$) & $-1.61$ & $-7.4$ [$-9.8$, $-4.9$] & 18.7 & $+37.6$ [$+34.2$, $+41.2$] & $-3.8$ & \cellcolor[HTML]{FBE0DA} no \\
 & SynthID-Text & $-1.50$ & $-13.5$ [$-16.4$, $-10.5$] & 44.8 & $+58.5$ [$+54.9$, $+62.1$] & $-5.7$ & \cellcolor[HTML]{FBE0DA} no \\
 & Random keys & $-0.72$ & $-0.8$ [$-2.4$, $+0.8$] & 2.6 & $+3.3$ [$+0.7$, $+5.9$] & $-0.7$ & \cellcolor[HTML]{D9F0F2} yes \\
 & Stratified keys & $-0.44$ & $+0.3$ [$-1.3$, $+1.9$] & 1.6 & $-0.3$ [$-2.7$, $+2.1$] & $+0.6$ & \cellcolor[HTML]{D9F0F2} yes \\
 & Context history & $-0.61$ & $+0.7$ [$-0.4$, $+1.9$] & 1.6 & $+0.2$ [$-1.4$, $+1.8$] & $+0.1$ & \cellcolor[HTML]{D9F0F2} yes \\
\rowcolor[HTML]{DCEBFA}  & SOLO & $+0.06$ & $-0.5$ [$-1.7$, $+0.7$] & 1.5 & $+0.7$ [$-1.0$, $+2.5$] & $+0.1$ & \cellcolor[HTML]{D9F0F2} yes \\
\midrule
TriviaQA, JSON & Fixed key & $-2.39$ & $-30.2$ [$-33.3$, $-27.3$] & 85.8 & $+92.6$ [$+90.8$, $+94.3$] & $-11.9$ & \cellcolor[HTML]{FBE0DA} no \\
 & Dedup \texttt{all} & $-2.22$ & $-30.7$ [$-33.7$, $-27.7$] & 85.4 & $+92.9$ [$+91.3$, $+94.5$] & $-11.8$ & \cellcolor[HTML]{FBE0DA} no \\
 & GumbelSoft ($\tau{=}0.5$) & $-1.44$ & $-3.4$ [$-5.6$, $-1.3$] & 1.3 & $+19.1$ [$+16.2$, $+22.1$] & $-3.3$ & \cellcolor[HTML]{FBE0DA} no \\
 & GumbelSoft ($\tau{=}0.3$) & $-1.83$ & $-7.6$ [$-9.9$, $-5.4$] & 5.8 & $+39.4$ [$+36.6$, $+42.5$] & $-5.6$ & \cellcolor[HTML]{FBE0DA} no \\
 & SynthID-Text & $-1.83$ & $-14.9$ [$-17.6$, $-12.4$] & 30.3 & $+61.1$ [$+58.1$, $+64.2$] & $-7.8$ & \cellcolor[HTML]{FBE0DA} no \\
 & Random keys & $+0.17$ & $-0.4$ [$-2.0$, $+1.2$] & 0.0 & $+5.4$ [$+3.0$, $+7.7$] & $-1.0$ & \cellcolor[HTML]{D9F0F2} yes \\
 & Stratified keys & $+0.44$ & $-0.6$ [$-2.1$, $+0.8$] & 0.0 & $+0.2$ [$-2.0$, $+2.4$] & $+0.0$ & \cellcolor[HTML]{D9F0F2} yes \\
 & Context history & $+0.17$ & $-0.7$ [$-2.1$, $+0.7$] & 0.0 & $+1.2$ [$-0.5$, $+3.0$] & $-0.1$ & \cellcolor[HTML]{D9F0F2} yes \\
\rowcolor[HTML]{DCEBFA}  & SOLO & $+0.44$ & $-0.7$ [$-2.1$, $+0.7$] & 0.1 & $+1.8$ [$-0.0$, $+3.7$] & $+0.6$ & \cellcolor[HTML]{D9F0F2} yes \\
\bottomrule
\end{tabular}}
\caption{Repeated-sampling metrics of every watermark relative to unwatermarked sampling, Qwen2.5-14B (continued, see \cref{tab:uqsupp} for the columns).}\label{tab:uqsupp-q14b}
\end{table*}

\begin{table*}[!t]
\centering
{\scriptsize\setlength{\tabcolsep}{6pt}
\begin{tabular}{llrrrrrc}
\toprule
Task & Watermark & $\Delta$maj & $\Delta$AUROC [95\% CI] & Ident. & Infl. (\%) [95\% CI] & $\Delta$cov. & Adm. \\
\midrule
GSM8K & Fixed key & $-2.83$ & $-1.4$ [$-4.5$, $+1.6$] & 3.1 & $+22.5$ [$+17.7$, $+27.2$] & $-4.2$ & \cellcolor[HTML]{FBE0DA} no \\
 & Dedup \texttt{all} & $-1.44$ & $-0.8$ [$-3.1$, $+1.6$] & 0.2 & $+5.4$ [$+1.4$, $+9.7$] & $-1.8$ & \cellcolor[HTML]{D9F0F2} yes \\
 & GumbelSoft ($\tau{=}0.5$) & $-0.94$ & $+0.1$ [$-1.7$, $+2.0$] & 0.0 & $-8.0$ [$-12.5$, $-3.6$] & $-1.3$ & \cellcolor[HTML]{D9F0F2} yes \\
 & GumbelSoft ($\tau{=}0.3$) & $-1.06$ & $+1.2$ [$-1.1$, $+4.1$] & 0.0 & $+4.1$ [$-0.6$, $+8.5$] & $-1.6$ & \cellcolor[HTML]{D9F0F2} yes \\
 & SynthID-Text & $-4.44$ & $+3.0$ [$+0.1$, $+6.1$] & 0.1 & $-10.4$ [$-16.3$, $-4.9$] & $-3.3$ & \cellcolor[HTML]{FBE0DA} no \\
 & Random keys & $+0.33$ & $+0.1$ [$-2.6$, $+2.5$] & 0.0 & $+6.4$ [$+2.1$, $+10.5$] & $-0.6$ & \cellcolor[HTML]{D9F0F2} yes \\
 & Stratified keys & $+0.06$ & $+2.1$ [$-0.2$, $+4.8$] & 0.0 & $+2.0$ [$-1.9$, $+5.8$] & $-0.5$ & \cellcolor[HTML]{D9F0F2} yes \\
 & Context history & $+0.28$ & $+0.6$ [$-1.5$, $+2.8$] & 0.0 & $+2.3$ [$-1.2$, $+5.8$] & $-0.6$ & \cellcolor[HTML]{D9F0F2} yes \\
\rowcolor[HTML]{DCEBFA}  & SOLO & $+0.39$ & $+1.3$ [$-1.7$, $+4.1$] & 0.0 & $-0.4$ [$-4.3$, $+3.5$] & $-0.1$ & \cellcolor[HTML]{D9F0F2} yes \\
\midrule
GSM8K, JSON & Fixed key & $-1.22$ & $-5.8$ [$-10.9$, $-0.7$] & 7.1 & $+49.3$ [$+43.1$, $+55.3$] & $-1.9$ & \cellcolor[HTML]{FBE0DA} no \\
 & Dedup \texttt{all} & $-1.17$ & $-4.3$ [$-9.7$, $+1.0$] & 7.4 & $+48.8$ [$+42.7$, $+55.1$] & $-2.1$ & \cellcolor[HTML]{D9F0F2} yes \\
 & GumbelSoft ($\tau{=}0.5$) & $-0.89$ & $+2.0$ [$-0.8$, $+5.5$] & 0.0 & $+8.2$ [$+2.6$, $+13.3$] & $-0.2$ & \cellcolor[HTML]{D9F0F2} yes \\
 & GumbelSoft ($\tau{=}0.3$) & $-0.72$ & $-2.3$ [$-6.0$, $+1.4$] & 0.0 & $+18.1$ [$+12.5$, $+24.1$] & $-0.4$ & \cellcolor[HTML]{D9F0F2} yes \\
 & SynthID-Text & $-1.67$ & $-1.2$ [$-5.5$, $+3.5$] & 0.2 & $+9.7$ [$+3.1$, $+15.8$] & $-1.1$ & \cellcolor[HTML]{FBE0DA} no \\
 & Random keys & $-0.83$ & $-0.2$ [$-3.7$, $+3.2$] & 0.0 & $-2.8$ [$-9.2$, $+3.4$] & $-0.5$ & \cellcolor[HTML]{D9F0F2} yes \\
 & Stratified keys & $-0.50$ & $-0.0$ [$-4.4$, $+4.0$] & 0.0 & $-16.9$ [$-23.8$, $-10.9$] & $-0.4$ & \cellcolor[HTML]{D9F0F2} yes \\
 & Context history & $-0.17$ & $+1.2$ [$-2.1$, $+4.5$] & 0.0 & $+3.2$ [$-2.4$, $+9.0$] & $-0.3$ & \cellcolor[HTML]{D9F0F2} yes \\
\rowcolor[HTML]{DCEBFA}  & SOLO & $-0.39$ & $-0.9$ [$-5.5$, $+3.7$] & 0.0 & $+3.4$ [$-2.8$, $+9.0$] & $-0.3$ & \cellcolor[HTML]{D9F0F2} yes \\
\midrule
MATH-500, JSON & Fixed key & $-0.67$ & $-2.7$ [$-6.1$, $+0.5$] & 0.3 & $+24.7$ [$+20.6$, $+28.8$] & $-3.2$ & \cellcolor[HTML]{D9F0F2} yes \\
 & Dedup \texttt{all} & $-0.93$ & $-3.6$ [$-6.9$, $-0.5$] & 0.3 & $+25.0$ [$+20.7$, $+29.2$] & $-4.5$ & \cellcolor[HTML]{FBE0DA} no \\
 & GumbelSoft ($\tau{=}0.5$) & $-1.73$ & $-3.1$ [$-6.1$, $-0.2$] & 0.0 & $-1.7$ [$-6.2$, $+2.6$] & $-2.1$ & \cellcolor[HTML]{FBE0DA} no \\
 & GumbelSoft ($\tau{=}0.3$) & $+0.07$ & $-4.5$ [$-7.8$, $-1.4$] & 0.0 & $+9.9$ [$+5.6$, $+14.3$] & $-2.8$ & \cellcolor[HTML]{FBE0DA} no \\
 & SynthID-Text & $+0.53$ & $-2.8$ [$-5.8$, $-0.1$] & 0.0 & $+11.8$ [$+7.8$, $+15.8$] & $-2.2$ & \cellcolor[HTML]{FBE0DA} no \\
 & Random keys & $+0.53$ & $-3.6$ [$-6.7$, $-0.6$] & 0.0 & $-2.4$ [$-6.6$, $+1.6$] & $-0.3$ & \cellcolor[HTML]{FBE0DA} no \\
 & Stratified keys & $+1.33$ & $-3.8$ [$-6.6$, $-1.2$] & 0.0 & $+3.7$ [$-0.1$, $+7.2$] & $-0.7$ & \cellcolor[HTML]{D9F0F2} yes$^{\S}$ \\
 & Context history & $+1.27$ & $-3.5$ [$-6.5$, $-0.7$] & 0.0 & $-0.3$ [$-3.8$, $+3.0$] & $+0.3$ & \cellcolor[HTML]{D9F0F2} yes$^{\S}$ \\
\rowcolor[HTML]{DCEBFA}  & SOLO & $+1.00$ & $-0.6$ [$-3.6$, $+2.1$] & 0.0 & $-1.6$ [$-5.6$, $+2.2$] & $+0.0$ & \cellcolor[HTML]{D9F0F2} yes \\
\midrule
TriviaQA, CoT & Fixed key & $-5.22$ & $-18.8$ [$-21.6$, $-15.8$] & 39.0 & $+63.7$ [$+61.1$, $+66.3$] & $-11.7$ & \cellcolor[HTML]{FBE0DA} no \\
 & Dedup \texttt{all} & $-2.56$ & $-5.1$ [$-7.5$, $-2.6$] & 3.2 & $+21.5$ [$+19.0$, $+24.1$] & $-3.0$ & \cellcolor[HTML]{FBE0DA} no \\
 & GumbelSoft ($\tau{=}0.5$) & $-2.44$ & $-1.6$ [$-3.4$, $+0.3$] & 0.0 & $+0.7$ [$-1.5$, $+2.8$] & $-1.3$ & \cellcolor[HTML]{FBE0DA} no \\
 & GumbelSoft ($\tau{=}0.3$) & $-2.72$ & $-3.4$ [$-5.5$, $-1.3$] & 0.4 & $+13.4$ [$+11.0$, $+15.7$] & $-3.3$ & \cellcolor[HTML]{FBE0DA} no \\
 & SynthID-Text & $-2.89$ & $-4.5$ [$-6.8$, $-2.0$] & 6.2 & $+26.5$ [$+23.8$, $+29.2$] & $-4.3$ & \cellcolor[HTML]{FBE0DA} no \\
 & Random keys & $-1.89$ & $-1.1$ [$-2.6$, $+0.6$] & 0.0 & $+2.0$ [$+0.3$, $+3.5$] & $-1.4$ & \cellcolor[HTML]{FBE0DA} no \\
 & Stratified keys & $-1.17$ & $-1.0$ [$-2.5$, $+0.5$] & 0.0 & $-1.8$ [$-3.4$, $-0.3$] & $+0.1$ & \cellcolor[HTML]{D9F0F2} yes \\
 & Context history & $+0.11$ & $-0.2$ [$-1.8$, $+1.3$] & 0.0 & $-0.2$ [$-1.7$, $+1.2$] & $-0.3$ & \cellcolor[HTML]{D9F0F2} yes \\
\rowcolor[HTML]{DCEBFA}  & SOLO & $+0.22$ & $-0.9$ [$-2.3$, $+0.5$] & 0.0 & $-0.7$ [$-2.2$, $+0.7$] & $+0.2$ & \cellcolor[HTML]{D9F0F2} yes \\
\midrule
TriviaQA, JSON & Fixed key & $-10.94$ & $-25.5$ [$-28.6$, $-22.5$] & 55.3 & $+81.6$ [$+79.6$, $+83.5$] & $-16.8$ & \cellcolor[HTML]{FBE0DA} no \\
 & Dedup \texttt{all} & $-10.61$ & $-18.6$ [$-21.6$, $-15.4$] & 27.2 & $+73.9$ [$+71.7$, $+75.8$] & $-13.6$ & \cellcolor[HTML]{FBE0DA} no \\
 & GumbelSoft ($\tau{=}0.5$) & $-9.28$ & $-6.7$ [$-9.1$, $-4.3$] & 0.0 & $+10.0$ [$+7.2$, $+12.5$] & $-3.8$ & \cellcolor[HTML]{FBE0DA} no \\
 & GumbelSoft ($\tau{=}0.3$) & $-8.83$ & $-8.4$ [$-11.1$, $-6.0$] & 0.5 & $+25.2$ [$+22.6$, $+27.7$] & $-5.1$ & \cellcolor[HTML]{FBE0DA} no \\
 & SynthID-Text & $-4.61$ & $-6.0$ [$-8.3$, $-3.9$] & 5.1 & $+40.0$ [$+37.0$, $+42.9$] & $-5.1$ & \cellcolor[HTML]{FBE0DA} no \\
 & Random keys & $-2.94$ & $-1.6$ [$-3.4$, $+0.2$] & 0.0 & $+2.9$ [$+0.6$, $+5.1$] & $-0.4$ & \cellcolor[HTML]{FBE0DA} no \\
 & Stratified keys & $-1.83$ & $-0.6$ [$-2.0$, $+0.9$] & 0.0 & $-3.7$ [$-5.7$, $-1.6$] & $+0.8$ & \cellcolor[HTML]{D9F0F2} yes$^{\S}$ \\
 & Context history & $-0.72$ & $-0.7$ [$-2.2$, $+0.8$] & 0.0 & $-0.4$ [$-2.2$, $+1.4$] & $+0.9$ & \cellcolor[HTML]{D9F0F2} yes \\
\rowcolor[HTML]{DCEBFA}  & SOLO & $-0.44$ & $-0.3$ [$-1.8$, $+1.1$] & 0.0 & $-1.1$ [$-3.1$, $+1.0$] & $+0.7$ & \cellcolor[HTML]{D9F0F2} yes \\
\bottomrule
\end{tabular}}
\caption{Repeated-sampling metrics of every watermark relative to unwatermarked sampling, Ministral-3-14B (continued, see \cref{tab:uqsupp} for the columns).}\label{tab:uqsupp-min14b}
\end{table*}
\begin{table*}[!t]
\centering
\resizebox{\linewidth}{!}{%
\begin{tabular}{llccccccccc>{\columncolor[HTML]{DCEBFA}}c}
\toprule
Model & Task & None (\%) & Fixed key & Dedup \texttt{all} & GumbelSoft ($\tau{=}0.5$) & GumbelSoft ($\tau{=}0.3$) & SynthID-Text & Random keys & Stratified keys & Context history & SOLO \\
\midrule
Qwen2.5-7B & TriviaQA, CoT & 85.7 & \cellcolor[HTML]{FBE0DA} $-32.8$$^{*}$ & \cellcolor[HTML]{FBE0DA} $-10.1$$^{*}$ & \cellcolor[HTML]{FBE0DA} $-3.8$$^{*}$ & \cellcolor[HTML]{FBE0DA} $-9.0$$^{*}$ & \cellcolor[HTML]{FBE0DA} $-11.5$$^{*}$ & $+0.4$ & $-0.0$ & $+0.2$ & $+0.1$ \\
 & TriviaQA, JSON & 84.6 & \cellcolor[HTML]{FBE0DA} $-31.5$$^{*}$ & \cellcolor[HTML]{FBE0DA} $-30.9$$^{*}$ & \cellcolor[HTML]{FBE0DA} $-3.8$$^{*}$ & \cellcolor[HTML]{FBE0DA} $-7.3$$^{*}$ & \cellcolor[HTML]{FBE0DA} $-11.4$$^{*}$ & $-0.8$ & $-0.7$ & $+0.6$ & $-0.4$ \\
\midrule
Olmo-3-7B & TriviaQA, CoT & 84.4 & \cellcolor[HTML]{FBE0DA} $-29.2$$^{*}$ & \cellcolor[HTML]{FBE0DA} $-6.6$$^{*}$ & \cellcolor[HTML]{FBE0DA} $-3.7$$^{*}$ & \cellcolor[HTML]{FBE0DA} $-6.7$$^{*}$ & \cellcolor[HTML]{FBE0DA} $-7.0$$^{*}$ & $+0.6$ & $+0.9$ & $-0.2$ & $+0.8$ \\
 & TriviaQA, JSON & 86.6 & \cellcolor[HTML]{FBE0DA} $-33.3$$^{*}$ & \cellcolor[HTML]{FBE0DA} $-32.5$$^{*}$ & \cellcolor[HTML]{FBE0DA} $-3.2$$^{*}$ & \cellcolor[HTML]{FBE0DA} $-6.6$$^{*}$ & \cellcolor[HTML]{FBE0DA} $-8.0$$^{*}$ & $+0.7$ & $+0.5$ & $+0.3$ & $-0.8$ \\
\midrule
Ministral-3-8B & TriviaQA, CoT & 86.7 & \cellcolor[HTML]{FBE0DA} $-22.2$$^{*}$ & \cellcolor[HTML]{FBE0DA} $-2.5$$^{*}$ & \cellcolor[HTML]{FBE0DA} $-2.6$$^{*}$ & \cellcolor[HTML]{FBE0DA} $-4.5$$^{*}$ & \cellcolor[HTML]{FBE0DA} $-6.3$$^{*}$ & $+0.2$ & $-0.0$ & $-0.3$ & $+0.3$ \\
 & TriviaQA, JSON & 87.1 & \cellcolor[HTML]{FBE0DA} $-27.7$$^{*}$ & \cellcolor[HTML]{FBE0DA} $-24.1$$^{*}$ & \cellcolor[HTML]{FBE0DA} $-6.4$$^{*}$ & \cellcolor[HTML]{FBE0DA} $-7.9$$^{*}$ & \cellcolor[HTML]{FBE0DA} $-8.0$$^{*}$ & $-0.8$ & $+0.8$ & $-0.0$ & $-0.7$ \\
\midrule
Qwen2.5-14B & TriviaQA, CoT & 84.4 & \cellcolor[HTML]{FBE0DA} $-27.9$$^{*}$ & \cellcolor[HTML]{FBE0DA} $-11.3$$^{*}$ & \cellcolor[HTML]{FBE0DA} $-3.5$$^{*}$ & \cellcolor[HTML]{FBE0DA} $-7.3$$^{*}$ & \cellcolor[HTML]{FBE0DA} $-13.9$$^{*}$ & $-0.4$ & $+0.3$ & $+1.1$ & $-0.2$ \\
 & TriviaQA, JSON & 84.7 & \cellcolor[HTML]{FBE0DA} $-30.4$$^{*}$ & \cellcolor[HTML]{FBE0DA} $-30.6$$^{*}$ & \cellcolor[HTML]{FBE0DA} $-3.9$$^{*}$ & \cellcolor[HTML]{FBE0DA} $-8.1$$^{*}$ & \cellcolor[HTML]{FBE0DA} $-15.5$$^{*}$ & $-0.9$ & $-0.8$ & $-0.7$ & $-0.9$ \\
\midrule
Ministral-3-14B & TriviaQA, CoT & 85.3 & \cellcolor[HTML]{FBE0DA} $-17.9$$^{*}$ & \cellcolor[HTML]{FBE0DA} $-3.0$$^{*}$ & $-0.3$ & \cellcolor[HTML]{FBE0DA} $-2.3$$^{*}$ & \cellcolor[HTML]{FBE0DA} $-3.2$$^{*}$ & $+0.5$ & $+0.4$ & $-0.2$ & $-0.2$ \\
 & TriviaQA, JSON & 87.3 & \cellcolor[HTML]{FBE0DA} $-25.4$$^{*}$ & \cellcolor[HTML]{FBE0DA} $-18.9$$^{*}$ & \cellcolor[HTML]{FBE0DA} $-5.9$$^{*}$ & \cellcolor[HTML]{FBE0DA} $-7.3$$^{*}$ & \cellcolor[HTML]{FBE0DA} $-7.0$$^{*}$ & \cellcolor[HTML]{FBE0DA} $-1.7$$^{*}$ & $-1.4$ & $-1.2$ & $-0.5$ \\
\bottomrule
\end{tabular}
}
\caption{AUROC (\%) of semantic entropy on TriviaQA without watermark (None) and its difference in points for every watermark (600 held-out questions, $K = 10$, three seeds). Coral, $^{*}$: the 95\% CI of a paired item bootstrap excludes zero. Pale blue: SOLO.}\label{tab:se}
\end{table*}

\section{Repeated-Sampling Metrics of All Watermarks}\label{app:uq}
This appendix supports \cref{sec:exp-main}: per model and task, only the $K$-sample distortion-free schemes are admissible on every task.

\Cref{tab:uqsupp,tab:uqsupp-q14b,tab:uqsupp-olmo7b,tab:uqsupp-min8b,tab:uqsupp-min14b} list the repeated-sampling metrics behind the admissibility marks of \cref{tab:main}, one table per model (coral: not admissible, teal: admissible, pale blue: SOLO). Besides maj@$K$ and the vote-share AUROC, they report the fraction of requests with $K$ identical responses and the pairwise-agreement inflation of the extracted answers, the quantity that \cref{thm:inflation} predicts. Coupling raises the agreement. The inflation of SOLO stays below +5\% on every task, as does that of the other exact schemes, stratified keys and the context history. A fixed key inflates the agreement on every task, with an inflation of +10\% to +95\%. Coverage, the fraction of requests with at least one correct response and the quantity that bounds best-of-$n$ selection, drops by up to 21.9 points under a fixed key, 11.9 points under SynthID-Text, and 6.3 points under GumbelSoft at $\tau = 0.5$ (10.0 at $\tau = 0.3$), all on TriviaQA, whereas it changes by at most 1.1 points under SOLO.

\section{Semantic Entropy}\label{app:se}
This appendix supports \cref{sec:exp-main}: none of the exact schemes changes the semantic-entropy AUROC significantly.

\cref{tab:se} reports the AUROC of semantic entropy~\cite{kuhn2023semantic} for predicting the correctness of the majority answer on TriviaQA. Extracted answers are clustered by bidirectional entailment under DeBERTa-large-MNLI, and discrete semantic entropy is the entropy of the cluster frequencies among the ten responses. Because it is computed on the extracted answers, it differs from the entropy of the answer frequencies behind the vote share only where entailment merges differently worded answers. The fixed key, its \texttt{all} variant, SynthID-Text, and GumbelSoft lower the AUROC significantly on every model and format, except GumbelSoft at $\tau = 0.5$ with Ministral-3-14B in the CoT format, whereas SOLO stays within 1.0 points of unwatermarked sampling and the other two exact schemes within 1.4 points, none of them significantly.

\FloatBarrier
\clearpage
\nocite{yam2025myface,guo2026artwork,guo2024laaca,liu2026diffprotect,guo2025pca,lin2026priorasskill,qian2026tdcount,wang2026harmreduction,zhao2026exploring,liu2026agentsurvey,liu2026phase,zhao2026metacognitive,yang2026pointcloud,zhou2026posecompass,li2026fedkd,zhao2026p2p,ai2026hardlabel,duan2026gaitprotector,lei2026gazesync,li2026safetwinnet,li2026w2sdefense,liu2026graybox,guo2025thesis,hu2025syntactic,li2025guardfed,guo2025beyond,li2025fairseg,zhao2025roptester,lei2025transferable,li2025flsurvey,qian2025perspective,qian2025t2icount,lei2025instant,zhao2025backdoorsurvey,liu2024mmadpurify,zhao2024weaktostrong,guo2024signature,qian2024semisupervised,guo2023siamese}
\bibliographystyle{acl_natbib}
\bibliography{refs}

\end{document}